%% file: neurips_2026.tex
\documentclass{article}

\PassOptionsToPackage{numbers, compress}{natbib}

 \usepackage[main, final]{neurips_2026}

\usepackage[utf8]{inputenc} 
\usepackage[T1]{fontenc}    
\usepackage{hyperref}       
\usepackage{url}            
\usepackage{booktabs}       
\usepackage{amsfonts}       
\usepackage{nicefrac}       
\usepackage{microtype}      
\usepackage{xcolor}         

\usepackage[dvipsnames]{xcolor}
\usepackage[many]{tcolorbox}
\usepackage{amssymb}
\usepackage{colortbl}
\usepackage{pifont}
\usepackage{multirow}
\usepackage{graphicx}
\usepackage{tabularx}
\usepackage{threeparttable}
\usepackage{wrapfig}
\usepackage{enumitem}
\usepackage{subcaption}
\usepackage{gradient-text}

\usepackage{bm}
\usepackage{bbm}
\usepackage{wasysym}

\newcommand{\cmark}{\ding{51}}

\title{UniRAP: Towards Unified Part-level Physical Affordance Reasoning and Actionable Perception}

\author{
Linfei Li$^{1}$ \quad Ruining Hu$^{1}$
\quad Lin Zhang$^{1}$ \thanks{Corresponding author: Lin Zhang} 
\quad \textbf{Zhong Wang}$^{1}$  \\
\quad \textbf{Fengyi Zhang}$^{4}$
\quad \textbf{Ying Shen}$^{1}$ 
\quad \textbf{Binqiang Wang}$^{2}$ 
\quad \textbf{Xin Zhang}$^{3}$\\
$^1$School of Computer Science and Technology, Tongji University \\
$^2$IEIT SYSTEMS (Beijing) Co., Ltd \qquad $^3$NAIS, Neusoft Corporation \\
$^4$School of Electrical Engineering and Computer Science, The University of Queensland \\
\texttt{\{cslinfeili,2534075,cslinzhang,cszhongwang,yingshen\}@tongji.edu.cn} \\
\texttt{fengyi.zhang@uq.edu.au,wangbinqiang@ieisystem.com, zhang\_xin@neusoft.com} \\
\color{magenta}\url{https://lif314.github.io/projects/unirap}
}

\begin{document}

\maketitle

\input{sec/1_abstract}
\input{sec/2_introduction}
\input{sec/3_related_work}
\input{sec/4_method}
\input{sec/5_experiments}
\input{sec/6_conclusion}

\clearpage
\begin{ack}
This work was supported in part by the New Generation Artificial Intelligence-National Science and Technology Major Project under Grant 2025ZD0123701, in part by the National Natural Science Foundation of China under Grants 62672348, 62476202, and 62272343, in part by the Explorers Program of Shanghai (Basic Research Funding) from Shanghai Municipal Science and Technology Commission under Grant 26TS1403800, in part by the Innovation Ecosystem Development Program of Shanghai Municipal Science and Technology Commission under Grant 26510761200, and in part by the Fundamental Research Funds for the Central Universities.
\end{ack}
{
    \small
    \bibliographystyle{unsrt}
    \bibliography{main}
}



\input{sec/7_Appendix}


\newpage
\input{checklist.tex}

\end{document}

%% file: sec/1_abstract.tex
\begin{abstract}
Vision-language perception has achieved impressive progress in aligning natural language with visual observations, yet grounding high-level semantics into part-level physical interaction remains challenging. To address this gap, we propose UniRAP, a unified model for inferring part-level physical affordances and mapping language instructions to actionable geometric representations. UniRAP formulates this problem as a conditional multimodal generation task, integrating visual inputs, textual instructions, and optional prompts into a shared spatiotemporal representation space through a unified interface token mechanism. To predict executable contact geometry, we introduce a Unified Affordance Decoder (UAD), which jointly performs object detection, part-level affordance segmentation, and 4-DoF interaction pose estimation by leveraging intermediate segmentation features. In addition, we propose a curriculum-based transfer training strategy that progressively adapts the model from general visual parsing to interaction-aware perception, improving data efficiency under limited textual annotations. Experiments show that UniRAP achieves state-of-the-art performance on referring expression segmentation, affordance grounding, and interaction pose estimation, while maintaining strong spatiotemporal consistency in dynamic video scenarios. These results demonstrate the effectiveness of UniRAP as a unified perception framework for language-guided physical manipulation. 
\end{abstract}

%% file: sec/2_introduction.tex
\section{Introduction}

\begin{figure}[!ht]
\centering
\includegraphics[width=1.0\linewidth]{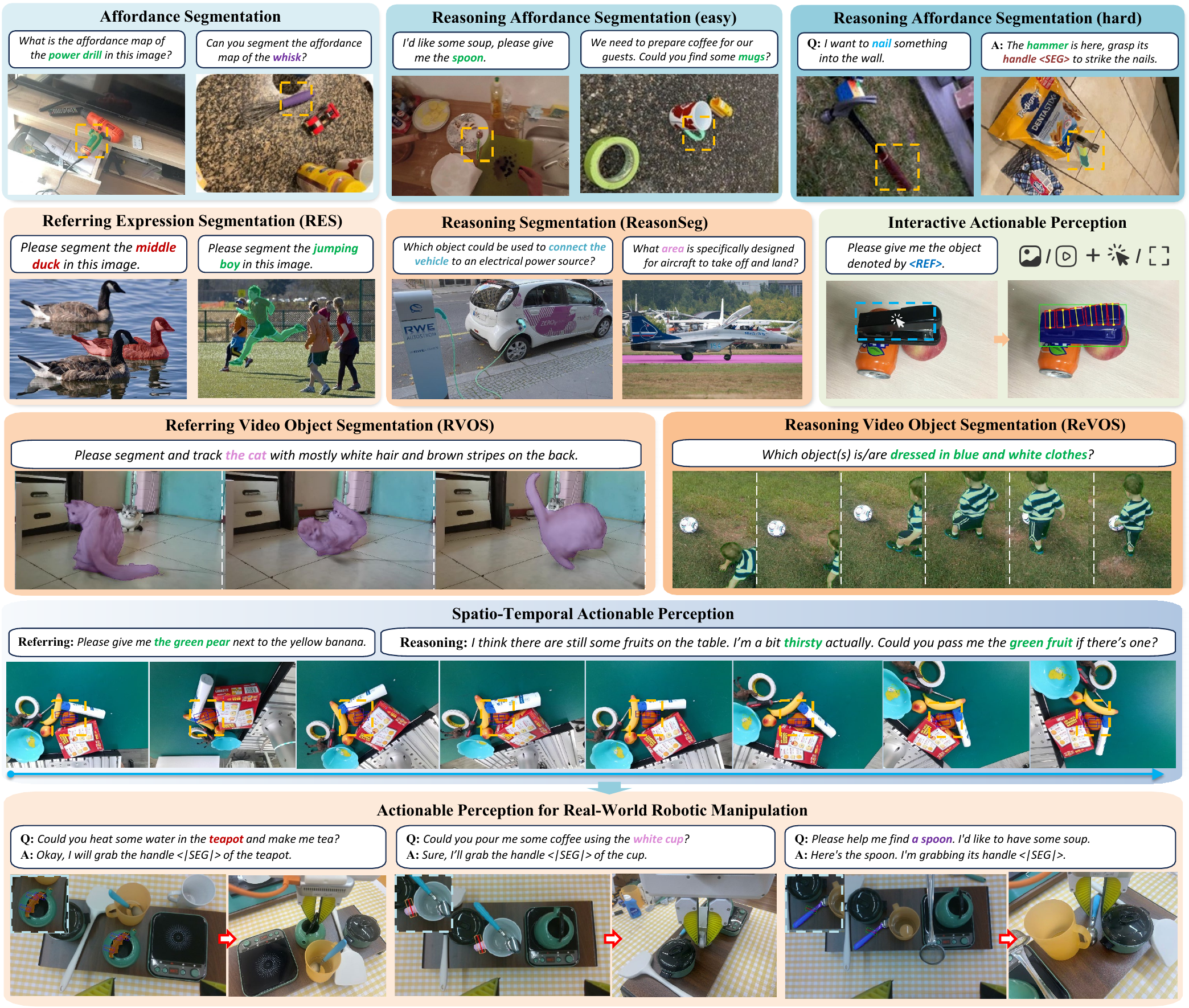}
\caption{\textbf{UniRAP} transfers large-scale generic segmentation pretraining to affordance segmentation and spatiotemporally actionable perception, improving language grounding for task-relevant regions and dynamics. It enhances referring segmentation, affordance grounding, and interaction modeling while maintaining strong spatio-temporal consistency.
}
\label{fig:task}
\vspace{-4mm}
\end{figure}

Vision-Language Perception (VLP) aims to establish semantic correspondences between natural language and visual observations, enabling models to interpret and localize relevant elements in a scene conditioned on linguistic instructions \cite{Vinyals2014ShowAT, Agrawal2015VQAVQ}. With the advancement of Vision-Language Models (VLMs), these systems have demonstrated strong capabilities in cross-modal alignment and semantic understanding in open-world settings, achieving notable success in tasks such as visual question answering \cite{Qwen-VL, Qwen3-VL, liu2023llava, Beyer2024PaliGemmaAV, chen2024internvl}. Building upon these foundations, subsequent works have extended such models to incorporate spatial perception, enabling them to generate coarse visual grounding from text, such as bounding boxes \cite{Qwen2.5-VL, Chen2023ShikraUM, cogvlm, You2023FerretRA, liu2025visualrft, jiang2025RexOmni} or segmentation masks \cite{lai2024lisa,liu2025unipixel, munasinghe2025videoglamm, cao2025refer, xia2024gsva, zhang2024psalm}. However, these capabilities remain largely confined to semantic understanding and alignment, lacking the fine-grained, actionable modeling required for physical interaction.

In embodied intelligence scenarios, this limitation becomes particularly critical: systems must not only understand \texttt{what} and \texttt{where}, but also further reason about \texttt{how to contact} \cite{krishna2017visualgenome, driess2023palme}. This requires models to possess \textbf{part-level actionable perception}, i.e., the ability to model interaction-relevant functional properties (affordances) \cite{do2018affordancenet, guo2023handal} and their corresponding operability, grounded in pixel-level spatial alignment and part-level structural understanding \cite{kirillov2023sam, he2022partimagenet, pan2023partseg, sun2023vlpart}. Without such capability, vision-language models struggle to translate language instructions into executable action representations, thereby limiting their applicability in real-world physical environments \cite{fei2025liberoplus, sapkota2025vlasurvey, zhou2025exploringvla}.

Despite notable progress, existing approaches still exhibit systematic limitations in advancing toward actionable perception. On one hand, cross-modal methods  \cite{vuong2024lgd, noh2025graspsam, shridhar2022cliport, sun2023vlpart, li2024ooal} based on CLIP/DINO-style visual representation alignment rely on static embedding spaces, which are inherently limited in modeling complex linguistic structures and compositional relationships. On the other hand, although VLM-based approaches demonstrate strong multi-step reasoning capabilities, their modeling granularity remains predominantly object-centric. Even when augmented with intermediate representations such as keypoints \cite{yuan25robopoint, zhou2025roborefer, nasiriany2024pivot, Yang2025magma,yuan2026embodiedr1,chen2025robovlm,li2024manipllm}, detection \cite{huang25a3vlm, jiang2025RexOmni, liu2025visualrft, huang2024manipvqa}, or segmentation \cite{qian2024affordancellm, xu2024naturalvlm, wu2025ragnet, lai2024lisa, liu2025unipixel}, these methods still fail to explicitly capture part-level functional affordances and contact regions, all of which are essential for reasoning about real-world interaction feasibility. Furthermore, RealVLG-R1 \cite{li2026realvlg} attempts to predict 4-DoF grasp poses via structured language generation. However, this approach relies on an autoregressive generation mechanism, which is prone to limited precision and instability, making it inadequate for pixel-level actionable tasks.

To address the aforementioned challenges, we propose \textbf{UniRAP}, a \textbf{Uni}fied vision-language model for \textbf{R}easoning-Aware \textbf{A}ctionable \textbf{P}erception. As shown in Fig. \ref{fig:task}, UniRAP supports both textual and visual prompts (e.g., points and bounding boxes) and is capable of producing multi-granularity affordance perception outputs over continuous data modalities, including images and videos. These outputs include object detection bounding boxes, part-level affordance segmentation masks, and interaction-oriented structured representations. Notably, UniRAP adopts a 4-DoF grasp pose as an explicit representation for actionable perception, which establishes a principled correspondence between high-level linguistic semantics and physical interaction.

To mitigate data imbalance and cross-task interference arising from heterogeneous perceptual granularity across tasks, a unified prompting and decoding paradigm is designed. Specifically, a structured \texttt{<REF>} token is introduced to integrate explicit visual prompts and implicit textual references into a shared \texttt{<SEG>} interface, thereby enabling consistent modeling of diverse localization signals. Furthermore, a Unified Affordance Decoder (UAD) is proposed to facilitate cross-task representation sharing through the reuse of intermediate features, while preserving general segmentation capability to the greatest extent possible. Finally, a curriculum-based transfer training strategy is adopted, allowing the model to progressively acquire stable and consistent actionable perception capabilities without compromising its original reasoning performance.

The main contributions of this work are summarized as follows:
\begin{enumerate}[left=1.5em,topsep=0pt,itemsep=2pt,parsep=2pt]

\item We propose UniRAP, a unified model that maps multimodal prompts to fine-grained physical contact geometry. It produces consistent multi-granularity affordance representations across both image and video domains.

\item A unified prompting and decoding pipeline is designed, together with a transfer-based training strategy, to alleviate task imbalance in cross-task learning and enable synergistic optimization across heterogeneous perception tasks.

\item Extensive experimental results demonstrate that UniRAP achieves a favorable trade-off between accuracy and stability across a range of tasks, including general segmentation, affordance perception, and part-level interaction grounding.
\end{enumerate}

%% file: sec/3_related_work.tex
\section{Related Work}

\textbf{Non-LLM Vision-Language Perception.}
Inspired by vision–language pretraining models such as CLIP \cite{radford2021clip, zhai2023siglip} and DINO \cite{caron202dino, oquab2023dinov2}, early VLP research primarily addresses coarse-grained tasks, including open-vocabulary object detection \cite{minderer2022owlvit,zhu2021deformable,ma2023codet,kuo2023openvocabulary,du2022learning,wu2023aligning,liu2024groundingdino,zhong2022regionclip, wang2025yoloe, cheng2024yoloworld, sapkota2025yolo26, carion2020detr, robinson2025rfdetr} and semantic segmentation \cite{shin2024PixelCLIP, pan2023partseg, yu2023fcclip, dong2023maskclip, wei2023ovparts,liang2023ovseg,cvpr2025dualseg,cvpr2025dpseg}. These methods rely on region–text alignment to associate visual and linguistic representations, enabling generalization beyond fixed category spaces. Subsequent studies extend VLP to affordance-aware perception, shifting from object recognition to modeling functional properties, typically formulated as part-level segmentation \cite{do2018affordancenet,Chuang2017LearningTA,jiang2025affordancesam, xu2025weaklysupervised,tang2025closed,moon2025selective,cvpr2025reasonmamba,li2023locate,tong2025oval} or region-level localization \cite{iros2017ITAFF,ijcv2022osoad,iccv2023cotdet,Wang_2024_CVPR}. More recently, this paradigm further expands to actionable perception, such as contact point prediction \cite{huang2024rekep,ju2024roboabc,xue2022useek,jiang2024doduo} and 4-DoF grasp pose estimation \cite{vuong2024lgd, noh2025graspsam, iroslightlgd, iros2025graspmamba, iros2024masklgd}, aiming to bridge the gap between visual perception and physical manipulation. However, due to limited textual understanding in pretrained models, these methods remain restricted to simple category-level language and struggle with compositional and reasoning-intensive instructions, which limits their applicability in open-world embodied scenarios.

\textbf{LLM-based Vision-Language Perception.}
In recent years, large-scale VLMs have demonstrated strong semantic understanding capabilities in tasks such as visual question answering \cite{Qwen-VL,liu2023llava,chen2024internvl,Qwen2-VL,Qwen2.5-VL,Qwen3-VL,liu2024improvedllava}. However, due to their reliance on global or region-level representations, they remain limited in pixel-level visual perception. To address this limitation, some approaches \cite{ma2024groma,zhang2024ferretv2,liu2025visualrft,sarch2025regrounded,jiang2025RexOmni,Qwen2.5-VL} enforce the generation of structured language through SFT \cite{nips2020gpt3,nips2022rlhf} or RLVR \cite{shao2024grpo, Zheng2025gspo}, enabling coarse-grained object localization. For dense segmentation tasks, VisionLLM \cite{nips2023visionllm} adopts a text-based polygon output paradigm, which is highly dependent on large-scale training data and still suffers from limited accuracy. To overcome these issues, LISA \cite{lai2024lisa} introduces the ``Embedding as Mask'' paradigm, which extracts implicit features from the language model via a special token and injects them into SAM \cite{kirillov2023sam}, enabling stable pixel-level predictions. Building on this line, subsequent studies further advance decoder design \cite{ren2024pixellm, nextchat, zhang2024groundhog, rasheed2024glamm}, scale up training data \cite{jang2025m2sa}, and extend capabilities to video settings \cite{bai2024videolisa, yan2024visa, yuan2025sa2va,munasinghe2025videoglamm}. UniPixel \cite{liu2025unipixel} further proposes the PixelQA task, unifying segmentation and region-referred visual question answering. In the context of affordance perception, existing methods largely remain limited to embodied question answering \cite{ji2025robobrain, huang2024manipvqa,kim2025manipgpt,chen2025robovlm} or keypoint prediction \cite{zhou2025roborefer,xu2025a0,ma2024glover,yuan25robopoint, nasiriany2024pivot, Yang2025magma,yuan2026embodiedr1,li2024manipllm}. To date, RAGNet \cite{wu2025ragnet} is the only work that extends LISA \cite{lai2024lisa} to affordance-level segmentation. However, it still lacks interaction-centric modeling and the capability to process video data. The most closely related work, RealVLG-R1 \cite{li2026realvlg}, employs reinforcement fine-tuning of VLMs to generate structured text for predicting bounding boxes and grasp poses, but is limited by unstable text representations, weak reasoning ability, and lack of video support. In contrast, UniRAP is the first to achieve end-to-end, reasoning-aware unified modeling of actionable affordance perception across both images and videos.

%% file: sec/4_method.tex
\section{UniRAP}
\label{sec:method}

\begin{figure}
\centering
\includegraphics[width=\linewidth]{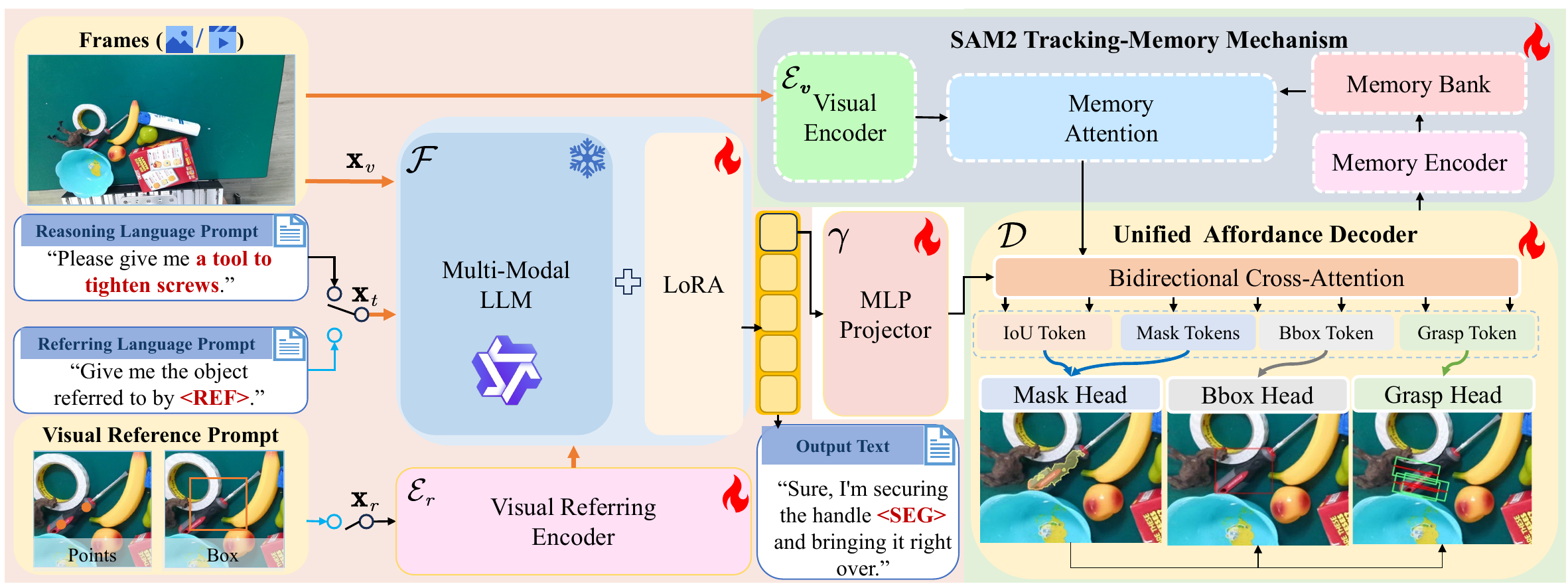}
\caption{\textbf{Overview of UniRAP Model.}  Given visual input $\mathbf{x}_v$, text prompt $\mathbf{x}_t$, and optional visual reference prompts $\mathbf{x}_r$, a multimodal model $\mathcal{F}$ produces \texttt{<SEG>}-conditioned outputs. When \texttt{<REF>} is present, visual prompts are encoded as embeddings and jointly modeled with text within a shared \texttt{<SEG>} representation space. Concurrently, visual features from $\mathbf{x}_v$ are aggregated with memory bank representations via memory attention. The resulting features are aligned with \texttt{<SEG>}-derived embeddings through an MLP projector and decoded by an affordance decoder $\mathcal{D}$ to generate segmentation masks, region detections, and 4-DoF interaction poses for unified vision–language spatiotemporal actionable perception.
}
\label{fig:framework}
\vspace{-4mm}
\end{figure}

\textbf{Overview.}
To enable a unified model for vision–language spatiotemporal actionable perception, UniRAP is built upon Qwen2.5-VL \cite{Qwen2.5-VL} and SAM2 \cite{ravi2024sam2}. As illustrated in Fig. \ref{fig:framework}, given a visual input (image or video) $\mathbf{x}_v$, a textual prompt $\mathbf{x}_t$, and optional visual reference prompts $\mathbf{x}_r$ (e.g., points or bounding boxes), UniRAP first leverages a multimodal large model $\mathcal{F}$ to generate textual responses containing \texttt{<SEG>} tokens. When the textual prompt $\mathbf{x}_t$ includes a \texttt{<REF>} token, the visual prompts are encoded into discrete embeddings via a visual referring encoder and injected into the textual sequence, which is then jointly processed by the multimodal model $\mathcal{F}$. These two forms of prompting are unified within a shared representation space centered on \texttt{<SEG>} tokens, facilitating transferability and generalization to tasks without explicit language referring cues. In the visual branch, the input $\mathbf{x}_v$ is encoded by a visual encoder $\mathcal{E}_v$, and the resulting features are fused with historical frame representations stored in a memory bank via a memory attention mechanism, enhancing temporal consistency and object persistence modeling. The fused features are then aligned with implicit prompt embeddings projected from \texttt{<SEG>} tokens through a MLP projector $\gamma$, and jointly fed into a unified affordance decoder $\mathcal{D}$. The model yields multi-task predictions, including affordance segmentation masks, region detection results, and interaction poses (4-DoF grasp poses), thereby enabling fine-grained actionable perception in complex spatiotemporal scenarios.

\subsection{Unified Vision-Language Spatiotemporal Modeling}
To address vision–language spatiotemporal affordance perception under low-data conditions, UniRAP extends the ``Embedding as Mask'' paradigm \cite{lai2024lisa,ren2024pixellm,bai2024videolisa,liu2025unipixel} by reformulating the problem as a conditional multimodal generation framework based on interface tokens. Specifically, the model takes as input visual observations (images or videos) $\mathbf{x}_v$, textual prompts $\mathbf{x}_t$, and optionally visual reference prompts $\mathbf{x}_r$ (e.g., points or bounding boxes). Through a unified interface token mechanism, it generates reasoning-aware textual outputs, thereby enabling spatial alignment and joint modeling of textual and visual referents. This design maps supervisory signals from heterogeneous modalities into a shared representation space, which not only facilitates cross-task co-optimization but also enhances the model’s transferability and generalization performance in multi-task settings.

Specifically, to enable effective transfer learning in scenarios without explicit textual referring expressions, UniRAP introduces a conditional visual referring prompt mechanism. When the input text $\mathbf{x}_t$ contains the \texttt{<REF>} token, the visual prompt $\mathbf{x}_r$ is encoded by a visual referring encoder $\mathcal{E}_r$ into a discrete embedding, which is then injected into the multimodal model $\mathcal{F}$ (Qwen2.5-VL \cite{Qwen2.5-VL}) for joint modeling. The model subsequently generates a reasoning-aware textual output $\hat{\mathbf{y}}$ that includes the \texttt{<SEG>} token. This process can be formally expressed as:
\begin{equation}
\hat{\mathbf{y}} = \mathcal{F}\big(\mathbf{x}_t, \mathbf{x}_v, \mathbf{z}_r\big), 
\quad 
\mathbf{z}_r =
\begin{cases}
\mathcal{E}_r(\mathbf{x}_r), & \text{if } \texttt{<REF>} \in \mathbf{x}_t \\
\mathbf{0}, & \text{otherwise},
\end{cases}
\end{equation}
where the referring expression encoder $\mathcal{E}_r$ adopts random Fourier mapping \cite{tancik2020fouriermapping} to encode the spatial coordinates and temporal information of each point, followed by a linear projection layer. Unlike prior methods \cite{yuan2025sa2va, ren2024pixellm, liu2025unipixel} designed for visual prompt-based question answering, we adapt this design to unify and align explicit textual referring and visual prompt-based referring within the \texttt{<SEG>} representation space, thereby enabling task transfer with reduced reliance on textual annotations.

Subsequently, to achieve stable and lightweight transfer-based spatiotemporal actionable perception, UniRAP extracts the final-layer hidden representation $\tilde{\mathbf{h}}_{\text{seg}}$ corresponding to the \texttt{<SEG>} token and projects it via a MLP projector $\gamma(\cdot)$ into a unified object-localization prompt embedding:
\begin{equation}
    \mathbf{h}_{\text{seg}} = \gamma(\tilde{\mathbf{h}}_{\text{seg}}).
\end{equation}
In parallel, SAM2 \cite{ravi2024sam2} encodes the visual input $\mathbf{x}_v$ into dense features $\mathbf{f}_v$ through its visual encoder $\mathcal{E}_v$. Owing to the tracking-memory mechanism, when processing video sequences, these visual features interact with historical frame features and target states stored in a memory bank, and temporal information is aggregated via memory attention, thereby maintaining temporal consistency across frames. Finally, the memory-enhanced visual features $\mathbf{f}_{\text{mem}}$ and the implicit prompt embedding $\mathbf{h}_{\text{seg}}$ are fused within the decoder $\mathcal{D}$ to jointly predict affordance-related outputs:
\begin{equation}
    {\hat{\mathbf{M}}, \hat{\mathbf{b}}, \hat{\mathbf{g}}}
= \mathcal{D}(\mathbf{h}_{\text{seg}}, \mathbf{f}_{\text{mem}}),
\end{equation}
where $\hat{\mathbf{M}}$ denotes the affordance segmentation mask, $\hat{\mathbf{b}}$ denotes region-level detection results, and $\hat{\mathbf{g}}$ denotes the 4-DoF interaction poses.

\subsection{Unified Affordance Decoder}
Given that vision–language segmentation data are relatively abundant, whereas annotations for fine-grained interaction perception (e.g., 4-DoF grasp poses) are scarce, we propose a Unified Affordance Decoder (UAD). The goal is to effectively transfer intermediate representations (visual features and implicit prompt embeddings) learned from large-scale coarse-grained perception data to fine-grained tasks, thereby improving affordance perception performance.

\begin{wrapfigure}{R}{0.7\textwidth}
\includegraphics[width=\linewidth]{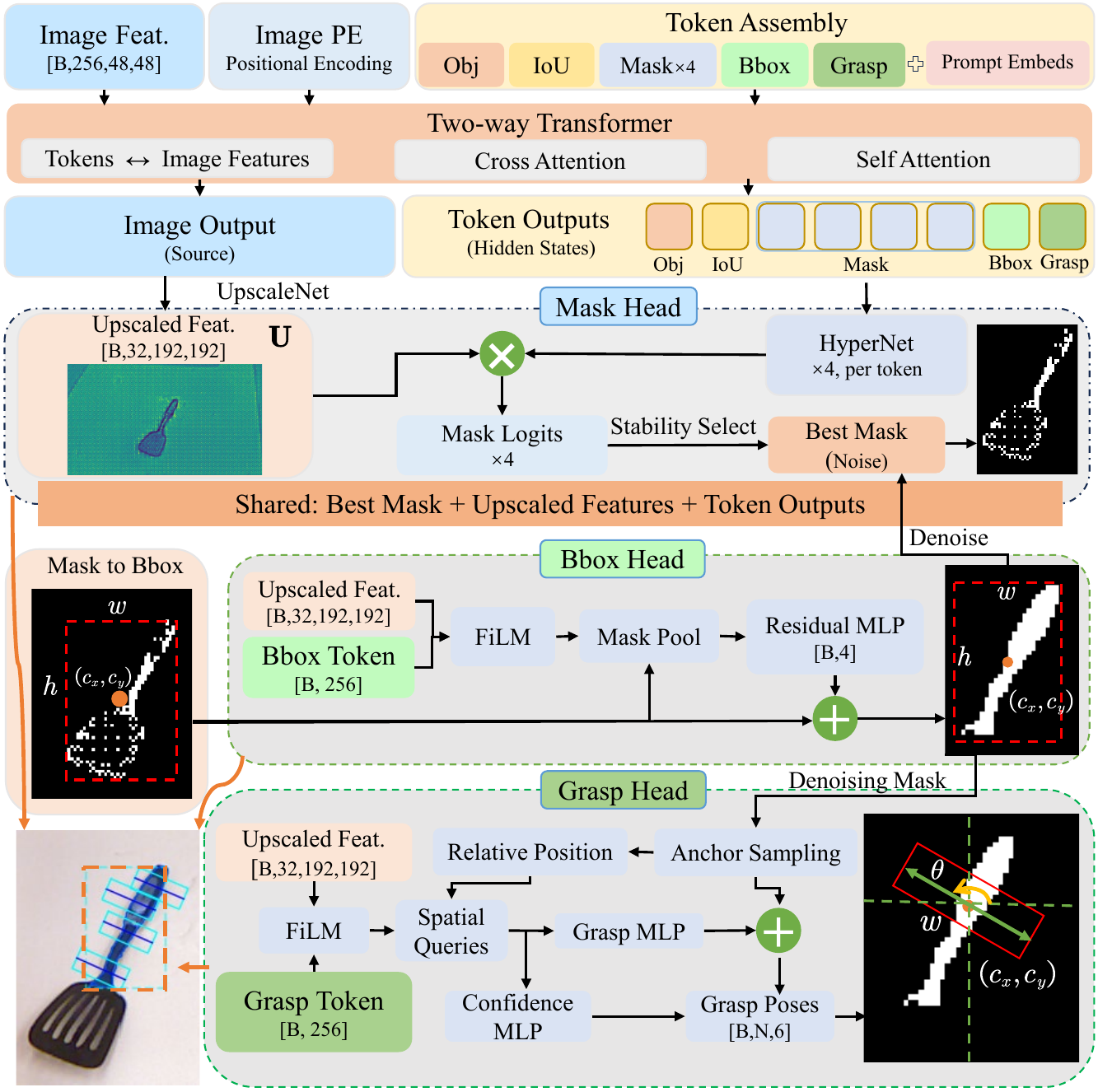}
\caption{\textbf{Architecture of Unified Affordance Decoder.}}
\label{fig:uad}
\vspace{-4mm}
\end{wrapfigure}

Specifically, as illustrated in Fig. \ref{fig:uad}, the UAD is built upon the Mask Decoder of SAM \cite{ravi2024sam2, kirillov2023sam}. During the decoding stage, a Two-way Transformer \cite{carion2020detr} takes as input the memory-enhanced visual features $\mathbf{f}_{\text{mem}}$, positional encodings $\mathbf{p}$, and task tokens augmented with object-localization prompt embeddings $\mathbf{h}_{\text{seg}}$, including the object pointer token, IoU token, mask tokens $\mathbf{t}_m$, bbox token $\mathbf{t}_b$, and grasp token $\mathbf{t}_g$. Through joint modeling via bidirectional cross-attention and self-attention, the model enables coordinated updates between visual features and token representations, resulting in context-enhanced representations.

To enable unified decoding across different granularities, UAD adopts a token-based feature modulation mechanism for prediction. In the Mask Head, the updated visual features are first upsampled via UpscaleNet to obtain a high-resolution representation $\mathbf{f}_{\text{up}}$. The mask tokens $\mathbf{t}_m$, projected through a HyperNet, are then used to perform element-wise modulation of $\mathbf{f}_{\text{up}}$, producing four candidate mask logits. A stability-based selection strategy is further applied to iteratively refine these candidates, resulting in the final optimal segmentation mask $\hat{\mathbf{M}}$.

For the object detection task, the Bbox Head adopts a DETR-style residual regression modeling strategy \cite{carion2020detr}. An initial bounding box $\mathbf{b}_0$ is first derived from the optimal segmentation mask $\hat{\mathbf{M}}$ via geometric construction. The bbox token $\mathbf{t}_b$ then modulates the upsampled features $\mathbf{f}_{\text{up}}$ via FiLM \cite{perez2018film}, followed by mask-guided weighted pooling under $\hat{\mathbf{M}}$. The resulting feature is fed into a ResidualMLP to predict a bounding box offset for refinement. This process can be formulated as:
\begin{equation}
    \hat{\mathbf{b}}
= \mathbf{b}_0
+ \mathrm{MLP}_{\text{res}}(
  \mathrm{Pool}_{\hat{\mathbf{M}}}(
  \mathrm{FiLM}(\mathbf{f}_{\text{up}}, \mathbf{t}_b))).
\end{equation}
To improve robustness, UniRAP further introduces a consistency constraint between the predicted bounding box and the segmentation mask, enabling mutual refinement and suppressing mask noise.

For contact-aware affordance perception, conventional heatmap-based methods \cite{grcnn, morrison2018ggcnn} exhibit relatively stable predictions but suffer from ground-truth coverage and uneven response distributions. To address these limitations, we reformulate 4-DoF grasp prediction within a regression-based framework, focusing on part-level contact region estimation in the object-centric spatial domain.

Specifically, given a refined segmentation mask $\hat{\mathbf{M}}$, the Grasp Head first applies farthest point sampling (FPS) within the foreground region to select $N_g$ grasp center candidates $(a_x^i, a_y^i)$. Each anchor is encoded as a normalized offset relative to the mask centroid and further scaled by the bounding box dimensions, forming a relative positional encoding $\mathbf{e}_i$ to enhance scale-consistent spatial representation. Next, the upsampled visual feature $\mathbf{f}_\text{up}$ is modulated by the grasp token via FiLM conditioning to inject task-specific semantic information. Based on this, $N_g$ learnable spatial queries interact with the modulated features through foreground-masked dot-product attention, producing per-candidate pooled feature representations $\mathbf{z}_i$:
\begin{equation}
\mathbf{z}_i
= \mathrm{Attn}\big(
\mathrm{FiLM}(\mathbf{f}_\text{up}, \mathbf{t}_g),\;
\mathrm{FPS}(\hat{\mathbf{M}}),\;
\mathbf{e}_i
\big).
\end{equation}
Subsequently, a GraspMLP predicts residual offsets and grasp geometry, while an independent ConfidenceMLP estimates the corresponding grasp quality score $q_i$:
\begin{equation}
(\Delta c_x^i, \Delta c_y^i, \cos 2\theta_i, \sin 2\theta_i, w_i)
= \mathrm{MLP}_{\text{grasp}}(\mathbf{z}_i), \quad
q_i = \mathrm{MLP}_{\text{conf}}(\mathbf{z}_i).
\end{equation}
The final grasp center is obtained via anchor-residual refinement: $c_x^i = a_x^i + \Delta c_x^i, c_y^i = a_y^i + \Delta c_y^i$. We thus obtain $N_g$ grasp candidates, denoted as (see Fig. \ref{fig:uad} for illustration):
\begin{equation}
\hat{\mathbf{g}} = \{(c_x^i, c_y^i, \cos 2\theta_i, \sin 2\theta_i, w_i, q_i)\}_{i=1}^{N_g}.
\end{equation}

Overall, UAD provides a unified decoding paradigm to integrate segmentation, detection, and part-level contact reasoning into a single representation space. By leveraging a shared interface, it enables synergistic knowledge transfer across heterogeneous tasks, ensuring that interaction-centric geometry is grounded in robust semantic features.

\subsection{Curriculum Transfer Training}
\begin{figure}
\centering
\includegraphics[width=\linewidth]{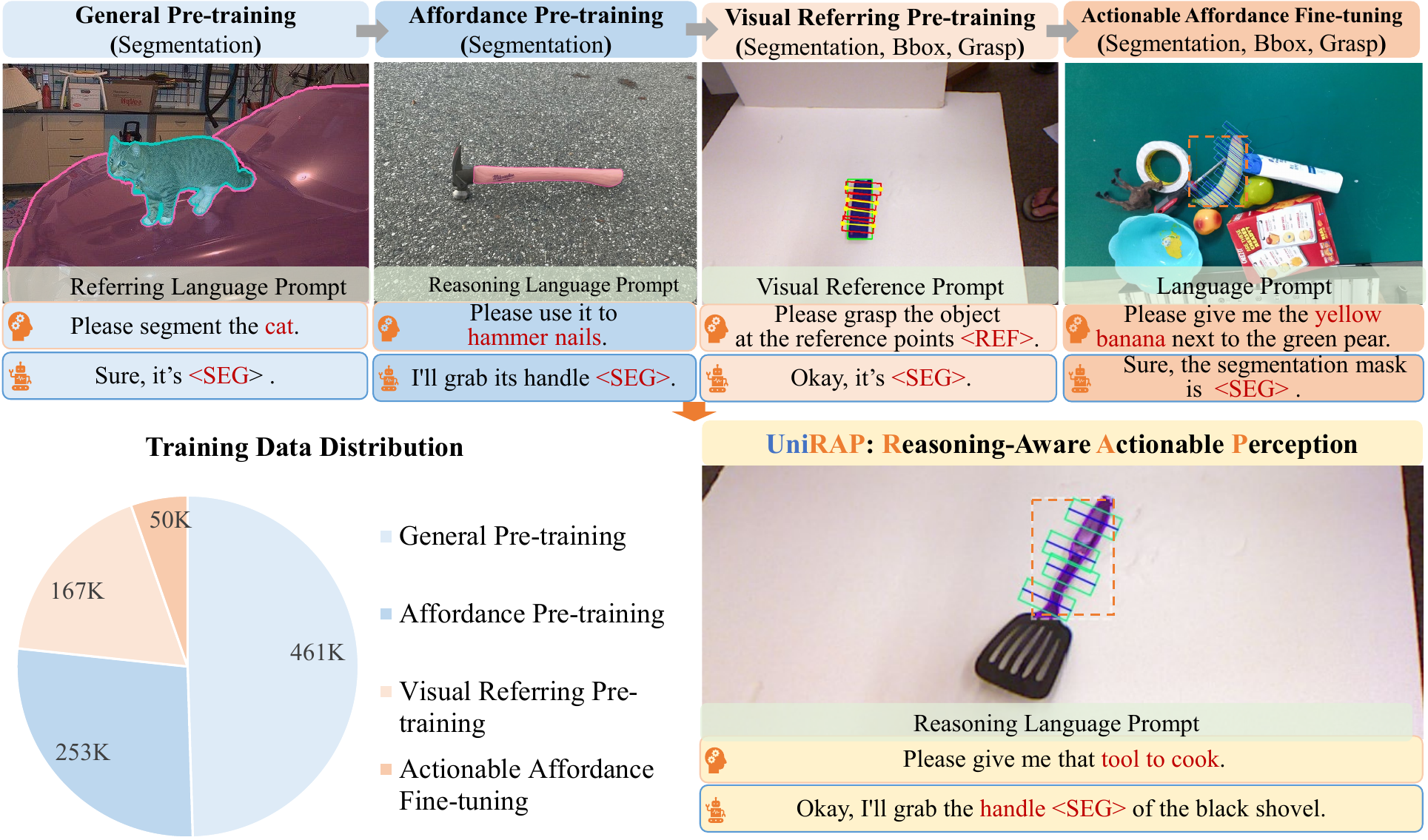}
\caption{\textbf{Curriculum Transfer Training Pipeline.} Details in Appendix \ref{appendix:train_recipe}.}
\label{fig:transfer_training}
\vspace{-4mm}
\end{figure}

\textbf{Training Pipeline.}
To address the imbalance between large-scale segmentation data and limited interaction annotations, we adopt a curriculum-based transfer learning framework that progressively transitions model capabilities from segmentation to interaction-aware perception. As illustrated in Fig. \ref{fig:transfer_training}, the training pipeline is organized into four stages with increasing task complexity: (1) general pre-training with language prompts to enhance visual–semantic alignment and mask prediction; (2) affordance pre-training to localize functionally meaningful regions, bridging category-level segmentation and part-level perception; (3) visual referring pre-training with visual prompts to implicitly learn interaction-relevant poses from grasp datasets without language annotations; and (4) actionable affordance fine-tuning using limited vision-language grasp data to jointly predict segmentation masks, bounding boxes, and 4-DoF interaction poses. This staged strategy enables effective knowledge transfer, thereby improving data efficiency and generalization performance.

\textbf{Training Objective.}
The training objective of UniRAP is formulated as a weighted linear combination of a language modeling loss and multiple task-specific losses. The mask decoding component includes a Focal Loss and a Dice Loss for mask prediction, a MAE loss for IoU estimation, and a cross-entropy loss for objectness classification. The object detection branch employs an L1 regression loss and a GIoU loss to supervise bounding box prediction, while the grasp pose estimation branch is optimized using a Huber Loss and a BCE Loss. Furthermore, to enforce spatial consistency across multi-granularity perception, additional constraints are introduced, including a mask-in-bbox loss to ensure containment between masks and bounding boxes, and a consistency loss that aligns grasp center points with the corresponding mask regions. Details are provided in Appendix \ref{appendix_loss}.

%% file: sec/5_experiments.tex
\section{ Experiments}
This section demonstrates that UniRAP preserves strong general segmentation performance while extending to affordance segmentation and actionable perception, enabling a more grounded understanding of functional regions and physical interactions. \textbf{Details on implementation (\ref{appendix_iImplementation}), evaluation metrics (\ref{appendix_metric}), qualitative results  (\ref{appendix_results}), and real-world robotic experiments (\ref{appendix_realworld_grasping}) are provided in the Appendix.}

\subsection{General Segmentation Performance}
\textbf{Referring Expression and Reasoning Segmentation.}
As shown in Tab. \ref{tab:res}, we evaluated UniRAP on RefCOCO/+/g \cite{refcoco,refcocog} and ReasonSeg \cite{lai2024lisa} to assess its image segmentation capability under both explicit and implicit textual queries. The results demonstrate that UniRAP achieves state-of-the-art performance on referring expression segmentation (RES) benchmarks. Among models of comparable scale, UniRAP significantly outperforms existing methods and, in most settings, attains performance on par with larger models such as UniPixel-7B \cite{liu2025unipixel} and Sa2VA-8B \cite{yuan2025sa2va}, indicating a favorable trade-off between parameter efficiency and performance. On ReasonSeg \cite{lai2024lisa}, UniRAP remains competitive but does not achieve the best performance. This is primarily because ReasonSeg requires stronger multi-step reasoning capabilities, while the current state-of-the-art method, VideoLISA \cite{bai2024videolisa}, is specifically designed and trained for such reasoning-intensive segmentation tasks.

\begin{table}[t]
\centering
\caption{\textbf{Referring Expression and Reasoning Segmentation.} cIoU on RefCOCO/+/g.}
\vspace{2mm}
\resizebox{0.9\linewidth}{!}{
\begin{tabular}{l|c|ccc|ccc|cc|cc}
\toprule
\multirow{2}{*}{Method} & \multirow{2}{*}{Size} & \multicolumn{3}{c|}{RefCOCO \cite{refcoco}} & \multicolumn{3}{c|}{RefCOCO+ \cite{refcoco}} & \multicolumn{2}{c|}{RefCOCOg \cite{refcocog}} & \multicolumn{2}{c}{ReasonSeg \cite{lai2024lisa}} \\

\cmidrule{3-12}
&& val & testA & testB & val & testA & testB & val(U) & test(U) & gIoU & cIoU \\

\midrule
\rowcolor{gray!10}\multicolumn{12}{l}{\textit{\textcolor{gray}{Non-LLM-based Specialists}}\vspace{0.5mm}} \\
ReLA \cite{liu2023gres} & -- & 73.8 & 76.5 & 70.2 & 66.0 & 71.0 & 57.7 & 65.0 & 66.0 & -- & -- \\
X-Decoder \cite{zou2023xdecoder} & -- & -- & -- & -- & -- & -- & -- & 64.6 & -- & 22.6 & 17.9 \\
SEEM \cite{zou2023seem} & -- & -- & -- & -- & -- & -- & -- & 65.7 & -- & 25.5 & 21.2 \\

\midrule
\rowcolor{gray!10}\multicolumn{12}{l}{\textit{\textcolor{gray}{LLM-based Image Generalists}}\vspace{0.5mm}} \\
NExT-Chat \cite{nextchat} & 7B & 74.7 & 78.9 & 69.5 & 65.1 & 71.9 & 56.7 & 67.0 & 67.0 & -- & -- \\
PixelLM \cite{ren2024pixellm} & 7B & 73.0 & 76.5 & 68.2 & 66.3 & 71.7 & 58.3 & 69.3 & 70.5 & -- & -- \\
LISA \cite{lai2024lisa} & 7B & 74.9 & 79.1 & 72.3 & 65.1 & 70.8 & 58.1 & 67.9 & 70.6 & 61.3 & 62.9 \\
Groundhog \cite{zhang2024groundhog} & 7B & 78.5 & 79.9 & 75.7 & 70.5 & 75.0 & 64.9 & 74.1 & 74.6 & 56.2 & -- \\
LaSagnA \cite{wei2024lasagna} & 7B & 76.8 & 78.7 & 73.8 & 66.4 & 70.6 & 60.1 & 70.6 & 71.9 & 48.8 & 47.2 \\
M$^{\rm 2}$SA \cite{jang2025m2sa} & 13B & 74.6 & 77.6 & 71.0 & 64.0 & 68.1 & 57.6 & 69.0 & 69.3 & -- & -- \\

\midrule
\rowcolor{gray!10}\multicolumn{12}{l}{\textit{\textcolor{gray}{LLM-based Video Generalists}}\vspace{0.5mm}} \\
VISA \cite{yan2024visa} & 7B & 72.4 & 75.5 & 68.1 & 59.8 & 64.8 & 53.1 & 65.5 & 66.4 & 52.7 & 57.8 \\
Vitron \cite{fei2024vitron} & 7B & 75.5 & 79.5 & 72.2 & 66.7 & 72.5 & 58.0 & 67.9 & 68.9 & -- & -- \\
UniPixel \cite{liu2025unipixel} & 7B & 80.8 & 83.0 & 77.4 & 75.3 & 80.1 & 70.0 & 76.4 & 77.1 & 60.5 & 58.7 \\
Sa2VA \cite{yuan2025sa2va} & 8B & \textbf{81.9} & -- & -- & \textbf{76.5} & -- & -- & \textbf{78.9} & -- & 57.9 & 44.7 \\

\midrule
VideoLISA \cite{bai2024videolisa} & 3.8B & 73.8 & 76.6 & 68.8 & 63.4 & 68.8 & 56.2 & 68.3 & 68.8 & \textbf{61.4} & \textbf{67.1} \\
Sa2VA \cite{yuan2025sa2va} & 4B & 78.9 & -- & -- & 71.7 & -- & -- & 74.1 & -- & 52.3 & 44.9 \\
\rowcolor{orange!10} \textbf{UniRAP} (Ours) & 3B & \textbf{80.9} & \textbf{82.6} & \textbf{78.1} & \textbf{75.1} & \textbf{79.3} & \textbf{70.4} & \textbf{76.1} & \textbf{77.5} & 57.9 & 59.0 \\

\bottomrule
\end{tabular}}
\label{tab:res}
\vspace{-6mm}
\end{table}

\textbf{Referring and Reasoning Video Object Segmentation.}
To evaluate UniRAP’s ability in mask prediction under implicit textual queries and spatiotemporal reasoning, we conducted experiments on ReVOS \cite{yan2024visa}. As shown in Tab. \ref{tab:revos}, UniRAP achieves strong overall performance on ReVOS across both referring and reasoning settings, outperforming representative 3B-scale baselines such as HyperSeg \cite{wei2024hyperseg} and InstructSeg \cite{wei2024instructseg}. It remains competitive compared to recent larger-scale models (e.g., UniPixel-7B), demonstrating a favorable efficiency–performance trade-off.

\begin{table}[t]
\scriptsize
\centering
\caption{\textbf{Referring and Reasoning Video Object Segmentation} on ReVOS  \texttt{val} \cite{yan2024visa} split.}
\vspace{2mm}
\resizebox{0.9\linewidth}{!}{
\begin{tabular}{l|c|ccc|ccc|ccc|c}
\toprule
\multirow{2.6}{*}{\textbf{Method}} & \multirow{2.6}{*}{\textbf{Size}} & \multicolumn{3}{c|}{\textbf{Referring}} & \multicolumn{3}{c|}{\textbf{Reasoning}} & \multicolumn{3}{c|}{\textbf{Overall}} & \multirow{2.6}{*}{$\mathcal{R}$} \\
\cmidrule{3-5} \cmidrule{6-8} \cmidrule{9-11}
&& $\mathcal{J}$ & $\mathcal{F}$ & $\mathcal{J\&F}$ & $\mathcal{J}$ & $\mathcal{F}$ & $\mathcal{J\&F}$ & $\mathcal{J}$ & $\mathcal{F}$ & $\mathcal{J\&F}$ \\
\midrule
\rowcolor{gray!10}\multicolumn{12}{l}{\textit{\textcolor{gray}{Non-LLM-based Specialists}}\vspace{0.5mm}} \\
MTTR \cite{mttr} & -- & 29.8 & 30.2 & 30.0 & 20.4 & 21.5 & 21.0 & 25.1 & 25.9 & 25.5 & 5.6 \\
LMPM \cite{ding2023mevis} & -- & 29.0 & 39.1 & 34.1 & 13.3 & 24.3 & 18.8 & 21.2 & 31.7 & 26.4 & 3.2 \\
ReferFormer \cite{wu2022referformer} & -- & 31.2 & 34.3 & 32.7 & 21.3 & 25.6 & 23.4 & 26.2 & 29.9 & 28.1 & 8.8 \\
\midrule
\rowcolor{gray!10}\multicolumn{12}{l}{\textit{\textcolor{gray}{LLM-based Generalists}}\vspace{0.5mm}} \\
LISA \cite{lai2024lisa} & 13B & 45.2 & 47.9 & 46.6 & 34.3 & 39.1 & 36.7 & 39.8 & 43.5 & 41.6 & 8.6 \\
VISA \cite{yan2024visa} & 13B & 55.6 & 59.1 & 57.4 & 42.0 & 46.7 & 44.3 & 48.8 & 52.9 & 50.9 & 14.5 \\

GLUS \cite{lin2025glus} & 7B & 56.0 & 60.7 & 58.3 & 48.8 & 53.9 & 51.4 & 52.4 & 57.3 & 54.9 & 17.9 \\
ViLLa \cite{zheng2025villa} & 6B & -- & -- & -- & -- & -- & -- & 54.9 & 59.1 & 57.0 & -- \\
UniPixel \cite{liu2025unipixel}& 7B & \textbf{63.9} & \textbf{67.8} & \textbf{65.8} & \textbf{59.4} & \textbf{63.7} & \textbf{61.5} & \textbf{61.7} & \textbf{65.7} & \textbf{63.7} & \textbf{19.4} \\
Sa2VA \cite{yuan2025sa2va} & 8B & -- & -- & -- & -- & -- & -- & -- & -- & 57.6 & -- \\

\midrule
HyperSeg \cite{wei2024hyperseg} & 3B & 56.0 & 60.9 & 58.5 & 50.2 & 55.8 & 53.0 & 53.1 & 58.4 & 55.7 & -- \\
InstructSeg \cite{wei2024instructseg} & 3B & 54.8 & 59.2 & 57.0 & 49.2 & 54.7 & 51.9 & 52.0 & 56.9 & 54.5 & -- \\
Sa2VA \cite{yuan2025sa2va} & 4B & -- & -- & -- & -- & -- & -- & -- & -- & 53.2 & -- \\
\rowcolor{orange!10} \textbf{UniRAP} (Ours) & 3B & \textbf{61.8} & \textbf{66.5} & \textbf{64.1} & \textbf{56.8} & \textbf{62.4} & \textbf{59.6} & \textbf{59.3} & \textbf{64.4} & \textbf{61.9} & \textbf{19.4} \\
\bottomrule
\end{tabular}}
\label{tab:revos}
\vspace{-2mm}
\end{table}

\begin{table}[t!]
\centering
\begin{minipage}{0.49\textwidth}
\centering
\caption{\textbf{Referring Affordance Segmentation.}}
\label{tbl:affordance_segmentation}
\scriptsize
\resizebox{\linewidth}{!}{
        \begin{tabular}{l|c|cc|cc|cc|cc}
            \toprule
            \multicolumn{1}{l|}{\multirow{2}*{Method}} &\multicolumn{1}{l|}{\multirow{2}*{Size}} & \multicolumn{2}{c|}{HANDAL} & \multicolumn{2}{c|}{GraspNet \texttt{seen}} & \multicolumn{2}{c|}{\cellcolor{gray!30} GraspNet \texttt{novel}} & \multicolumn{2}{c}{\cellcolor{gray!30}3DOI} \\
            & & gIoU & cIoU & gIoU & cIoU & \cellcolor{gray!30}gIoU & \cellcolor{gray!30}cIoU & \cellcolor{gray!30}gIoU & \cellcolor{gray!30}cIoU  \\
            \midrule
            
            \rowcolor{gray!10}\multicolumn{10}{l}{\textit{\textcolor{gray}{Non-LLM Foundation Models}}\vspace{0.5mm}} \\
            \midrule
            VLPart~\cite{sun2023vlpart} + SAM2~\cite{ravi2024sam2} & - & 40.9 & 28.9 & 40.7& 27.6 & - & - & - & - \\
            G-DINO~\cite{liu2024groundingdino} + SAM2~\cite{ravi2024sam2} & - & 34.7 & 26.8 & 34.9 & 26.9 & - & - & - & -  \\
            Florence 2~\cite{xiao2024florence2} + SAM2~\cite{ravi2024sam2} & - & 39.7 & 22.4 & 39.4 & 22.5& - & - & - & -   \\
            \midrule
            \rowcolor{gray!10}\multicolumn{10}{l}{\textit{\textcolor{gray}{LLM-based Generalists}}\vspace{0.5mm}} \\
            \midrule
            LISA~\cite{lai2024lisa} & 7B & 15.4 & 11.8 & 17.7 & 17.7 & 25.2 & 24.1 & 21.5 & 13.7  \\
            GLaMM~\cite{rasheed2024glamm} & 3.8B & 25.1 & 17.0 & 21.6 & 10.5 & 19.2 & 8.6 & 19.7 & 14.1  \\
        
            UniPixel \cite{liu2025unipixel} & 7B & 26.0 & 16.4 & 45.0 &12.5 & \underline{48.0} & 11.5& \underline{41.5} & 24.9  \\
            
            RAGNet \cite{wu2025ragnet} & 7B & \underline{60.5} & \underline{60.3} & \underline{63.3} &\underline{64.0} & 45.6 &\underline{33.2} & 37.4 & \underline{37.4} \\
        
            \midrule
            \rowcolor{orange!10} \textbf{UniRAP} (Ours) & 3B &  \textbf{63.9} & \textbf{65.7} & \textbf{70.6} &\textbf{64.1} & \textbf{51.6} &\textbf{42.3} & \textbf{52.1} &\textbf{37.5} \\
        
            \bottomrule
            \end{tabular}
}
\end{minipage}
\hfill
\begin{minipage}{0.49\textwidth}
\centering
\caption{\textbf{Reasoning Affordance Segmentation.}}
\label{tbl:reasoning_affordance}
\scriptsize
\resizebox{\linewidth}{!}{
        \begin{tabular}{l|c|cc|cc|cc}
            \toprule
            \multicolumn{1}{l|}{\multirow{2}*{Method}} & \multicolumn{1}{c|}{\multirow{2}*{Size}} & \multicolumn{2}{c|}{HANDAL (easy)}  & \multicolumn{2}{c|}{HANDAL (hard)} & \multicolumn{2}{c}{\cellcolor{gray!30}3DOI}\\
            
            & & gIoU & cIoU & gIoU & cIoU & \cellcolor{gray!30}gIoU & \cellcolor{gray!30}cIoU  \\
            \midrule
            G-DINO \cite{liu2024groundingdino}& -- & 3.6 & 3.0 & 3.4 & 3.1 & 4.1 & 3.9 \\
            LISA \cite{lai2024lisa} & 7B & 15.5 & 11.9 & 12.3 & 8.1 &12.3 &  8.1  \\
            GLaMM \cite{rasheed2024glamm} & 3.8B & 4.7 & 3.5 & 5.0 & 3.5 & 4.4 &2.9\\
            UniPixel \cite{liu2025unipixel} & 7B & 26.7 & 15.7 &24.5 & 11.6& 38.5 & 21.0 \\
            RAGNet \cite{wu2025ragnet} & 7B & \underline{58.3} & \underline{58.1}  & \underline{58.2} & \underline{57.8} & \underline{38.1} & \underline{39.4} \\
            
            \midrule
            \rowcolor{orange!10} \textbf{UniRAP} (Ours) & 3B & \textbf{63.2} & \textbf{64.1} & \textbf{62.9} & \textbf{63.8} & \textbf{55.7} & \textbf{54.5} \\
            \bottomrule
        \end{tabular}
}
\end{minipage}
\end{table}

\subsection{Referring and Reasoning Affordance Segmentation Performance}
\vspace{-2mm}
To evaluate UniRAP’s capability in affordance segmentation, we followed the evaluation protocol of RAGNet \cite{wu2025ragnet} and conducted systematic experiments on the annotated HANDAL \cite{guo2023handal}, GraspNet \cite{fang2020graspnet}, and 3DOI \cite{qian20233doi} datasets. \setlength{\fboxsep}{2pt}\colorbox{gray!30}{Gray} indicates zero-shot benchmarks.

As shown in Tab. \ref{tbl:affordance_segmentation} and Tab. \ref{tbl:reasoning_affordance}, UniRAP achieves strong performance on both explicit text-guided and reasoning-based affordance segmentation tasks. In referring affordance segmentation, UniRAP significantly outperforms RAGNet, which is also designed for affordance modeling, and demonstrates notably better generalization in unseen scenarios. Benefiting from its transfer-learning-oriented design, this advantage becomes even more pronounced in reasoning-based affordance segmentation. On the zero-shot 3DOI dataset, UniRAP improves gIoU and cIoU by 17.6 and 15.1, respectively, indicating stronger robustness and adaptability in complex semantic reasoning and spatial localization.

\begin{table*}[t]
\centering
\caption{
\textbf{Actionable Perception Performance on RealVLG-Benchmark} \cite{li2026realvlg}.
}
\label{tab:realvlg_benchmark}
\resizebox{0.98\textwidth}{!}{
\begin{tabular}{l|c|ccc|ccc|ccc}
\toprule
\multirow{2}{*}{Method} & \multirow{2}{*}{Size}
& \multicolumn{3}{c|}{Seen} 
& \multicolumn{3}{c|}{Similar} 
& \multicolumn{3}{c}{\colorbox{gray!20}{Novel}} \\

\cmidrule(lr){3-5} \cmidrule(lr){6-8} \cmidrule(lr){9-11}

& & Bbox & Seg. & Grasp
& Bbox & Seg. & Grasp
& Bbox & Seg. & Grasp \\
& & gIoU/cIoU & F$_\beta$/S$_\alpha$ & mIoU/gAcc
& gIoU/cIoU & F$_\beta$/S$_\alpha$ & mIoU/gAcc
& gIoU/cIoU & F$_\beta$/S$_\alpha$ & mIoU/gAcc \\
\midrule

Qwen-VL-Max \cite{alibaba2025qwenvlmax} & -
& 56.6/57.1 & 62.8/32.0 & 2.3/2.4
& 57.5/57.9 & 57.6/28.9 & 2.7/1.3
& 59.8/60.1 & 30.3/24.7 & 1.8/1.2 \\

Gemini2.5-Flash \cite{deepmind_gemini_flash} & - 
& 50.1/51.3 & 2.8/11.8 & 3.2/1.2
& 47.9/49.0 & 3.1/10.6 & 2.5/2.2
& 59.5/60.6 & 3.7/14.2 & 3.4/1.9 \\

\midrule

Qwen2.5VL \cite{Qwen2.5-VL} & 7B 
& 56.8/58.3 & 80.6/43.8 & 1.8/0.8
& 57.1/57.7 & 77.7/42.2 & 1.6/0.0
& 55.6/56.8 & 48.3/39.2 & 1.0/0.0 \\

Qwen2.5VL+SFT \cite{Qwen2.5-VL}& 7B 
& 64.8/63.5 & 82.2/44.1 & 4.1/2.5
& 66.7/62.7 & 79.6/42.6 & 3.9/2.3
& 64.9/65.2 & 48.5/39.4 & 2.7/1.1 \\

RealVLG-R1(GRPO) \cite{li2026realvlg} & 7B 
& \underline{88.0}/\underline{88.5} & \underline{89.2}/\underline{45.3} & 32.1/39.1
& 83.4/84.2 & 85.0/43.8 & \underline{28.7}/\underline{36.2}
& \underline{84.6}/\underline{85.4} & 52.2/39.7 & \underline{16.8}/\underline{18.9} \\

RealVLG-R1(GSPO) \cite{li2026realvlg}& 7B 
& \textbf{89.0}/\textbf{89.6} & 88.9/\textbf{45.4} & 33.6/32.8
& \textbf{86.4}/\textbf{86.9} & \underline{86.6}/\underline{43.8} & 27.9/30.2
& \textbf{88.5}/\textbf{88.9} & \underline{52.0}/\underline{39.7} & 16.5/18.3 \\

\midrule

Qwen2.5VL \cite{Qwen2.5-VL} & 3B 
& 49.6/51.5 & 74.5/43.2 & 4.0/2.7
& 48.4/49.5 & 71.0/40.1 & 3.7/1.3
& 51.3/52.1 & 44.5/37.5 & 3.0/1.3 \\

Qwen2.5VL+SFT \cite{Qwen2.5-VL} & 3B 
& 56.4/57.1 & 76.2/43.4 & 3.4/1.7
& 54.7/55.9 & 75.4/43.2 & 3.6/2.1
& 57.2/58.6 & 46.2/38.9 & 4.4/1.5 \\

RealVLG-R1(GRPO) \cite{li2026realvlg} & 3B 
& 83.8/84.2 & 83.9/43.4 & \underline{34.7}/\underline{40.3}
& \underline{84.7}/\underline{85.3} & 84.1/43.0 & 28.2/31.9
& 77.1/78.1 & 49.4/38.7 & 16.3/17.1 \\

RealVLG-R1(GSPO) \cite{li2026realvlg} & 3B 
& 87.7/88.4 & 88.7/45.3 & 29.2/33.6
& 84.6/85.3 & 85.0/43.7 & 27.6/30.6
& 78.0/79.3 & 49.1/39.6 & 15.3/9.1 \\

\rowcolor{orange!10} \textbf{UniRAP} (Ours) & 3B 
& 85.5/86.1 & \textbf{92.0}/43.6 & \textbf{43.8}/\textbf{56.7}
& 83.6/83.0 & \textbf{89.4}/\textbf{46.9} & \textbf{40.8}/\textbf{56.7}
& 68.4/72.3 & \textbf{75.8}/\textbf{42.4} & \textbf{39.5}/\textbf{63.3} \\
\bottomrule
\end{tabular}
}
\end{table*}

\subsection{Actionable Perception Performance}
We evaluated the actionable perception performance of UniRAP on the RealVLG-Benchmark \cite{li2026realvlg}, focusing on its capability in both coarse-grained segmentation and fine-grained interaction grounding. Given a natural language description of a target object, the benchmark requires the model to reason about its bounding box, segmentation mask, and 4-DoF grasp pose. Details of the evaluation metrics are provided in Appendix \ref{appendix_metric}.

As illustrated in Tab. \ref{tab:realvlg_benchmark}, for coarse-grained object detection, RealVLG-R1 \cite{li2026realvlg} retains a certain advantage, benefiting from the large-scale detection pretraining of the foundation model Qwen2.5-VL \cite{Qwen2.5-VL}. However, for segmentation tasks, existing baselines are unable to perform end-to-end prediction and instead rely on a two-stage ``bbox + SAM2'' pipeline, which introduces error accumulation and leads to slightly inferior performance compared to UniRAP, particularly in zero-shot \texttt{Novel} scenarios. For fine-grained 4-DoF grasp pose estimation, text-based generation baselines struggle to capture pixel-level localization, whereas UniRAP’s end-to-end formulation demonstrates clear advantages, enabling both stable grasp prediction and stronger generalization. In terms of grasp accuracy (gAcc), especially in novel settings, UniRAP achieves up to a threefold improvement over RealVLG-R1.

\subsection{Ablation Study}
\begin{table*}[t]
\centering
\caption{\textbf{Ablation Study} of different pre-training stages,
fine-tuning strategies, and prediction heads.}
\label{tbl:ablation}
\small
\setlength{\tabcolsep}{3.5pt}
\resizebox{\textwidth}{!}{
\begin{tabular}{cccccccccccc}
\toprule
\multicolumn{4}{c}{Training Strategy} &
\multicolumn{3}{c}{Prediction Head} &
\multicolumn{5}{c}{Evaluation Benchmark} \\
\cmidrule(lr){1-4}
\cmidrule(lr){5-7}
\cmidrule(lr){8-12}

\shortstack{General\\Pre-train} &
\shortstack{Affordance\\Pre-train} &
\shortstack{Visual Referring\\Pre-train} &
\shortstack{Actionable Affordance\\Fine-tune} &
\shortstack{Bbox\\Head} &
\shortstack{Heatmap Grasp\\Head} &
\shortstack{UniRAP Grasp\\Head} &
\shortstack{ReasonSeg\\(gIoU)} &
\shortstack{ReVOS\\(J\&F)} &
\shortstack{3DOI\\(Ref. gIoU)} &
\shortstack{3DOI\\(Reason, gIoU)} &
\shortstack{RealVLG-Bench\\(Grasp, gAcc)} \\
\midrule

\checkmark & & & & & & &
55.4 & 58.3 & 40.5 & 37.5 & -- \\

\checkmark & \checkmark & & & & & &
56.4 & 59.2 & 50.4 & 52.9 & -- \\

\checkmark & \checkmark & \checkmark & &
\checkmark & & \checkmark &
57.3 & 60.2 & 51.6 & 54.6 & 19.6 \\

\checkmark & \checkmark & & \checkmark &
\checkmark & & \checkmark &
57.0 & 59.8 & 50.8 & 54.2 & 26.4 \\

\checkmark & \checkmark & \checkmark & \checkmark &
& & \checkmark &
56.8 & 60.4 & 51.3 & 53.8 & 24.7 \\

\checkmark & \checkmark & \checkmark & \checkmark &
\checkmark & \checkmark & &
57.6 & 61.7 & \textbf{52.3} & 55.6 & 13.4 \\

\rowcolor{orange!10} \checkmark & \checkmark & \checkmark & \checkmark &
\checkmark & & \checkmark &
 \textbf{57.9} & \textbf{61.9} & 52.1 &
 \textbf{55.7} & \textbf{63.3} \\

\bottomrule
\end{tabular}
}
\vspace{-4mm}
\end{table*}

We conducted ablation studies on selected key components of UniRAP, as summarized in Tab.~\ref{tbl:ablation}. Affordance pre-training substantially improved affordance understanding and reasoning segmentation, increasing the Referring and Reasoning gIoU on 3DOI by 9.9 and 15.4 points, respectively. Visual referring pre-training further strengthened the association between language instructions and actionable interaction poses, improving grasp accuracy from 26.4 to 63.3 when combined with our proposed Grasp Head. Actionable affordance fine-tuning jointly optimized segmentation, localization, and grasp prediction, improving grasp accuracy from 19.6 to 63.3 while maintaining competitive segmentation performance. Regarding prediction heads, the Bbox Head provided additional spatial constraints for the SAM2 decoder, suppressing background interference and improving ReasonSeg, ReVOS, and 3DOI performance by 1.1, 1.5, and up to 1.9 points, respectively. Finally, replacing our regression-based Grasp Head with a heatmap-based alternative reduced grasp accuracy from 63.3 to 13.4, demonstrating the effectiveness of direct grasp prediction from Transformer representations.

\begin{wrapfigure}{r}{0.65\columnwidth}
\vspace{-10pt}
\centering

\begin{minipage}{\linewidth}
\centering
\includegraphics[width=\linewidth]{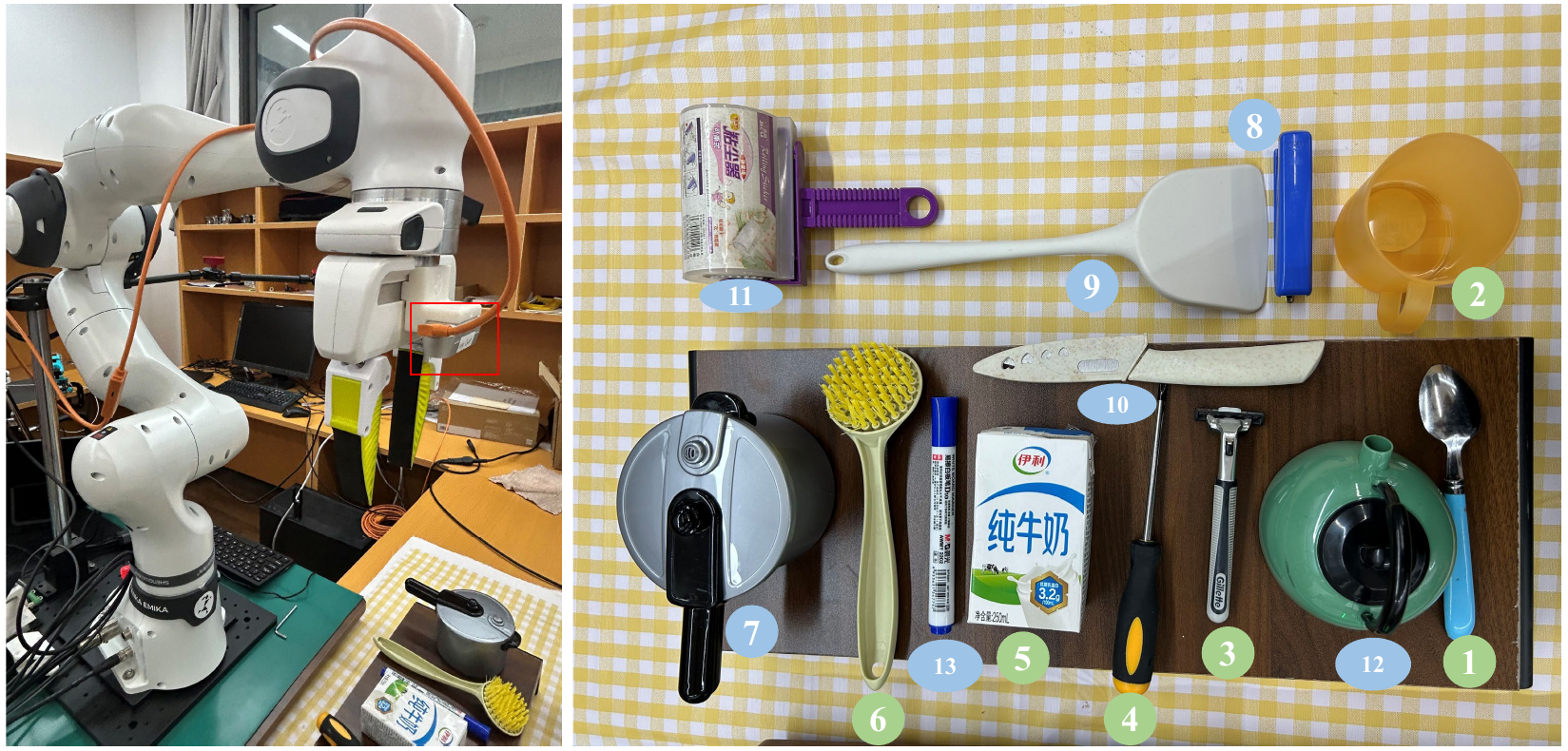}
\vspace{-3mm}

\caption{
\textbf{Real-World Experimental Setup.}
A Franka FR3 robotic arm is equipped with an
eye-in-hand RealSense D435i camera.
Objects 1--6 evaluate referring grasping,
while objects 7--13 evaluate reasoning-based grasping.
}
\label{fig:robot_setting_main}
\end{minipage}

\vspace{2mm}

\begin{minipage}{\linewidth}
\centering

\captionof{table}{
\textbf{Real-world robotic grasping success rates (\%).}
}
\label{tab:realworld_grasping_main}

\vspace{1mm}
\scriptsize
\setlength{\tabcolsep}{2pt}

\resizebox{\linewidth}{!}{
\begin{tabular}{c|*{13}{c}|c}
\toprule
\multirow{2}{*}{\textbf{Method}}
& \multicolumn{13}{c|}{\textbf{Object ID}}
& \multirow{2}{*}{\textbf{Avg. SR (\%)}} \\
\cmidrule(lr){2-14}
& 1 & 2 & 3 & 4 & 5 & 6 & 7
& 8 & 9 & 10 & 11 & 12 & 13 & \\
\midrule

RAGNet + GraspNet
& 50 & 60 & 40 & 60 & 50 & 60 & 30
& 40 & 30 & 40 & 30 & 40 & 40 & 43.8 \\

UniRAP + GraspNet
& 70 & 70 & 60 & 70 & 60 & 50 & 40
& 70 & 50 & 50 & 40 & 40 & 50 & 55.4 \\

RealVLG-R1 (4-DoF)
& 80 & 60 & 70 & 80 & 70 & 70 & 50
& \textbf{80} & 40 & 70 & 40 & 20 & 80 & 62.3 \\

\midrule
\rowcolor{orange!10}
\textbf{UniRAP (4-DoF)}
& \textbf{100} & \textbf{90} & \textbf{80}
& \textbf{100} & \textbf{90} & \textbf{100}
& \textbf{80} & \textbf{80} & \textbf{70}
& \textbf{90} & \textbf{80} & \textbf{90}
& \textbf{100} & \textbf{88.5} \\

\bottomrule
\end{tabular}
}
\end{minipage}

\vspace{-10pt}
\end{wrapfigure}
\subsection{Application: Actionable Perception for Robotic Grasping}
To evaluate the transferability of UniRAP to real-world robotic manipulation, we conducted grasping experiments using a Franka FR3 robotic arm equipped with an eye-in-hand RealSense D435i camera, as illustrated in Fig.~\ref{fig:robot_setting_main}. The evaluation includes 13 household objects, with six objects for referring grasping and seven for reasoning-based grasping. We compared UniRAP against RAGNet and RealVLG-R1 under two manipulation paradigms: a modular perception-to-grasping pipeline that combines affordance segmentation with GraspNet, and direct 4-DoF grasp prediction. As shown in Table~\ref{tab:realworld_grasping_main}, RAGNet+GraspNet and UniRAP+GraspNet achieve average success rates of 43.8\% and 55.4\%, respectively, indicating that improved affordance perception alone does not fully resolve the limitations of downstream grasp generation. In particular, the GraspNet-based pipeline remains sensitive to small objects, reflective surfaces, and unreliable depth observations, where inaccurate geometric reconstruction can compromise grasp estimation even when the target region is correctly identified. In contrast, directly predicting actionable 4-DoF grasp poses allows UniRAP to reduce its dependence on a separate geometry-based grasp detector, achieving an average success rate of 88.5\%, substantially outperforming RealVLG-R1 (62.3\%) and UniRAP+GraspNet (55.4\%). These results demonstrate the benefits of unifying language-grounded affordance understanding and grasp prediction for robust real-world robotic manipulation. More details are provided in Appendix \ref{appendix_realworld_grasping}.

%% file: sec/6_conclusion.tex
\section{Conclusion}
We proposed UniRAP, a unified vision-language model that bridges high-level vision-language reasoning with fine-grained physical contact perception. By introducing a token-based interface between the VLM backbone and a dense decoder, UniRAP enables end-to-end mapping from language instructions to interaction-centric geometric representations, supporting affordance segmentation, target region detection, and 4-DoF interaction pose estimation. A curriculum-based transfer learning strategy further improves generalization across heterogeneous tasks. Extensive experiments show consistent gains in both perception and interaction alignment, which establishes a more robust foundation for language-guided robotic systems in dynamic environments. To the best of our knowledge, UniRAP is the first end-to-end, reasoning-aware unified model for actionable affordance perception across both images and videos.

%% file: sec/7_Appendix.tex
\clearpage
\newpage
\appendix

\section*{Appendix}

\begin{enumerate}[label=\textbf{\Alph*.}]
    \item UniRAP Model
    \begin{enumerate}[label=\textbf{\Alph{enumi}.\arabic*.}]
        \item Architecture
        \item Visual Referring Encoder
    \end{enumerate}

    \item Experiments
    \begin{enumerate}[label=\textbf{\Alph{enumi}.\arabic*.}]
        \item Implementation Details
        \item Training Recipe
        \item Training Objective
        \item Evaluation Metrics
        \item Qualitative Results
    \end{enumerate}

    \item Transferring Actionable Perception to Real-World Robotic Manipulation
    \begin{enumerate}[label=\textbf{\Alph{enumi}.\arabic*.}]
        \item Motivation
        \item UniRAP for Reasoning-Aware Robotic Grasping
    \end{enumerate}

    \item Discussions
    \begin{enumerate}[label=\textbf{\Alph{enumi}.\arabic*.}]
        \item Limitations \& Future Work
        \item Potential Societal Impacts
    \end{enumerate}

    \item Licenses
\end{enumerate}

\section{UniRAP Model}
\subsection{Architecture}
Within the unified vision–language segmentation paradigm, recent approaches \cite{liu2025unipixel, yuan2025sa2va, wu2025ragnet, lai2024lisa} commonly adopt an embedding-as-mask design. In this framework, multimodal large language models generate specific tokens (e.g., \texttt{<SEG>} \cite{lai2024lisa}, \texttt{<AFF> \cite{wu2025ragnet}}, and \texttt{TRK} \cite{bai2024videolisa}) during text decoding, and the corresponding high-level semantic representations are treated as prompt embeddings to guide segmentation models, such as SAM \cite{kirillov2023sam}, in producing target regions. This paradigm has been systematically formalized in LISA \cite{lai2024lisa} and its subsequent extensions, and has been widely reused across diverse tasks. Although these methods differ in implementation details and input modalities, for instance, UniPixel \cite{liu2025unipixel} supports both image and video inputs and introduces finer-grained supervision via the PixelQA task, they largely share an implicit assumption: visual understanding can be modeled as a direct mapping from semantic representations to pixel space, i.e., a semantic-to-dense prediction process. Under this assumption, variations across methods primarily manifest in task formulation and supervision granularity rather than in the underlying capability paradigm.

However, this assumption becomes inadequate in scenarios involving physical interaction. Taking contact point prediction as an example, the objective extends beyond identifying ``what the object is'' (semantic level) and ``where the object is'' (spatial level); it further requires reasoning about ``which local region of the object should be contacted'' and ``whether such contact is geometrically and functionally feasible and stable''. This problem inherently introduces additional dimensions of reasoning, including local geometric constraints, potential interaction affordances, and task-relevant functional semantics, thereby exceeding the expressive capacity of conventional segmentation formulations.

Motivated by this observation, UniRAP extends the applicability of the unified paradigm. Rather than treating the \texttt{<SEG>} token solely as a semantic anchor for region localization, UniRAP leverages its high-level representation to encode potential physical interaction intent, and, through coupling with a segmentation model, enables a cross-level mapping from semantic understanding to contact reasoning. In contrast to existing methods that primarily focus on pixel-wise correspondences, UniRAP addresses a more structured prediction problem: localizing contact regions in continuous space that satisfy both physical and functional constraints. From a capability perspective, UniRAP therefore does not merely apply existing architectures to a new task, but rather generalizes the paradigm from perception-driven dense prediction to interaction-oriented reasoning under physical constraints. This extension suggests that language-guided segmentation frameworks are not only applicable to conventional visual understanding tasks, but also hold potential for advancing toward higher-level embodied intelligence.

\subsection{Visual Referring Encoder}
For UniRAP, both point and box prompts are uniformly represented as spatiotemporal coordinates $(x, y, t)$, where bounding boxes are discretized into representative points such as corners or uniformly sampled boundary locations. We adopt a Fourier mapping \cite{tancik2020fouriermapping} with random frequency bases to embed low-dimensional coordinates into a high-dimensional representation space.

Specifically, we define the positional encoding as:
\begin{equation}
\gamma(\mathbf{c}) =
\left[
\sin(\mathbf{B}\mathbf{c}),\;
\cos(\mathbf{B}\mathbf{c})
\right],
\quad \mathbf{c} \in \mathbb{R}^{3}
\end{equation}
where $\mathbf{c} = (x, y, t)^\top$, and $\mathbf{B} \in \mathbb{R}^{L \times 3}$ is a random Gaussian matrix whose entries are sampled as:
\begin{equation}
\mathbf{B}_{ij} \sim \mathcal{N}(0, \sigma^2)
\end{equation}
This random Fourier feature embedding allows the model to capture high-frequency spatial and temporal variations in a linearly separable representation space, which is essential for precise localization and interaction-sensitive reasoning.

Next, the encoded prompt embeddings are transformed into discrete tokens via a learnable linear projection and further interact within LLM, enabling the joint modeling of language-driven semantics and geometry-aware visual prompts. While this design is methodologically inspired by the SAM \cite{kirillov2023sam}, it differs fundamentally in the pathway of information interaction: the output of the Visual Referring Encoder is no longer fused directly with visual encoder features. Instead, it engages in interaction with the LLM, allowing prompt features to be explicitly aggregated into the semantic representation of the \texttt{<SEG>} token, thereby achieving cross-modal alignment within a unified semantic space.

Furthermore, this mechanism is distinct from approaches such as SA2VA \cite{yuan2025sa2va} and UniPixel \cite{liu2025unipixel}. These methods typically convert structured prompts, such as points and bounding boxes, into mask representations for region-based visual question answering tasks. In contrast, our proposed approach adopts a token-level interaction paradigm, mapping tasks without explicit textual annotations into a text-guided \texttt{<SEG>} token representation space. This unification reduces distributional discrepancies across tasks at the representation level and mitigates the complexity of cross-task transfer learning.

\section{Experiments}
\label{appendix_experiments}

\subsection{Implementation Details}
\label{appendix_iImplementation}
UniRAP is built upon Qwen2.5-VL (3B) \cite{Qwen2.5-VL} and SAM 2.1 \cite{ravi2024sam2}, with the hidden dimensionality of the visual prompt encoder set to 256.

Within the affordance decoder, all ground-truth annotations are uniformly resampled to a fixed resolution of $768\times768$ via center padding. This design differs from the direct resizing strategy adopted in SAM2. The motivation is that tasks requiring coordinate inference based on object geometry, such as bounding box estimation and grasp pose prediction, are sensitive to geometric distortions, which can be significantly exacerbated by naive resizing and consequently degrade prediction accuracy.

For bounding box representation, we adopt the parameterization $(c_x, c_y, w, h)$. After center padding, all elements are normalized with respect to the scale of 768. For 4-DoF grasp pose prediction, the representation $(c_x, c_y, \cos 2\theta, \sin 2\theta, w)$ is employed, where $(c_x, c_y)$ denotes the grasp center and $\theta$ represents the counterclockwise angle relative to the x-axis. To address the periodicity of angular variables, the angle is encoded as $(\cos 2\theta, \sin 2\theta)$. The parameter $w$ corresponds to the width of the grasp rectangle (see Fig. \ref{fig:uad} for illustration).  The number of candidate grasp poses $N_g$ is 32.

During training, LoRA is applied to fine-tune Qwen2.5-VL, with the rank and scaling factor set to 128 and 256, respectively. For video inputs, 8 frames are randomly sampled per instance. All experiments are conducted on 8 A800 (80GB) GPUs, and the model is trained for a single epoch.

\subsection{Training Recipe}
\label{appendix:train_recipe}
The four stages of Curriculum Transfer Training and their corresponding dataset configurations are summarized in Tab. \ref{tab:training}.

In the first stage, a segmentation dataset comprising 461K samples is used to pre-train the LoRA adapters, the \texttt{<SEG>} token projection layer, and all parameters of SAM2. Within the Unified Affordance Decoder, only the segmentation head (Seg Head) is optimized, while the remaining components are kept frozen or excluded from training.

The second stage follows the same training configuration as the first stage, but employs a 253K-scale affordance segmentation dataset. Notably, due to the substantial redundancy of scenes in the HANDAL \cite{guo2023handal} dataset, only half of the samples from each scene are randomly selected for training to mitigate potential bias introduced by duplicated data.

In the third stage, both segmentation datasets with bounding box annotations and grasp datasets without language annotations are utilized to train the visual prompt encoder, as well as the bbox and grasp heads. This stage also involves optimizing the shared Two-way Transformer, with the objective of aligning \texttt{<REF>} prompt features to the feature space of the \texttt{<SEG>} token.

In the fourth stage, the model is initialized with the weights obtained from the previous stage, and only the Unified Affordance Decoder is trained. While preserving the original textual reasoning capability of the vision-language model, this stage further fine-tunes the decoder to align with the prompt features encoded in the \texttt{<SEG>} token space.

\begin{table}[t]
\scriptsize
\centering
\caption{\textbf{Distribution of Training Datasets for UniRAP.} $\Circle$ denotes random selection between two prompt types per sample; $^\dagger$ indicates text prompts annotated by RAGNet \cite{wu2025ragnet}. The fourth stage uses the RealVLG dataset \cite{li2026realvlg}.  Stage 1: General Pre-training; Stage 2: Affordance Pre-training; Stage 3: Visual Referring Pre-training; Stage 4: Actionable Affordance Fine-tuning.}
\vspace{1.2mm}
\resizebox{0.98\textwidth}{!}{
\begin{tabular}{c|p{2.85cm}<{\raggedright}|ccccc|cccc|rr}

\toprule
\multirow{2}{*}{\textbf{Stage}} & \multirow{2}{*}{\textbf{Dataset}} & \multicolumn{5}{c|}{\textbf{Inputs}} & \multicolumn{4}{c|}{\textbf{Outputs}} & \multirow{2}{*}{\textbf{\#Samples}} & \multirow{2}{*}{\textbf{\#Repeat}} \\
\cmidrule{3-8} \cmidrule{9-10}
&& Text & Image & Video & Point & Box & Text & Mask & Box & Grasp && \\

\midrule
\rowcolor{blue!3.5} & RefCOCO \cite{refcoco} & \cmark & \cmark &&&& \cmark & \cmark &&& 17K & 5 \\
\rowcolor{blue!3.5} & RefCOCO+ \cite{refcoco} & \cmark & \cmark &&&& \cmark & \cmark &&& 17K & 5 \\
\rowcolor{blue!3.5} & RefCOCOg \cite{refcocog} & \cmark & \cmark &&&& \cmark & \cmark &&& 22K & 5 \\
\rowcolor{blue!3.5} & RefClef \cite{refcoco} & \cmark & \cmark &&&& \cmark & \cmark &&& 18K & 5  \\
\rowcolor{blue!3.5} & ReasonSeg \cite{lai2024lisa} & \cmark & \cmark &&&& \cmark & \cmark &&& 1.6K & 10  \\
\rowcolor{blue!3.5} & ADE20K \cite{ade20k} & \cmark & \cmark &&&& \cmark &\cmark &&& 20K &3  \\
\rowcolor{blue!3.5} & COCOStuff \cite{caesar2018cocostuff} &\cmark &\cmark &&&& \cmark &\cmark &&& 118K & 3 \\
\rowcolor{blue!3.5} 1 & MapillaryVistas \cite{mapillary} & \cmark & \cmark &&&& \cmark & \cmark &&& 18K & 3 \\
\rowcolor{blue!3.5} & PACO-LVIS \cite{ramanathan2023paco} & \cmark & \cmark &&&& \cmark & \cmark &&& 46K & 3  \\
\rowcolor{blue!3.5} & PASCAL-Part \cite{chen2014pacopart} & \cmark & \cmark &&&& \cmark & \cmark &&& 4.4K & 3  \\
\rowcolor{blue!3.5} & Ref-YouTube-VOS \cite{urvos} & \cmark && \cmark &&& \cmark & \cmark &&& 13K & 3 \\
\rowcolor{blue!3.5} & Ref-DAVIS17 \cite{davis2017} & \cmark && \cmark &&& \cmark & \cmark &&& 0.6K & 10  \\
\rowcolor{blue!3.5} & Ref-SAV \cite{yuan2025sa2va} & \cmark && \cmark &&& \cmark & \cmark &&& 56K & 3  \\
\rowcolor{blue!3.5} & MeViS \cite{ding2023mevis} & \cmark && \cmark &&& \cmark & \cmark &&& 23K & 5  \\
\rowcolor{blue!3.5} & LV-VIS \cite{wang2023lvvis} & \cmark && \cmark &&& \cmark & \cmark &&& 11K & 3 \\
\rowcolor{blue!3.5} & ViCaS \cite{athar2025vicas} & \cmark && \cmark &&& \cmark & \cmark &&& 41K & 3 \\
\rowcolor{blue!3.5} & ReVOS \cite{yan2024visa} & \cmark && \cmark &&& \cmark & \cmark &&& 29K & 5 \\
\rowcolor{blue!3.5} & GroundMoRe \cite{deng2025groundmore} & \cmark && \cmark &&& \cmark & \cmark &&& 5.6K & 3 \\

\midrule
\rowcolor{blue!8.5} & ReasonSeg \cite{lai2024lisa} & \cmark & \cmark &&&& \cmark & \cmark &&& 1.6K & 10 \\
\rowcolor{blue!8.5} & $^\dagger$HANDAL \cite{guo2023handal} & \cmark & \cmark &&&& \cmark & \cmark & & & 211K & 3\\
\rowcolor{blue!8.5} 2 & $^\dagger$EgoObjects \cite{zhu2023egoobjects} & \cmark & \cmark &&&& \cmark & \cmark & & & 12K & 3 \\ 
\rowcolor{blue!8.5} & $^\dagger$OpenX \cite{o2024openx} & \cmark & \cmark &&&& \cmark & \cmark & & & 11K & 3 \\
\rowcolor{blue!8.5} & $^\dagger$RLBench \cite{james2020rlbench} & \cmark & \cmark &&&& \cmark & \cmark & & & 5.6K & 3 \\
\rowcolor{blue!8.5} & $^\dagger$GraspNet \cite{fang2020graspnet} & \cmark & \cmark &&&& \cmark & \cmark & & & 12K & 3  \\

\midrule
\rowcolor{orange!8.5} & RefCOCO \cite{refcoco} & \cmark & \cmark &&\Circle&\Circle& \cmark & \cmark &&& 17K & 5 \\
\rowcolor{orange!8.5} & RefCOCO+ \cite{refcoco} & \cmark & \cmark &&\Circle&\Circle& \cmark & \cmark &&& 17K & 5 \\
\rowcolor{orange!8.5} & RefCOCOg \cite{refcocog} & \cmark & \cmark &&\Circle&\Circle& \cmark & \cmark &&& 22K & 5  \\
\rowcolor{orange!8.5} & RefClef \cite{refcoco} & \cmark & \cmark &&\Circle&\Circle& \cmark & \cmark &&& 18K & 5  \\
\rowcolor{orange!8.5} 3 & Jacquard \cite{depierre2018jacquard} & \cmark & \cmark &&\Circle&\Circle& \cmark & \cmark &\cmark&\cmark& 54K & 1\\
\rowcolor{orange!8.5} & Cornell \cite{cornell} & \cmark & \cmark &&\Circle&\Circle& \cmark & \cmark &\cmark&\cmark& 0.9K & 10 \\
\rowcolor{orange!8.5} & OCID-Grasp \cite{ocidgrasp} & \cmark & \cmark &&\Circle&\Circle& \cmark & \cmark &\cmark&\cmark& 12K & 5 \\
\rowcolor{orange!8.5} & GraspNet \cite{fang2020graspnet} & \cmark & \cmark &&\Circle&\Circle& \cmark & \cmark &\cmark&\cmark& 26K & 2 \\

\midrule
\rowcolor{orange!35} & Cornell-VLG \cite{cornell,li2026realvlg} & \cmark & \cmark &&&& \cmark & \cmark &\cmark&\cmark& 0.9K & 5\\
\rowcolor{orange!35} & OCID-VLG \cite{ocidgrasp,li2026realvlg} & \cmark & \cmark &&&& \cmark & \cmark &\cmark&\cmark& 12K & 5 \\
\rowcolor{orange!35} 4 & VMRD-VLG \cite{vmrd,li2026realvlg} & \cmark & \cmark &&&& \cmark & \cmark &\cmark&\cmark& 11K & 5 \\
\rowcolor{orange!35} & GraspNet-VLG \cite{fang2020graspnet,li2026realvlg} & \cmark & \cmark &&&& \cmark & \cmark &\cmark&\cmark& 26K & 5  \\
\bottomrule
\end{tabular}}

\label{tab:training}
\vspace{-15px}
\end{table}

\subsection{Training Objective}
\label{appendix_loss}

We optimize a weighted sum of task-specific loss functions:
\begin{equation}
\mathcal{L} = \mathcal{L}_{\text{txt}} + 100\,\mathcal{L}_{\text{mask}} + 5\,\mathcal{L}_{\text{dice}} + 5\,\mathcal{L}_{\text{iou}} + 5\,\mathcal{L}_{\text{cls}} + 0.5\,\mathcal{L}_{\text{L1}} + 0.5\,\mathcal{L}_{\text{GIoU}} + 2\,\mathcal{L}_{\text{grasp}} + 0.1\,\mathcal{L}_{\text{mib}} + 0.1\,\mathcal{L}_{\text{gcm}}.
\end{equation}
The weighting coefficients were determined through validation-based tuning, adhering to segmentation-centric practices. This designates $\hat{\mathbf{M}}$ as the primary spatial signal, to which the remaining branches are anchored.

\textbf{Language Modeling.}
The loss $\mathcal{L}_{\text{txt}}$ corresponds to the standard autoregressive cross-entropy computed over the predicted output tokens.

\textbf{Affordance Mask.}
Among $K$ candidate masks, only the optimal candidate receives gradient updates:
\begin{equation}
k^* = \arg\min_k \left( \lambda_{\text{mask}}\mathcal{L}_{\text{mask}}^{(k)} + \lambda_{\text{dice}}\mathcal{L}_{\text{dice}}^{(k)} \right),
\end{equation}
while all other candidates employ stop-gradient. The individual loss terms are defined as follows:
\begin{align}
\mathcal{L}_{\text{mask}} &= -\frac{1}{N}\sum_i \alpha_t (1-p_t)^\gamma \log p_t, \\
\mathcal{L}_{\text{dice}} &= 1 - \frac{2\sum_i \sigma(\hat{\mathbf{M}}_i)\,\mathbf{M}_i + 1}{\sum_i \sigma(\hat{\mathbf{M}}_i) + \sum_i \mathbf{M}_i + 1}.
\end{align}
An IoU prediction head produces the estimate $\hat{u}$:
\begin{equation}
\mathcal{L}_{\text{iou}} = \left\| \hat{u} - \frac{|\hat{\mathbf{M}}_{>0} \cap \mathbf{M}|}{|\hat{\mathbf{M}}_{>0} \cup \mathbf{M}|} \right\|_1,
\end{equation}
where the IoU target is computed from thresholded masks and treated as detached. Supervision is applied exclusively to $k^*$. An objectness focal loss $\mathcal{L}_{\text{cls}}$ gates all mask-related terms.

\textbf{Detection.}
Bounding box predictions $\hat{\mathbf{b}}$ are supervised through the following objectives:
\begin{equation}
\mathcal{L}_{\text{L1}} = \|\hat{\mathbf{b}} - \mathbf{b}\|_1, \quad
\mathcal{L}_{\text{GIoU}} = 1 - \text{GIoU}(\hat{\mathbf{b}}, \mathbf{b}).
\end{equation}

\textbf{Interaction Pose.}
Each grasp configuration $(c_x, c_y, \cos 2\theta, \sin 2\theta, w, q)$ is matched to the ground truth via a greedy assignment procedure. The resulting loss combines a Huber loss (applied to the center coordinates, width, and angle) with a binary cross-entropy loss for the quality score.

\textbf{Spatial Consistency.}
We introduce two asymmetric constraints to promote spatial coherence. The mask-in-bounding-box loss penalizes activations outside the predicted bounding box $\hat{\mathbf{b}}$:
\begin{equation}
\mathcal{L}_{\text{mib}} = \frac{1}{|\Omega_{\text{out}}|} \sum_{(i,j)\in\Omega_{\text{out}}} \log\bigl(1 + e^{\hat{\mathbf{M}}_{ij}}\bigr).
\end{equation}
Conversely, the grasp-center-in-mask loss encourages the predicted grasp centers to lie within $\hat{\mathbf{M}}$:
\begin{equation}
\mathcal{L}_{\text{gcm}} = \frac{1}{|\mathcal{G}_{\text{valid}}|} \sum_k \log\bigl(1 + e^{-\hat{\mathbf{M}}(\hat{c}_x^k,\,\hat{c}_y^k)}\bigr),
\end{equation}
where $\hat{\mathbf{M}}(\cdot,\cdot)$ denotes bilinear interpolation of the predicted mask evaluated at the specified coordinates. Both loss terms are normalized on a per-object basis and accumulated over the decoding steps.

\subsection{Evaluation Metrics}
\label{appendix_metric}
For image segmentation tasks (Tab. \ref{tab:res}, Tab. \ref{tbl:affordance_segmentation}, and Tab. \ref{tbl:reasoning_affordance}), we use gIoU and cIoU as the primary evaluation metrics. For video segmentation (Tab. \ref{tab:revos}), we adopt region similarity $\mathcal{J}$,  contour accuracy $\mathcal{F}$, and $\mathcal{J}\&\mathcal{F}$ for performance evaluation.  As for the evaluation of hallucination, we adopt the robustness score 
$\mathcal{R}$ introduced in R2VOS \cite{li2023r2vos}.

In the RealVLG-Benchmark (Tab. \ref{tab:realvlg_benchmark}), we follow the evaluation protocol of RealVLG-R1 \cite{li2026realvlg}: object detection is assessed using gIoU and cIoU, segmentation quality is measured by F-measure (F$_\beta$) and S-measure (S$_\alpha$), and grasping performance is evaluated using mIoU and gAcc. Here, gAcc (Grasp Accuracy) is defined as a successful grasp when the predicted grasp box achieves an IoU greater than 0.25 and an angular deviation less than 30°.

\subsection{Qualitative Results}
\label{appendix_results}
\begin{figure}[!ht]
\centering
\includegraphics[width=1.0\linewidth]{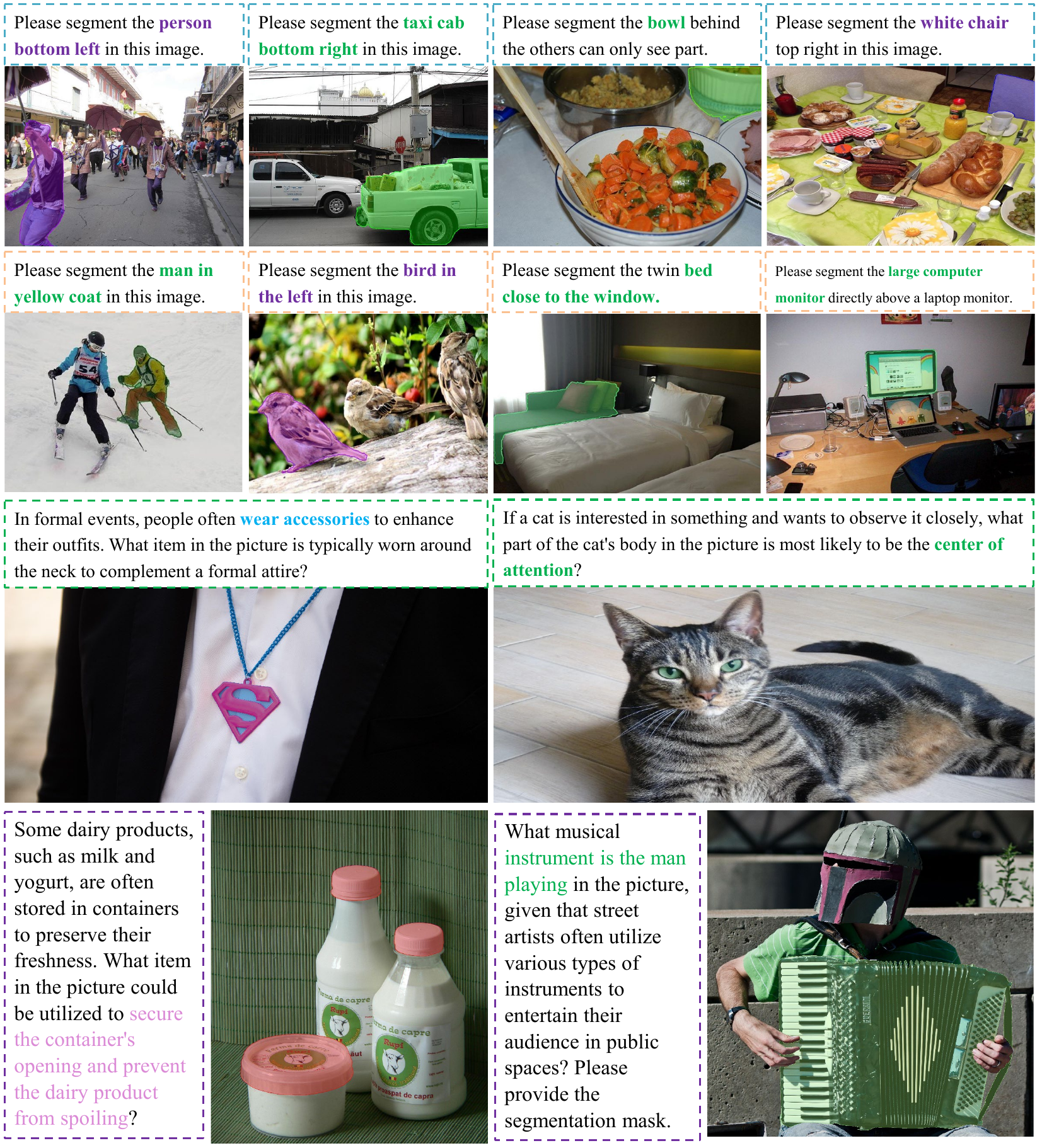}
\caption{\textbf{Qualitative Results of Referring Expression and Reasoning Segmentation.}}
\label{fig:res_vis}
\end{figure}

Fig. \ref{fig:res_vis} presents qualitative results for Referring Expression and Reasoning Segmentation on RefCOCO/+/g \cite{refcoco, refcocog} and ReasonSeg \cite{lai2024lisa}.

\begin{figure}[!ht]
\centering
\includegraphics[width=1.0\linewidth]{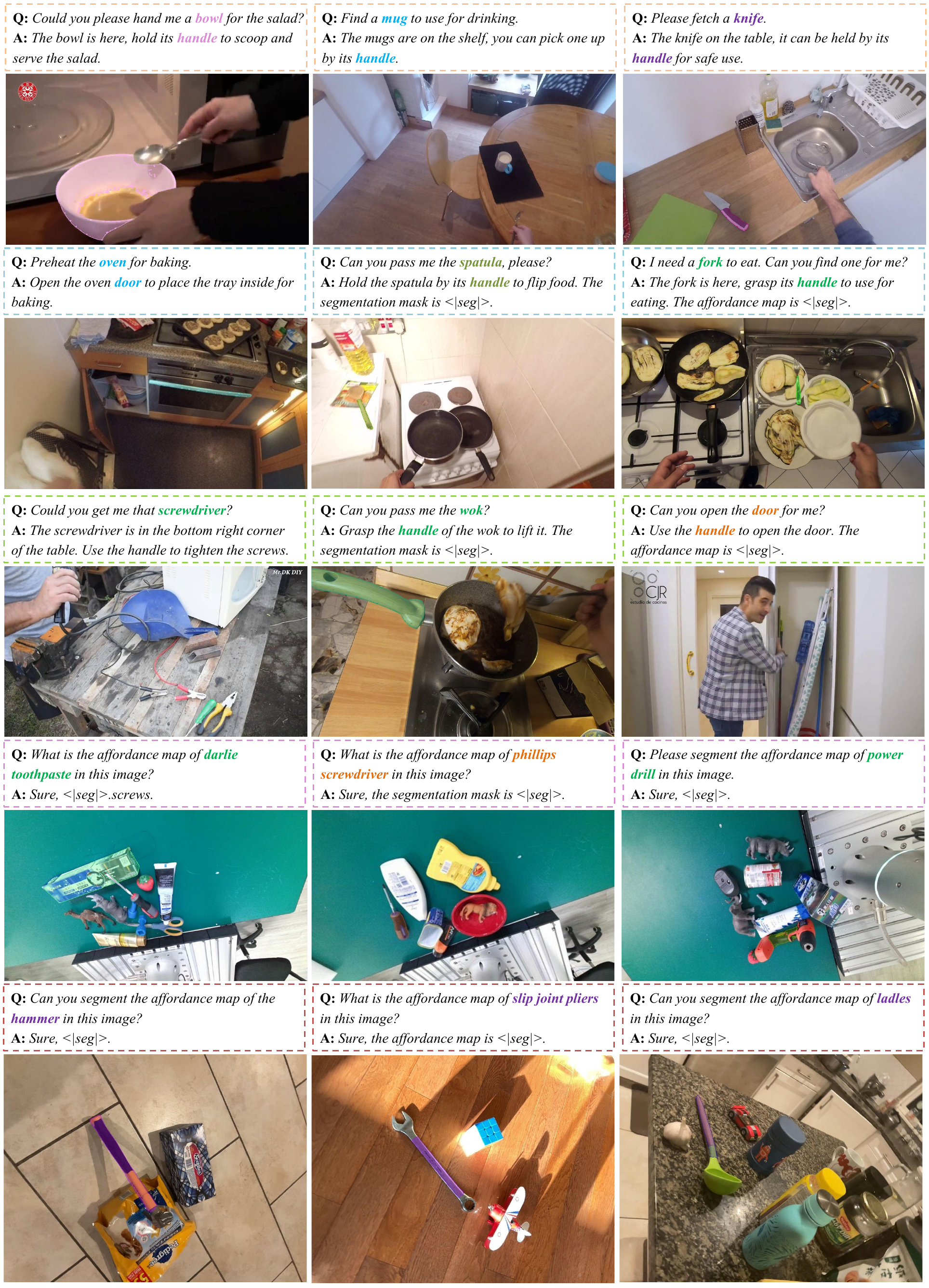}
\caption{\textbf{Qualitative Results of Affordance Segmentation.}}
\label{fig:aff_seg_vis}
\end{figure}

Fig. \ref{fig:aff_seg_vis} shows qualitative results for Affordance Segmentation on 3DOI \cite{qian20233doi}, HANDAL \cite{guo2023handal}, and GraspNet \texttt{Novel} \cite{fang2020graspnet}.

\begin{figure}[!ht]
\centering
\includegraphics[width=1.0\linewidth]{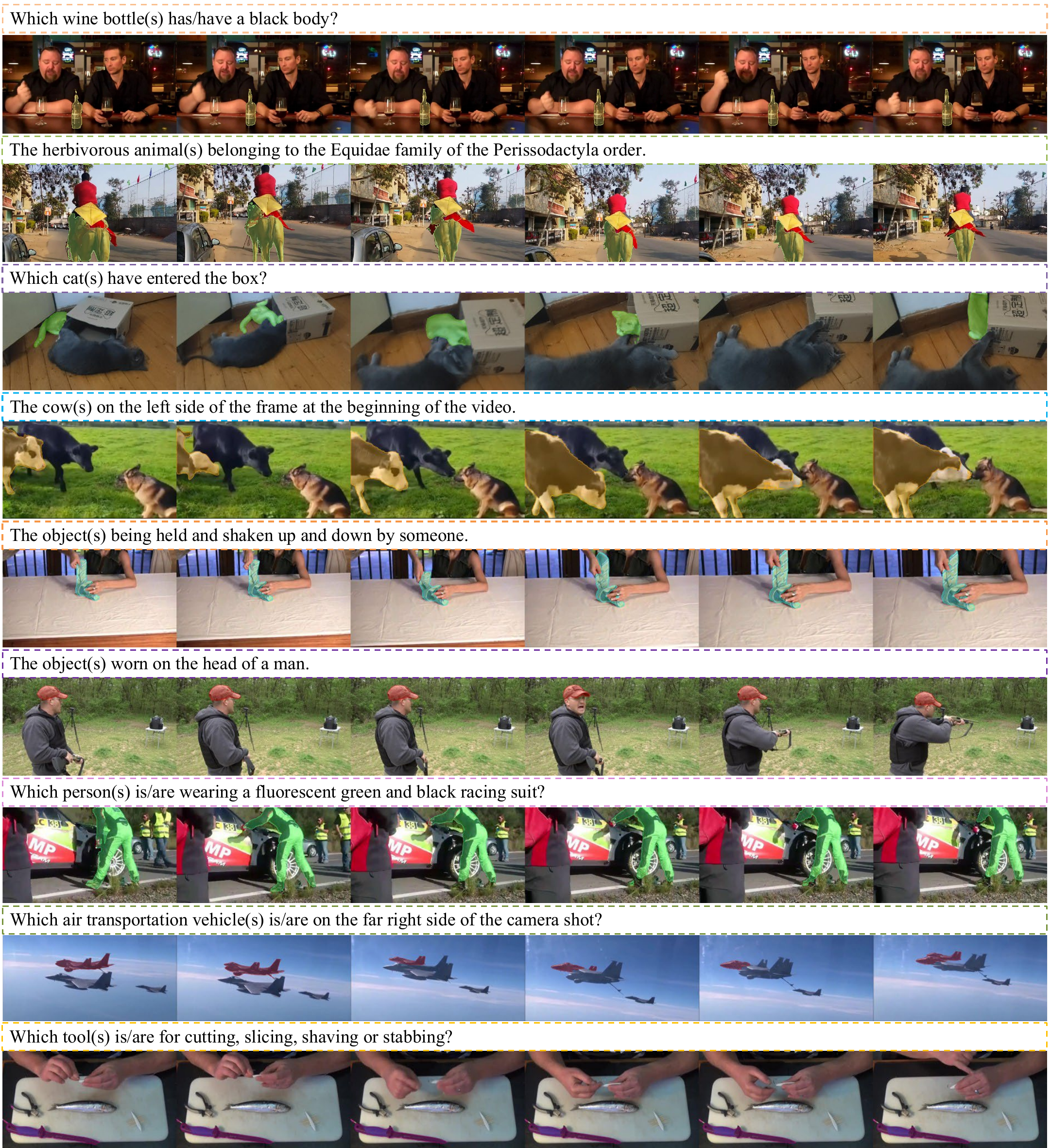}
\caption{\textbf{Qualitative Results of Referring and Reasoning Video Object Segmentation.}}
\label{fig:video_seg_vis}
\end{figure}

Fig. \ref{fig:video_seg_vis} illustrates qualitative results for Referring and Reasoning Video Object Segmentation on ReVOS \cite{yan2024visa}.

\begin{figure}[!ht]
\centering
\includegraphics[width=1.0\linewidth]{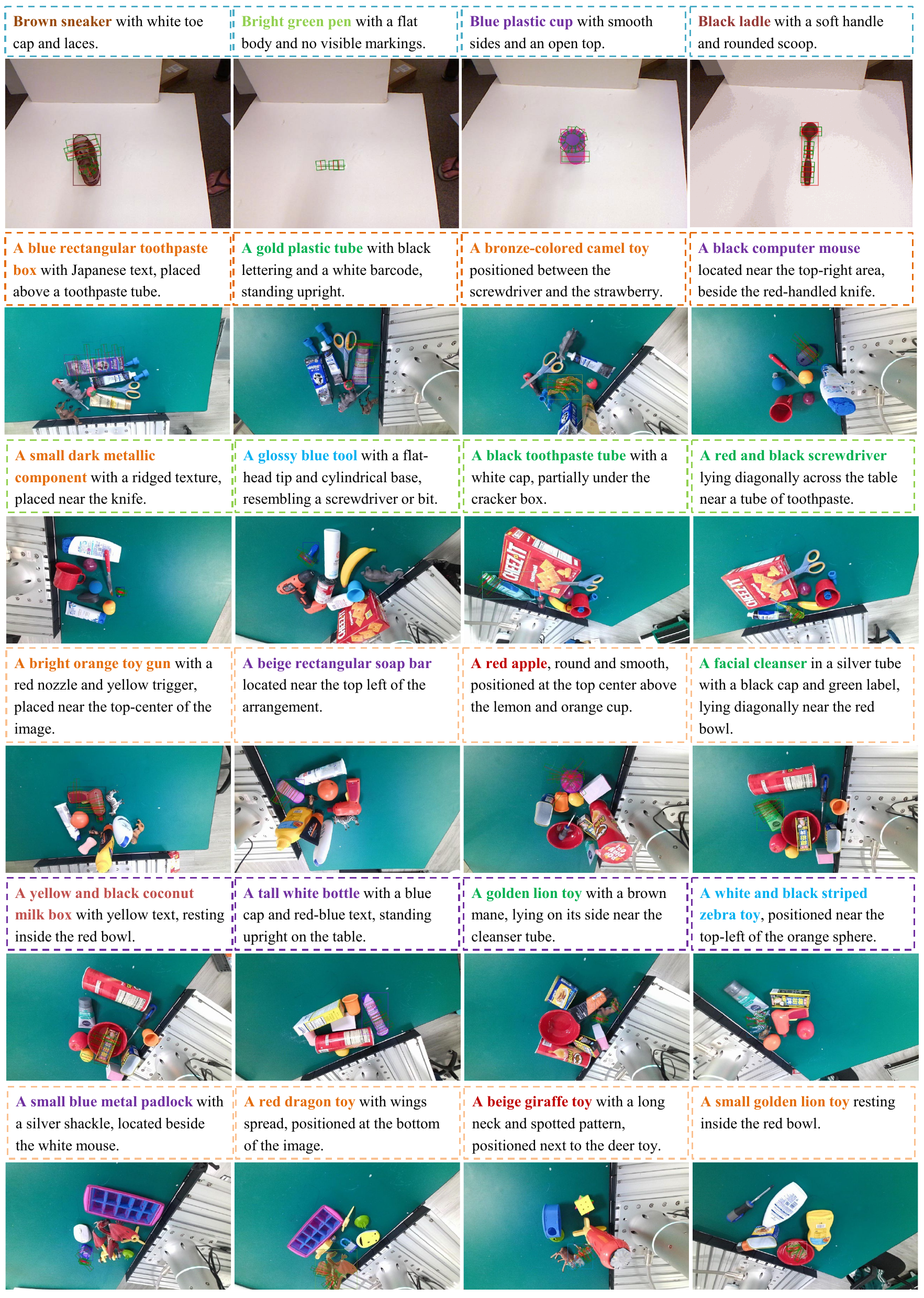}
\caption{\textbf{Qualitative Results of Actionable Perception on RealVLG-Benchmark} \cite{li2026realvlg}. }
\label{fig:realvlg_img_vis}
\vspace{-5mm}
\end{figure}

Fig. \ref{fig:realvlg_img_vis} presents qualitative results of UniRAP’s actionable perception on the RealVLG-Benchmark \cite{li2026realvlg}.

\begin{figure}[!ht]
\centering
\includegraphics[width=1.0\linewidth]{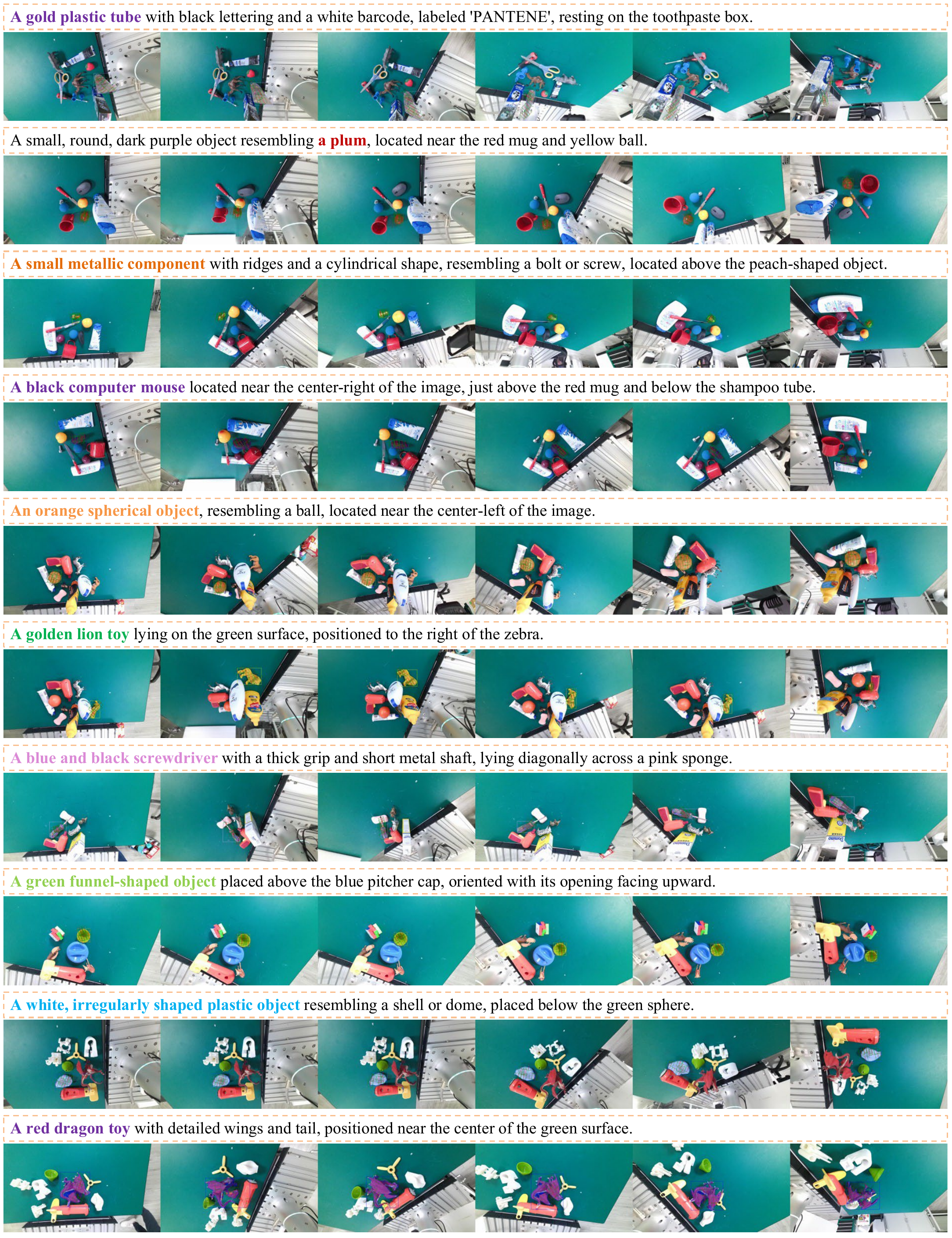}
\caption{\textbf{Qualitative Results of Spatio-Temporal Actionable Perception.}
To evaluate the temporal consistency of UniRAP’s actionable perception, we construct pseudo-sequences by randomly sampling 6 frames from each scene in the RealVLG-GraspNet subset, due to the absence of video-based grasping datasets.
}
\label{fig:realvlg_video_vis}
\vspace{-5mm}
\end{figure}

To further evaluate spatio-temporal consistency, we construct pseudo video clips by randomly sampling 6 frames from each scene in the RealVLG-GraspNet subset and arranging them into short sequences. As shown in Fig. \ref{fig:realvlg_video_vis}, UniRAP maintains consistent actionable predictions across frames under variations in viewpoint, occlusion, and scene configuration.

\section{Transferring Actionable Perception to Robotic Manipulation}
\subsection{Motivation}
Recent Vision–Language–Action (VLA) systems have achieved substantial progress in robotic manipulation tasks. However, these approaches still exhibit notable limitations in fine-grained visual localization required for precise interaction, as well as in their ability to generalize across complex scenes, object variations, and diverse language instructions \cite{sapkota2025vlasurvey, zhou2025exploringvla}. In particular, establishing pixel-level correspondences between natural language commands and actionable regions (e.g., grasp points or interaction surfaces) remains challenging. Existing methods often demonstrate instability and insufficient reliability, especially in cluttered or contact-dense environments.

Although perception–action decoupling strategies may not achieve the theoretical upper bound performance of end-to-end approaches, they offer a key advantage in leveraging the generalization capability of perception models. This enables more robust transfer across variations in environments, embodiments, camera configurations, and linguistic instructions. Drawing inspiration from the development trajectory of autonomous driving systems such as Tesla FSD, we argue that a more feasible short-term pathway is to first deploy decoupled solutions in relatively simple scenarios, progressively accumulate embodiment-specific data, and subsequently transition toward end-to-end training in a staged manner.

Accordingly, this work focuses on modeling executable vision–language localization as an intermediate representation for robotic manipulation. Specifically, rather than directly learning an end-to-end control policy, we generate reliable pixel-level or pose-level interaction cues and integrate them with classical geometric planning modules. This hybrid formulation enhances both the robustness and interpretability of robotic execution in complex environments.

\subsection{UniRAP for Reasoning-Aware Robotic Grasping}
\label{appendix_realworld_grasping}
UniRAP can be integrated into robotic grasping pipelines from two complementary perspectives. First, its coarse-grained perceptual outputs, such as bounding boxes or segmentation masks, can be directly combined with geometry-based grasp synthesis methods (e.g., GraspNet \cite{fang2020graspnet}). In this setting, UniRAP serves as a semantic front-end that filters task-relevant regions, while downstream geometric planners operate within these regions to generate feasible grasp candidates. This hybrid design preserves the robustness and efficiency of classical pipelines while introducing task-level semantic awareness.

\begin{figure}[!ht]
\centering
\includegraphics[width=0.9\linewidth]{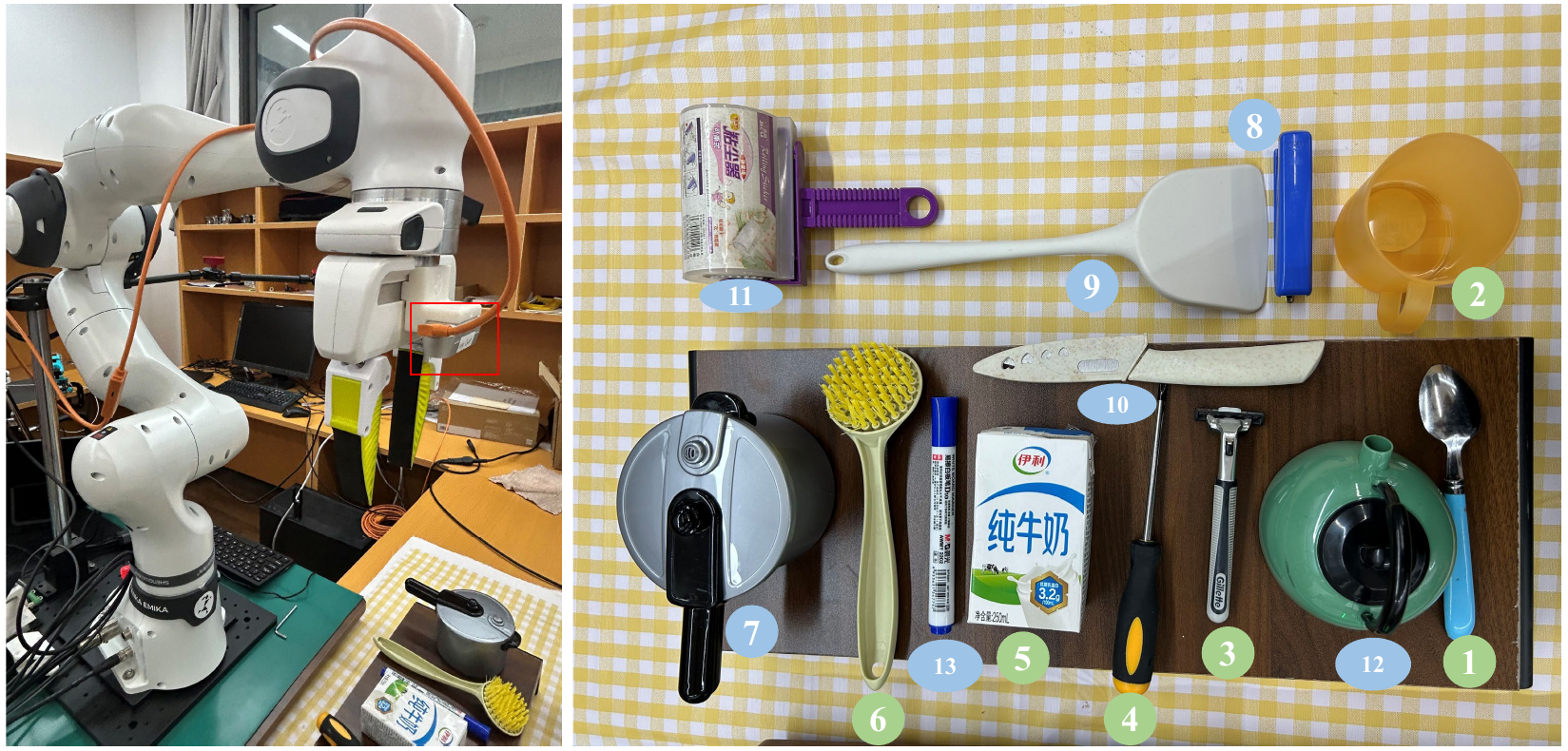}
\caption{\textbf{Real-World Experimental Setup.} A Franka FR3 robotic arm is equipped with an eye-in-hand RealSense D435i camera mounted on the end-effector. The test set consists of 13 common household objects, where objects 1–6 are used to evaluate referring grasping performance, and objects 7–13 are used to assess reasoning-based grasping performance.}
\label{fig:robot_setting}
\vspace{-10px}
\end{figure}

\begin{table}[t]
\centering
\caption{\textbf{Object Prompts for Real-World Robotic Vision-Language Grasping.}}
\resizebox{\linewidth}{!}{
\begin{tabular}{c|c|l}
\hline
Object ID & Object Name & Prompt \\
\hline
\multicolumn{3}{c}{\textbf{Referring}} \\
\hline
1 & Blue Spoon & Please give me the blue spoon. \\
2 & Orange Cup & Please grasp the orange cup. \\
3 & Razor & Please pick up the grey and black razor. \\
4 & Screwdriver & Please give me the screwdriver. \\
5 & Milk Box & Please give me the milk box. \\
6 & Green Brush & Please grasp the green brush. \\
\hline
\multicolumn{3}{c}{\textbf{Reasoning}} \\
\hline
7 & Pressure Cooker & I want to braise pork ribs. Can you find a suitable kitchen utensil for me in the image? \\
8 & Blue Stapler & Please give me the blue tool on the induction cooker. \\
9 & Shovel & Please use the tool to stir-fry the vegetables in the pan. \\
10 & Knife & Please use the tool on the table to cut some fruit for me. \\
11 & Rolling Sticker & Please use the tool on the table that can roll and stick lint to help me deal with my clothes. \\
12 & Teapot & I want to make a cup of tea. Please help me find a container that can hold hot water and steep tea leaves. \\
13 & Marker Pen & Please find a pen that can write on a whiteboard and be erased. \\
\hline
\end{tabular}
}
\label{tab:robot_prompt}
\vspace{-10px}
\end{table}

\begin{table}[ht!]
\centering
\caption{\textbf{Grasp Success Rates (\%) on 13 Household Objects.}}
\resizebox{0.8\linewidth}{!}{
\begin{tabular}{c|cccccc|ccccccc|c}
\hline
Task & \multicolumn{6}{c|}{Referring} & \multicolumn{7}{c|}{Reasoning} & Avg. \\
\hline
ObjectID & 1 & 2 & 3 & 4 & 5 & 6 & 7 & 8 & 9 & 10 & 11 & 12 & 13 & -- \\
\hline
RAGNet \cite{wu2025ragnet} + GraspNet \cite{fang2020graspnet}
& 50 & 60 & 40 & 60 & 50 & 60 & 30
& 40 & 30 & 40 & 30 & 40 & 40 & 43.8 \\

UniRAP + GraspNet \cite{fang2020graspnet}
& 70 & 70 & 60 & 70 & 60 & 50 & 40
& 70 & 50 & 50 & 40 & 40 & 50 & 55.4 \\

RealVLG-R1 \cite{li2026realvlg} & 80 & 60 & 70 & 80 & 70 & 70 & 50 & 80 & 40 & 70 & 40 & 20 & 80 & 62.3 \\
\rowcolor{orange!8.5} \textbf{UniRAP} (Ours)  & \textbf{100} & \textbf{90} & \textbf{80} & \textbf{100} & \textbf{90} & \textbf{100} & \textbf{80} & \textbf{80} & \textbf{70} & \textbf{90} & \textbf{80} & \textbf{90} & \textbf{100} & \textbf{88.5} \\
\hline
\end{tabular}
}
\label{tab:realworld_compare}
\vspace{-10px}
\end{table}

Second, UniRAP produces interaction-aware predictions in the form of 4-DoF grasp poses, which encode both spatial location and orientation constraints aligned with the inferred contact intent. These poses can be directly mapped into the robot coordinate system via standard hand–eye calibration procedures, enabling execution without requiring additional policy learning. Compared to conventional grasp detection outputs that are often agnostic to task semantics, UniRAP’s predictions are conditioned on both visual context and language-driven intent, thereby reducing ambiguity in grasp selection.

To evaluate the effectiveness of the contact poses predicted by UniRAP in real-world vision–language robotic grasping tasks, we adopt a pose-based execution protocol. Specifically, the predicted 4-DoF grasp poses are transformed into the robot base frame using hand–eye calibration parameters and subsequently executed directly by the robotic manipulator. As illustrated in Fig. \ref{fig:robot_setting}, the experimental platform is built upon a Franka FR3 robotic arm equipped with a RealSense D435i eye-in-hand camera mounted on the end-effector. The test set consists of 13 common household objects across different scenarios, with their names and corresponding prompts used in the experiments listed in Tab. \ref{tab:robot_prompt}.

As shown in Tab. \ref{tab:realworld_compare}, we further evaluate the performance of UniRAP and the baseline method RealVLG-R1 (3B) \cite{li2026realvlg} on two types of tasks, namely referring and reasoning language instructions, across 13 test objects, with 10 independent trials per object. RealVLG-R1 directly predicts two contact points to generate grasp poses. Experimental results show that UniRAP achieves a 26.2\% higher overall grasp success rate compared to RealVLG-R1, and its performance advantage is particularly pronounced when handling reasoning-based prompts.

\begin{figure}[!ht]
\centering
\includegraphics[width=1.0\linewidth]{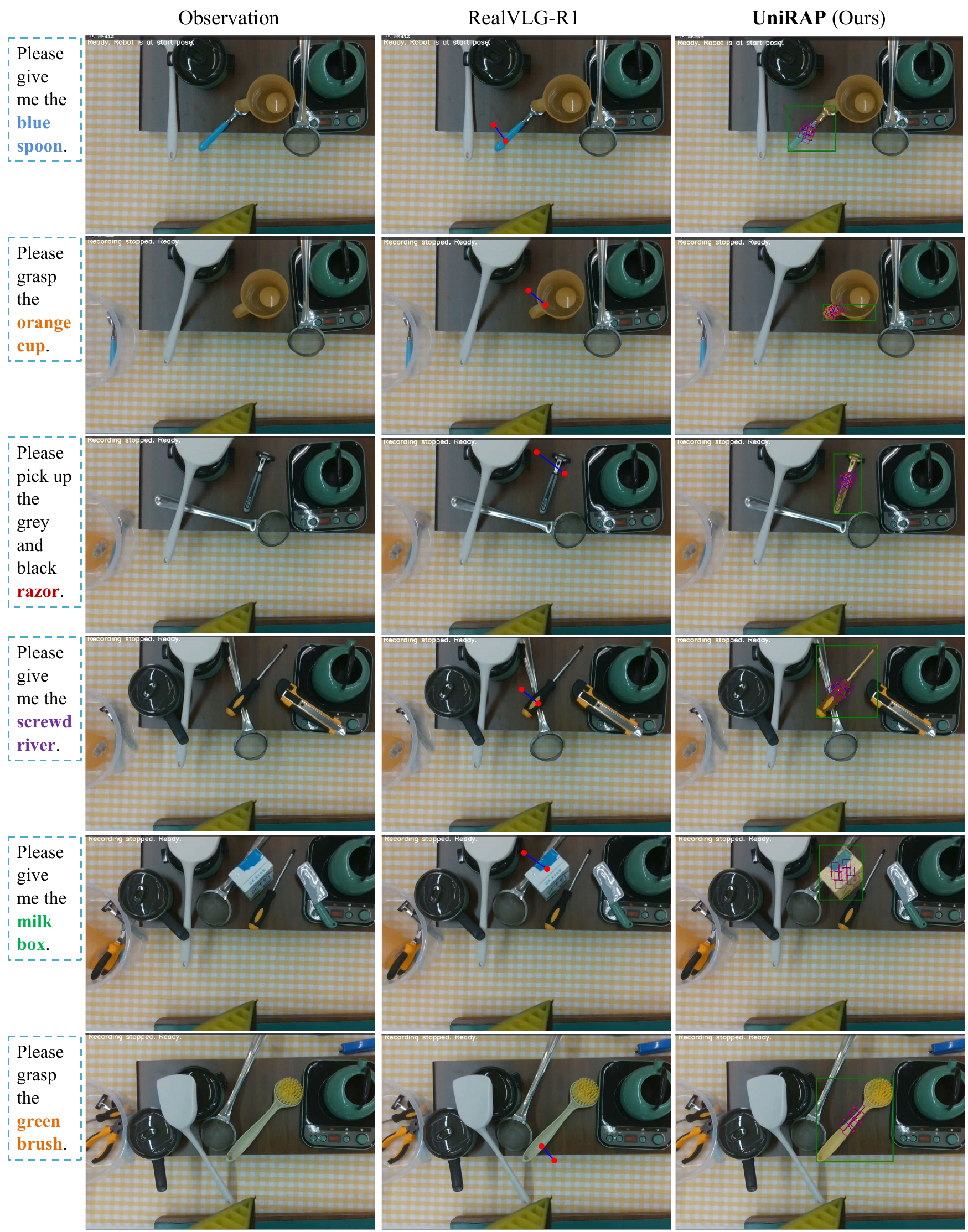}
\caption{\textbf{Qualitative Results of Real-World Referring Vision-Language Grasping.}
}
\label{fig:robot_realworld_s1}
\vspace{-5mm}
\end{figure}

\begin{figure}[!ht]
\centering
\includegraphics[width=1.0\linewidth]{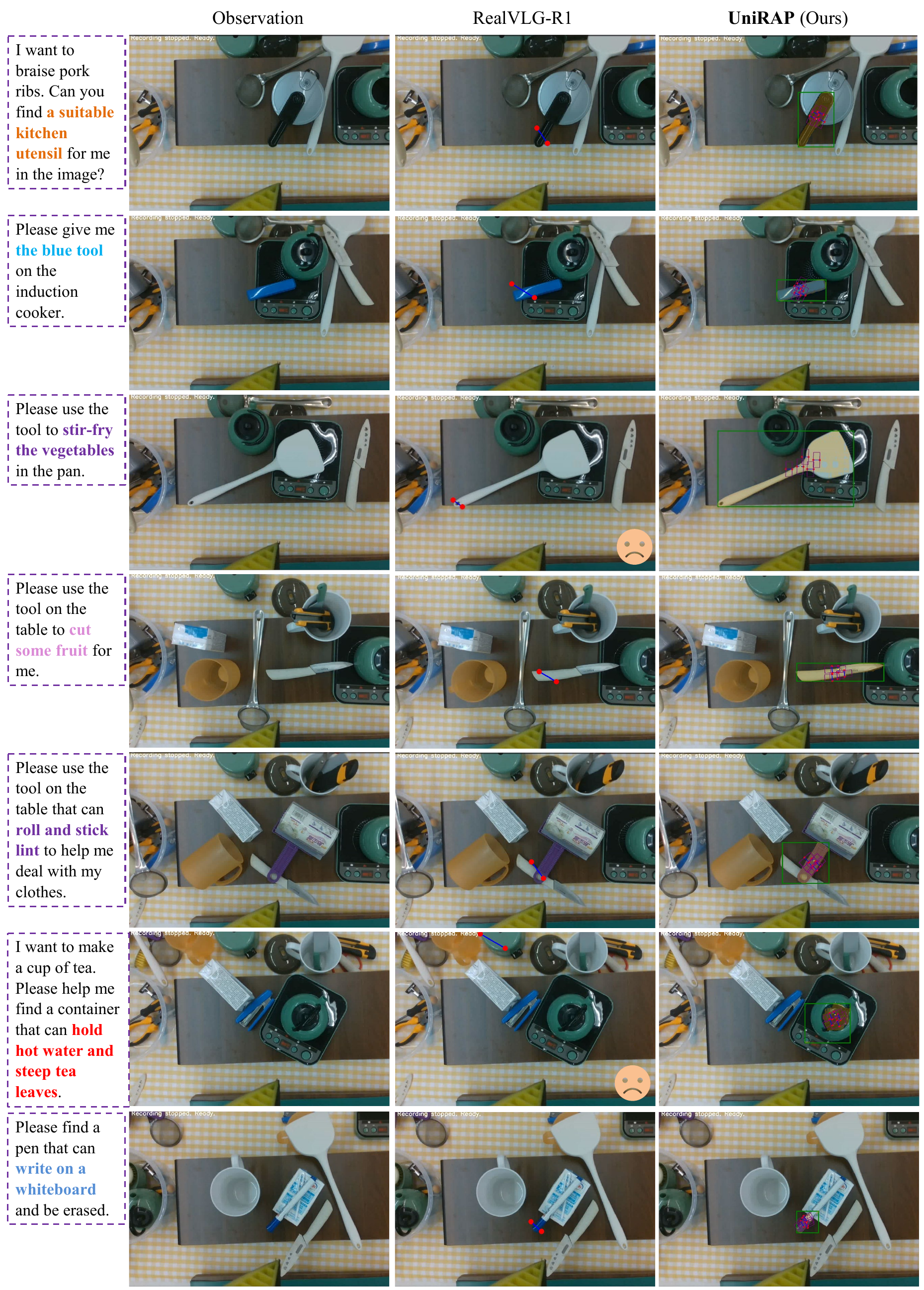}
\caption{\textbf{Qualitative Results of Real-World Reasoning-Based Vision-Language  Grasping.} Compared to RealVLG-R1, UniRAP exhibits a stronger affordance modeling capability, enabling it to predict more physically plausible grasp poses, such as correctly identifying appropriate grasp regions on objects like a \texttt{Shovel}. In addition, UniRAP demonstrates more robust localization performance. For instance, RealVLG-R1 fails to accurately localize the \texttt{Teapot}, whereas UniRAP successfully identifies a valid interaction region and produces a stable grasp prediction.
}
\label{fig:robot_realworld_s2}
\vspace{-5mm}
\end{figure}

In addition, qualitative results are presented in Fig. \ref{fig:robot_realworld_s1} (Referring) and Fig. \ref{fig:robot_realworld_s2} (Reasoning), further demonstrating the robustness and generalization capability of UniRAP across diverse scenarios. \textbf{Video demonstrations of the experiments are provided at \color{magenta}\url{https://lif314.github.io/projects/unirap}.}

\section{Discussions}
\subsection{Limitations \& Future Work}
\label{appendix_limitation}
Due to computational constraints, we were unable to further scale the training data or increase the model size, which in turn limited potential performance gains. Although incorporating pixel-level visual question answering tasks, such as PixelQA \cite{liu2025unipixel}, may offer additional benefits, these were not included within the scope of this study.

Given that this work focuses on perception models for embodied scenarios, we argue that it is important to investigate the capability boundaries of smaller models, with an emphasis on achieving real-time actionable perception. Such an approach facilitates deployment on edge devices and reduces the overall computational demands of real-world systems.

\subsection{Potential Societal Impacts}
\label{appendix_impact}
This work presents a pixel-level vision–language perception framework tailored for embodied scenarios, enabling a unified formulation that spans coarse-grained object detection and segmentation, functional segmentation, and physical interaction pose estimation. The proposed approach is particularly relevant to application domains requiring frequent and adaptive interaction between agents and their environments, such as robotic home assistance, security, and healthcare. These settings typically demand fine-grained understanding of images and videos, as well as the capability to reason about physical interactions. Moreover, in other embodied applications involving multimodal intelligent agents, the proposed framework can serve as a more expressive and practical alternative.

To the best of our knowledge, this work does not raise any specific potential negative societal impacts that require explicit disclosure.

\section{Licenses}

Our UniRAP is built based on the pre-trained Qwen2.5-VL \cite{Qwen2.5-VL} and SAM2 \cite{ravi2024sam2} models. They are both licensed under the Apache License 2.0 (\url{https://www.apache.org/licenses/LICENSE-2.0}).

%% file: checklist.tex
\clearpage
\section*{NeurIPS Paper Checklist}

The checklist is designed to encourage best practices for responsible machine learning research, addressing issues of reproducibility, transparency, research ethics, and societal impact. Do not remove the checklist: {\bf The papers not including the checklist will be desk rejected.} The checklist should follow the references and follow the (optional) supplemental material.  The checklist does NOT count towards the page
limit. 

Please read the checklist guidelines carefully for information on how to answer these questions. For each question in the checklist:
\begin{itemize}
    \item You should answer \answerYes{}, \answerNo{}, or \answerNA{}.
    \item \answerNA{} means either that the question is Not Applicable for that particular paper or the relevant information is Not Available.
    \item Please provide a short (1--2 sentence) justification right after your answer (even for \answerNA). 
\end{itemize}

{\bf The checklist answers are an integral part of your paper submission.} They are visible to the reviewers, area chairs, senior area chairs, and ethics reviewers. You will also be asked to include it (after eventual revisions) with the final version of your paper, and its final version will be published with the paper.

The reviewers of your paper will be asked to use the checklist as one of the factors in their evaluation. While \answerYes{} is generally preferable to \answerNo{}, it is perfectly acceptable to answer \answerNo{} provided a proper justification is given (e.g., error bars are not reported because it would be too computationally expensive'' or ``we were unable to find the license for the dataset we used''). In general, answering \answerNo{} or \answerNA{} is not grounds for rejection. While the questions are phrased in a binary way, we acknowledge that the true answer is often more nuanced, so please just use your best judgment and write a justification to elaborate. All supporting evidence can appear either in the main paper or the supplemental material, provided in appendix. If you answer \answerYes{} to a question, in the justification please point to the section(s) where related material for the question can be found.

IMPORTANT, please:
\begin{itemize}
    \item {\bf Delete this instruction block, but keep the section heading ``NeurIPS Paper Checklist"},
    \item  {\bf Keep the checklist subsection headings, questions/answers and guidelines below.}
    \item {\bf Do not modify the questions and only use the provided macros for your answers}.
\end{itemize}


\begin{enumerate}

\item {\bf Claims}
    \item[] Question: Do the main claims made in the abstract and introduction accurately reflect the paper's contributions and scope?
    \item[] Answer: \answerYes{} 
    \item[] Justification:  The abstract and introduction outline the key challenges addressed, our contributions, and a summary of the experimental findings.

\item {\bf Limitations}
    \item[] Question: Does the paper discuss the limitations of the work performed by the authors?
    \item[] Answer: \answerYes{} 
    \item[] Justification:  We include the limitations of our work in Appendix \ref{appendix_limitation}.

\item {\bf Theory assumptions and proofs}
    \item[] Question: For each theoretical result, does the paper provide the full set of assumptions and a complete (and correct) proof?
    \item[] Answer: \answerNA{} 
    \item[] Justification: This is not a theoretical paper.

    \item {\bf Experimental result reproducibility}
    \item[] Question: Does the paper fully disclose all the information needed to reproduce the main experimental results of the paper to the extent that it affects the main claims and/or conclusions of the paper (regardless of whether the code and data are provided or not)?
    \item[] Answer: \answerYes{} 
    \item[] Justification: We have described the experimental details and have properly cited the opensource assets and materials necessary for reproducing the experiments. The code will be made publicly available upon acceptance.

\item {\bf Open access to data and code}
    \item[] Question: Does the paper provide open access to the data and code, with sufficient instructions to faithfully reproduce the main experimental results, as described in supplemental material?
    \item[] Answer: \answerYes{} 
    \item[] Justification: We used the public dataset, and we plan to publicly release our code once the status of the paper is confirmed.

\item {\bf Experimental setting/details}
    \item[] Question: Does the paper specify all the training and test details (e.g., data splits, hyperparameters, how they were chosen, type of optimizer) necessary to understand the results?
    \item[] Answer: \answerYes{} 
    \item[] Justification: Details are summarized in Appendix \ref{appendix_experiments}.

\item {\bf Experiment statistical significance}
    \item[] Question: Does the paper report error bars suitably and correctly defined or other appropriate information about the statistical significance of the experiments?
    \item[] Answer: \answerNo{} 
    \item[] Justification: Due to the limited computational resources, we conducted each experiment only once.

\item {\bf Experiments compute resources}
    \item[] Question: For each experiment, does the paper provide sufficient information on the computer resources (type of compute workers, memory, time of execution) needed to reproduce the experiments?
    \item[] Answer: \answerYes{} 
    \item[] Justification:  We have a Appendix \ref{appendix_iImplementation} on this.
    
\item {\bf Code of ethics}
    \item[] Question: Does the research conducted in the paper conform, in every respect, with the NeurIPS Code of Ethics \url{https://neurips.cc/public/EthicsGuidelines}?
    \item[] Answer: \answerYes{} 
    \item[] Justification: We fully adhere to the NeurIPS Code of Ethics.

\item {\bf Broader impacts}
    \item[] Question: Does the paper discuss both potential positive societal impacts and negative societal impacts of the work performed?
    \item[] Answer: \answerYes{} 
    \item[] Justification: See Appendix \ref{appendix_impact}.
    
\item {\bf Safeguards}
    \item[] Question: Does the paper describe safeguards that have been put in place for responsible release of data or models that have a high risk for misuse (e.g., pre-trained language models, image generators, or scraped datasets)?
    \item[] Answer: \answerNA{} 
    \item[] Justification: The paper poses no such risks.

\item {\bf Licenses for existing assets}
    \item[] Question: Are the creators or original owners of assets (e.g., code, data, models), used in the paper, properly credited and are the license and terms of use explicitly mentioned and properly respected?
    \item[] Answer: \answerYes{} 
    \item[] Justification: The paper properly credits the creators or original owners of the assets used, such as code, data, and models. We also have correctly cited the relevant literature.

\item {\bf New assets}
    \item[] Question: Are new assets introduced in the paper well documented and is the documentation provided alongside the assets?
    \item[] Answer: \answerNA{} 
    \item[] Justification:  We will disclose the code after the confirmation of the status of the paper, but currently no new assets is available.

\item {\bf Crowdsourcing and research with human subjects}
    \item[] Question: For crowdsourcing experiments and research with human subjects, does the paper include the full text of instructions given to participants and screenshots, if applicable, as well as details about compensation (if any)? 
    \item[] Answer: \answerNA{} 
    \item[] Justification:  The paper does not involve crowdsourcing nor research with human subjects.

\item {\bf Institutional review board (IRB) approvals or equivalent for research with human subjects}
    \item[] Question: Does the paper describe potential risks incurred by study participants, whether such risks were disclosed to the subjects, and whether Institutional Review Board (IRB) approvals (or an equivalent approval/review based on the requirements of your country or institution) were obtained?
    \item[] Answer: \answerNA{} 
    \item[] Justification: The paper does not involve crowdsourcing nor research with human subjects.

\item {\bf Declaration of LLM usage}
    \item[] Question: Does the paper describe the usage of LLMs if it is an important, original, or non-standard component of the core methods in this research? Note that if the LLM is used only for writing, editing, or formatting purposes and does \emph{not} impact the core methodology, scientific rigor, or originality of the research, declaration is not required.
    \item[] Answer: \answerNA{} 
    \item[] Justification: The core method development in this research does not involve LLMs as any important, original, or non-standard components.

\end{enumerate}

%% file: main.bib
@String(IJCV = {Int. J. Comput. Vis.})

@String(CVPR= {IEEE Conf. Comput. Vis. Pattern Recog.})

@String(ICCV= {Int. Conf. Comput. Vis.})

@String(ECCV= {Eur. Conf. Comput. Vis.})

@String(NIPS= {Adv. Neural Inform. Process. Syst.})

@String(ICLR = {Int. Conf. Learn. Represent.})

@String(AAAI = {AAAI})

@String(IJCV  = {IJCV})

@String(CVPR  = {CVPR})

@String(ICCV  = {ICCV})

@String(ECCV  = {ECCV})

@String(NIPS  = {NeurIPS})

@String(ICML  = {ICML})

@String(ICLR  = {ICLR})

@String(EMNLP = {EMNLP})

@String(RAL = {RA-L})

@String(IROS = {IROS})

@String(ICRA = {ICRA})

@String(CoRL = {CoRL})

@article{Vinyals2014ShowAT,
  title={{Show and Tell: A Neural Image Caption Generator}},
  author={Oriol Vinyals and Alexander Toshev and Samy Bengio and D. Erhan},
  journal=CVPR,
  year={2014},
  pages={3156-3164},
}

@article{Agrawal2015VQAVQ,
  title={{VQA: Visual Question Answering}},
  author={Aishwarya Agrawal and Jiasen Lu and Stanislaw Antol and Margaret Mitchell and C. Lawrence Zitnick and Devi Parikh and Dhruv Batra},
  journal=IJCV,
  year={2015},
  volume={123},
  pages={4-31},
}

@article{Beyer2024PaliGemmaAV,
  title={{PaliGemma: A versatile 3B VLM for transfer}},
  author={Lucas Beyer and Andreas Steiner and Andr{\'e} Susano Pinto and Alexander Kolesnikov and Xiao Wang and Daniel M. Salz and Maxim Neumann and Ibrahim M. Alabdulmohsin and Michael Tschannen and Emanuele Bugliarello and Thomas Unterthiner and Daniel Keysers and Skanda Koppula and Fangyu Liu and Adam Grycner and Alexey A. Gritsenko and Neil Houlsby and Manoj Kumar and Keran Rong and Julian Martin Eisenschlos and Rishabh Kabra and Matthias Bauer and Matko Bovsnjak and Xi Chen and Matthias Minderer and Paul Voigtlaender and Ioana Bica and Ivana Balazevic and Joan Puigcerver and Pinelopi Papalampidi and Olivier H{\'e}naff and Xi Xiong and Radu Soricut and Jeremiah Harmsen and Xiao-Qi Zhai},
  journal={ArXiv},
  year={2024},
  volume={abs/2407.07726},
}

@inproceedings{chen2024internvl,
  title={{InternVL: Scaling up Vision Foundation Models and Aligning for Generic Visual-Linguistic Tasks}},
  author={Chen, Zhe and Wu, Jiannan and Wang, Wenhai and Su, Weijie and Chen, Guo and Xing, Sen and Zhong, Muyan and Zhang, Qinglong and Zhu, Xizhou and Lu, Lewei and others},
  booktitle=CVPR,
  pages={24185--24198},
  year={2024}
}

@article{Chen2023ShikraUM,
  title={{Shikra: Unleashing Multimodal LLM's Referential Dialogue Magic}},
  author={Ke Chen and Zhao Zhang and Weili Zeng and Richong Zhang and Feng Zhu and Rui Zhao},
  journal={ArXiv},
  year={2023},
  volume={abs/2306.15195},
}

@inproceedings{cogvlm,
author = {Wang, Weihan and Lv, Qingsong and Yu, Wenmeng and Hong, Wenyi and Qi, Ji and Wang, Yan and Ji, Junhui and Yang, Zhuoyi and Zhao, Lei and Song, Xixuan and Xu, Jiazheng and Chen, Keqin and Xu, Bin and Li, Juanzi and Dong, Yuxiao and Ding, Ming and Tang, Jie},
title = {{CogVLM: Visual Expert for Pretrained Language Models}},
year = {2024},
booktitle =NIPS,
}

@article{You2023FerretRA,
  title={{Ferret: Refer and Ground Anything Anywhere at Any Granularity}},
  author={Haoxuan You and Haotian Zhang and Zhe Gan and Xianzhi Du and Bowen Zhang and Zirui Wang and Liangliang Cao and Shih-Fu Chang and Yinfei Yang},
  journal={ArXiv},
  year={2023},
  volume={abs/2310.07704},
}

@inproceedings{liu2025visualrft,
  title={{Visual-RFT: Visual Reinforcement Fine-Tuning}},
  author={Liu, Ziyu and Sun, Zeyi and Zang, Yuhang and Dong, Xiaoyi and Cao, Yuhang and Duan, Haodong and Lin, Dahua and Wang, Jiaqi},
  booktitle=ICCV,
  pages={2034--2044},
  year={2025}
}

@inproceedings{ma2024groma,
  title={{Groma: Localized Visual Tokenization for Grounding Multimodal Large Language Models}},
  author={Ma, Chuofan and Jiang, Yi and Wu, Jiannan and Yuan, Zehuan and Qi, Xiaojuan},
  booktitle=ECCV,
  pages={417--435},
  year={2024},
}

@article{zhang2024ferretv2,
  title={{Ferret-v2: An Improved Baseline for Referring and Grounding with Large Language Models}},
  author={Zhang, Haotian and You, Haoxuan and Dufter, Philipp and Zhang, Bowen and Chen, Chen and Chen, Hong-You and Fu, Tsu-Jui and Wang, William Yang and Chang, Shih-Fu and Gan, Zhe and others},
  journal={arXiv preprint arXiv:2404.07973},
  year={2024}
}

@inproceedings{jiang2025RexOmni,
  title={{Detect Anything via Next Point Prediction}},
  author={Jiang, Qing and Huo, Junan and Chen, Xingyu and Xiong, Yuda and Zeng, Zhaoyang and Chen, Yihao and Ren, Tianhe and Yu, Junzhi and Zhang, Lei},
  booktitle=CVPR,
  year={2026}
}

@article{sarch2025regrounded,
  title={Grounded reinforcement learning for visual reasoning},
  author={Sarch, Gabriel and Saha, Snigdha and Khandelwal, Naitik and Jain, Ayush and Tarr, Michael J and Kumar, Aviral and Fragkiadaki, Katerina},
  journal={arXiv preprint arXiv:2505.23678},
  year={2025}
}

@inproceedings{nips2023visionllm,
author = {Wang, Wenhai and Chen, Zhe and Chen, Xiaokang and Wu, Jiannan and Zhu, Xizhou and Zeng, Gang and Luo, Ping and Lu, Tong and Zhou, Jie and Qiao, Yu and Dai, Jifeng},
title = {{VisionLLM: Large Language Model is also an Open-Ended Decoder for Vision-Centric Tasks}},
year = {2023},
booktitle = NIPS,
articleno = {2688},
numpages = {13},
}

@inproceedings{cao2025refer,
  title={{Refer to Any Segmentation Mask Group With Vision-Language Prompts}},
  author={Cao, Shengcao and Wei, Zijun and Kuen, Jason and Liu, Kangning and Zhang, Lingzhi and Gu, Jiuxiang and Jung, HyunJoon and Gui, Liang-Yan and Wang, Yu-Xiong},
  booktitle={ICCV},
  year={2025}
}

@inproceedings{xia2024gsva,
  title={{GSVA: Generalized Segmentation via Multimodal Large Language Models}},
  author={Xia, Zhuofan and Han, Dongchen and Han, Yizeng and Pan, Xuran and Song, Shiji and Huang, Gao},
  booktitle=CVPR,
  pages={3858--3869},
  year={2024}
}

@inproceedings{zhang2024psalm,
  title={{PSALM: Pixelwise SegmentAtion with Large Multi-Modal Model}},
  author={Zhang, Zheng and Ma, Yeyao and Zhang, Enming and Bai, Xiang},
  booktitle=ECCV,
  pages={74--91},
  year={2024},
}

@inproceedings{li2026realvlg,
  title     = {{RealVLG-R1: A Large-Scale Real-World Visual-Language Grounding Benchmark for Robotic Perception and Manipulation}},
  author    = {Li, Linfei and Zhang, Lin and Shen, Ying},
  booktitle = CVPR,
  year      = {2026},
}

@inproceedings{huang2024manipvqa,
  title={{ManipVQA: Injecting Robotic Affordance and Physically Grounded Information into Multi-Modal Large Language Models}},
  author={Huang, Siyuan and Ponomarenko, Iaroslav and Jiang, Zhengkai and Li, Xiaoqi and Hu, Xiaobin and Gao, Peng and Li, Hongsheng and Dong, Hao},
  booktitle=IROS,
  pages={7580--7587},
  year={2024},
}

@inproceedings{kim2025manipgpt,
  title={{ManipGPT: Is Affordance Segmentation by Large Vision Models Enough for Articulated Object Manipulation?}},
  author={Kim, Taewhan and Bae, Hojin and Li, Zeming and Li, Xiaoqi and Ponomarenko, Iaroslav and Wu, Ruihai and Dong, Hao},
  booktitle=IROS,
  pages={20974--20981},
  year={2025},
}

@inproceedings{li2024ooal,
  title={{One-Shot Open Affordance Learning with Foundation Models}},
  author={Li, Gen and Sun, Deqing and Sevilla-Lara, Laura and Jampani, Varun},
  booktitle=CVPR,
  pages={3086--3096},
  year={2024}
}

@inproceedings{li2024manipllm,
  title={{ManipLLM: Embodied Multimodal Large Language Model for Object-Centric Robotic Manipulation}},
  author={Li, Xiaoqi and Zhang, Mingxu and Geng, Yiran and Geng, Haoran and Long, Yuxing and Shen, Yan and Zhang, Renrui and Liu, Jiaming and Dong, Hao},
  booktitle=CVPR,
  pages={18061--18070},
  year={2024}
}

@article{xu2024naturalvlm,
  title={{NaturalVLM: Leveraging Fine-grained Natural Language for Affordance-Guided Visual Manipulation}},
  author={Xu, Ran and Shen, Yan and Li, Xiaoqi and Wu, Ruihai and Dong, Hao},
  journal=RAL,
  volume={9},
  number={12},
  pages={10842--10849},
  year={2024},
}

@article{fei2025liberoplus,
  title={{LIBERO-Plus: In-depth Robustness Analysis of Vision-Language-Action Models}},
  author={Fei, Senyu and Wang, Siyin and Shi, Junhao and Dai, Zihao and Cai, Jikun and Qian, Pengfang and Ji, Li and He, Xinzhe and Zhang, Shiduo and Fei, Zhaoye and others},
  journal={arXiv preprint arXiv:2510.13626},
  year={2025}
}

@inproceedings{zhou2025exploringvla,
    title={Exploring the Limits of Vision-Language-Action Manipulation in Cross-task Generalization},
    author={Jiaming Zhou and Ke Ye and Jiayi Liu and Teli Ma and Zifan Wang and Ronghe Qiu and Kun-Yu Lin and Zhilin Zhao and Junwei Liang},
    booktitle=NIPS,
    year={2025},
}

@article{krishna2017visualgenome,
  title={{Visual Genome: Connecting Language and Vision Using Crowdsourced Dense Image Annotations}},
  author={Krishna, Ranjay and Zhu, Yuke and Groth, Oliver and Johnson, Justin and Hata, Kenji and Kravitz, Joshua and Chen, Stephanie and Kalantidis, Yannis and Li, Li-Jia and Shamma, David A and others},
  journal=IJCV,
  volume={123},
  number={1},
  pages={32--73},
  year={2017},
}

@article{driess2023palme,
  title={{PaLM-E: An Embodied Multimodal Language Model}},
  author={Driess, Danny and Xia, Fei and Sajjadi, Mehdi SM and Lynch, Corey and Chowdhery, Aakanksha and Ichter, Brian and Wahid, Ayzaan and Tompson, Jonathan and Vuong, Quan and Yu, Tianhe and others},
  journal={arXiv preprint arXiv:2303.03378},
  year={2023}
}

@inproceedings{do2018affordancenet,
  title={{AffordanceNet: An End-to-End Deep Learning Approach for Object Affordance Detection}},
  author={Do, Thanh-Toan and Nguyen, Anh and Reid, Ian},
  booktitle=ICRA,
  pages={5882--5889},
  year={2018},
}

@article{sapkota2025vlasurvey,
  title={{Vision-Language-Action (VLA) Models: Concepts, Progress, Applications and Challenges}},
  author={Sapkota, Ranjan and Cao, Yang and Roumeliotis, Konstantinos I and Karkee, Manoj},
  journal={arXiv preprint arXiv:2505.04769},
  year={2025}
}

@inproceedings{radford2021clip,
  title={{Learning Transferable Visual Models From Natural Language Supervision}},
  author={Radford, Alec and Kim, Jong Wook and Hallacy, Chris and Ramesh, Aditya and Goh, Gabriel and Agarwal, Sandhini and Sastry, Girish and Askell, Amanda and Mishkin, Pamela and Clark, Jack and others},
  booktitle=ICML,
  pages={8748--8763},
  year={2021},
}

@inproceedings{zhai2023siglip,
  title={{Sigmoid Loss for Language Image Pre-Training}},
  author={Zhai, Xiaohua and Mustafa, Basil and Kolesnikov, Alexander and Beyer, Lucas},
  booktitle=ICCV,
  pages={11975--11986},
  year={2023}
}

@inproceedings{caron202dino,
  title={{Emerging Properties in Self-Supervised Vision Transformers}},
  author={Caron, Mathilde and Touvron, Hugo and Misra, Ishan and J{\'e}gou, Herv{\'e} and Mairal, Julien and Bojanowski, Piotr and Joulin, Armand},
  booktitle=ICCV,
  pages={9650--9660},
  year={2021}
}

@article{oquab2023dinov2,
  title={{DINOv2: Learning Robust Visual Features without Supervision}},
  author={Oquab, Maxime and Darcet, Timoth{\'e}e and Moutakanni, Th{\'e}o and Vo, Huy and Szafraniec, Marc and Khalidov, Vasil and Fernandez, Pierre and Haziza, Daniel and Massa, Francisco and El-Nouby, Alaaeldin and others},
  journal={arXiv preprint arXiv:2304.07193},
  year={2023}
}

@article{sapkota2025yolo26,
  title={{YOLO26: Key Architectural Enhancements and Performance Benchmarking for Real-Time Object Detection}},
  author={Sapkota, Ranjan and Cheppally, Rahul Harsha and Sharda, Ajay and Karkee, Manoj},
  journal={arXiv preprint arXiv:2509.25164},
  year={2025}
}

@inproceedings{cheng2024yoloworld,
  title={{YOLO-World: Real-Time Open-Vocabulary Object Detection}},
  author={Cheng, Tianheng and Song, Lin and Ge, Yixiao and Liu, Wenyu and Wang, Xinggang and Shan, Ying},
  booktitle=CVPR,
  pages={16901--16911},
  year={2024}
}

@inproceedings{wang2025yoloe,
  title={{YOLOE: Real-Time Seeing Anything}},
  author={Wang, Ao and Liu, Lihao and Chen, Hui and Lin, Zijia and Han, Jungong and Ding, Guiguang},
  booktitle=ICCV,
  pages={24591--24602},
  year={2025}
}

@inproceedings{carion2020detr,
  title={{End-to-End Object Detection with Transformers}},
  author={Carion, Nicolas and Massa, Francisco and Synnaeve, Gabriel and Usunier, Nicolas and Kirillov, Alexander and Zagoruyko, Sergey},
  booktitle=ECCV,
  pages={213--229},
  year={2020},
}

@article{robinson2025rfdetr,
  title={{RF-DETR: Neural Architecture Search for Real-Time Detection Transformers}},
  author={Robinson, Isaac and Robicheaux, Peter and Popov, Matvei and Ramanan, Deva and Peri, Neehar},
  journal={arXiv preprint arXiv:2511.09554},
  year={2025}
}

@inproceedings{shridhar2022cliport,
  title={{CLIPort: What and Where Pathways for Robotic Manipulation}},
  author={Shridhar, Mohit and Manuelli, Lucas and Fox, Dieter},
  booktitle=CoRL,
  pages={894--906},
  year={2022},
}

@inproceedings{zhou2025roborefer,
  title={{RoboRefer: Towards Spatial Referring with Reasoning in Vision-Language Models for Robotics}},
  author={Zhou, Enshen and An, Jingkun and Chi, Cheng and Han, Yi and Rong, Shanyu and Zhang, Chi and Wang, Pengwei and Wang, Zhongyuan and Huang, Tiejun and Sheng, Lu and others},
  booktitle=NIPS,
  year={2025}
}

@InProceedings{yuan25robopoint,
  title = 	 {{RoboPoint: A Vision-Language Model for Spatial Affordance Prediction in Robotics}},
  author =       {Yuan, Wentao and Duan, Jiafei and Blukis, Valts and Pumacay, Wilbert and Krishna, Ranjay and Murali, Adithyavairavan and Mousavian, Arsalan and Fox, Dieter},
  booktitle =CoRL,
  pages = 	 {4005--4020},
  year = 	 {2025},
  volume = 	 {270},
}

@article{nasiriany2024pivot,
  title={{PIVOT: Iterative Visual Prompting Elicits Actionable Knowledge for VLMs}},
  author={Nasiriany, Soroush and Xia, Fei and Yu, Wenhao and Xiao, Ted and Liang, Jacky and Dasgupta, Ishita and Xie, Annie and Driess, Danny and Wahid, Ayzaan and Xu, Zhuo and others},
  journal={arXiv preprint arXiv:2402.07872},
  year={2024}
}

@InProceedings{Yang2025magma,
    author    = {Yang, Jianwei and Tan, Reuben and Wu, Qianhui and Zheng, Ruijie and Peng, Baolin and Liang, Yongyuan and Gu, Yu and Cai, Mu and Ye, Seonghyeon and Jang, Joel and Deng, Yuquan and Gao, Jianfeng},
    title     = {{Magma: A Foundation Model for Multimodal AI Agents}},
    booktitle = CVPR,
    year      = {2025},
    pages     = {14203-14214}
}

@InProceedings{yuan2026embodiedr1,
  title={{Embodied-R1: Reinforced Embodied Reasoning for General Robotic Manipulation}},
  author={Yuan, Yifu and Cui, Haiqin and Huang, Yaoting and Chen, Yibin and Ni, Fei and Dong, Zibin and Li, Pengyi and Zheng, Yan and Tang, Hongyao and Hao, Jianye},
  booktitle=ICLR,
  year={2026}
}

@inproceedings{chen2025robovlm,
title={{Robo2VLM: Improving Visual Question Answering using Large-Scale Robot Manipulation Data}},
author={Kaiyuan Chen and Shuangyu Xie and Zehan Ma and Pannag R Sanketi and Ken Goldberg},
booktitle=NIPS,
year={2025},
}

@InProceedings{huang25a3vlm,
  title = 	 {{A3VLM: Actionable Articulation-Aware Vision Language Model}},
  author =       {Huang, Siyuan and Chang, Haonan and Liu, Yuhan and Zhu, Yimeng and Dong, Hao and Boularias, Abdeslam and Gao, Peng and Li, Hongsheng},
  booktitle=CoRL,
  pages = 	 {1675--1690},
  year = 	 {2025},
  volume = 	 {270},
}

@inproceedings{qian2024affordancellm,
  title={{AffordanceLLM: Grounding Affordance from Vision Language Models}},
  author={Qian, Shengyi and Chen, Weifeng and Bai, Min and Zhou, Xiong and Tu, Zhuowen and Li, Li Erran},
  booktitle=CVPR,
  pages={7587--7597},
  year={2024}
}

@InProceedings{liu2023llava,
  title={{Visual Instruction Tuning}},
  author={Liu, Haotian and Li, Chunyuan and Wu, Qingyang and Lee, Yong Jae},
  booktitle=NIPS,
  volume={36},
  pages={34892--34916},
  year={2023}
}

@inproceedings{liu2024improvedllava,
  title={{Improved baselines with visual instruction tuning}},
  author={Liu, Haotian and Li, Chunyuan and Li, Yuheng and Lee, Yong Jae},
  booktitle=CVPR,
  pages={26296--26306},
  year={2024}
}

@article{Qwen3-VL,
      title={{Qwen3-VL Technical Report}}, 
      author={Shuai Bai and Yuxuan Cai and Ruizhe Chen and Keqin Chen and Xionghui Chen and Zesen Cheng and Lianghao Deng and Wei Ding and Chang Gao and Chunjiang Ge and Wenbin Ge and Zhifang Guo and Qidong Huang and Jie Huang and Fei Huang and Binyuan Hui and Shutong Jiang and Zhaohai Li and Mingsheng Li and Mei Li and Kaixin Li and Zicheng Lin and Junyang Lin and Xuejing Liu and Jiawei Liu and Chenglong Liu and Yang Liu and Dayiheng Liu and Shixuan Liu and Dunjie Lu and Ruilin Luo and Chenxu Lv and Rui Men and Lingchen Meng and Xuancheng Ren and Xingzhang Ren and Sibo Song and Yuchong Sun and Jun Tang and Jianhong Tu and Jianqiang Wan and Peng Wang and Pengfei Wang and Qiuyue Wang and Yuxuan Wang and Tianbao Xie and Yiheng Xu and Haiyang Xu and Jin Xu and Zhibo Yang and Mingkun Yang and Jianxin Yang and An Yang and Bowen Yu and Fei Zhang and Hang Zhang and Xi Zhang and Bo Zheng and Humen Zhong and Jingren Zhou and Fan Zhou and Jing Zhou and Yuanzhi Zhu and Ke Zhu},
	  journal={arXiv preprint arXiv:2511.21631},
      year={2025}
}

@article{Qwen2.5-VL,
  title={{Qwen2.5-VL Technical Report}},
  author={Bai, Shuai and Chen, Keqin and Liu, Xuejing and Wang, Jialin and Ge, Wenbin and Song, Sibo and Dang, Kai and Wang, Peng and Wang, Shijie and Tang, Jun and Zhong, Humen and Zhu, Yuanzhi and Yang, Mingkun and Li, Zhaohai and Wan, Jianqiang and Wang, Pengfei and Ding, Wei and Fu, Zheren and Xu, Yiheng and Ye, Jiabo and Zhang, Xi and Xie, Tianbao and Cheng, Zesen and Zhang, Hang and Yang, Zhibo and Xu, Haiyang and Lin, Junyang},
  journal={arXiv preprint arXiv:2502.13923},
  year={2025}
}

@article{Qwen2-VL,
  title={{Qwen2-VL: Enhancing Vision-Language Model's Perception of the World at Any Resolution}},
  author={Wang, Peng and Bai, Shuai and Tan, Sinan and Wang, Shijie and Fan, Zhihao and Bai, Jinze and Chen, Keqin and Liu, Xuejing and Wang, Jialin and Ge, Wenbin and Fan, Yang and Dang, Kai and Du, Mengfei and Ren, Xuancheng and Men, Rui and Liu, Dayiheng and Zhou, Chang and Zhou, Jingren and Lin, Junyang},
  journal={arXiv preprint arXiv:2409.12191},
  year={2024}
}

@article{Qwen-VL,
  title={{Qwen-VL: A Versatile Vision-Language Model for Understanding, Localization, Text Reading, and Beyond}},
  author={Bai, Jinze and Bai, Shuai and Yang, Shusheng and Wang, Shijie and Tan, Sinan and Wang, Peng and Lin, Junyang and Zhou, Chang and Zhou, Jingren},
  journal={arXiv preprint arXiv:2308.12966},
  year={2023}
}

@inproceedings{lai2024lisa,
  title={{LISA: Reasoning Segmentation via Large Language Model}},
  author={Lai, Xin and Tian, Zhuotao and Chen, Yukang and Li, Yanwei and Yuan, Yuhui and Liu, Shu and Jia, Jiaya},
  booktitle=CVPR,
  pages={9579--9589},
  year={2024}
}

@inproceedings{kirillov2023sam,
  title={{Segment Anything}},
  author={Kirillov, Alexander and Mintun, Eric and Ravi, Nikhila and Mao, Hanzi and Rolland, Chloe and Gustafson, Laura and Xiao, Tete and Whitehead, Spencer and Berg, Alexander C and Lo, Wan-Yen and others},
  booktitle=ICCV,
  pages={4015--4026},
  year={2023}
}

@article{ravi2024sam2,
  title={{SAM 2: Segment Anything in Images and Videos}},
  author={Ravi, Nikhila and Gabeur, Valentin and Hu, Yuan-Ting and Hu, Ronghang and Ryali, Chaitanya and Ma, Tengyu and Khedr, Haitham and R{\"a}dle, Roman and Rolland, Chloe and Gustafson, Laura and others},
  journal={arXiv preprint arXiv:2408.00714},
  year={2024}
}

@inproceedings{liu2025unipixel,
  title={{UniPixel: Unified Object Referring and Segmentation for Pixel-Level Visual Reasoning}},
  author={Liu, Ye and Ma, Zongyang and Pu, Junfu and Qi, Zhongang and Wu, Yang and Ying, Shan and Chen, Chang Wen},
  booktitle=NIPS,
  year={2025}
}

@article{shao2024grpo,
  title={{DeepSeekMath: Pushing the Limits of Mathematical Reasoning in Open Language Models}},
  author={Shao, Zhihong and Wang, Peiyi and Zhu, Qihao and Xu, Runxin and Song, Junxiao and Bi, Xiao and Zhang, Haowei and Zhang, Mingchuan and Li, YK and Wu, Yang and others},
  journal={arXiv preprint arXiv:2402.03300},
  year={2024}
}

@article{Zheng2025gspo,
  title={{Group Sequence Policy Optimization}},
  author={Chujie Zheng and Shixuan Liu and Mingze Li and Xiong-Hui Chen and Bowen Yu and Chang Gao and Kai Dang and Yuqiong Liu and Rui Men and An Yang and Jingren Zhou and Junyang Lin},
  journal={ArXiv},
  year={2025},
  volume={abs/2507.18071},
}

@inproceedings{liu2023gres,
  title={{GRES: Generalized Referring Expression Segmentation}},
  author={Liu, Chang and Ding, Henghui and Jiang, Xudong},
  booktitle=CVPR,
  pages={23592--23601},
  year={2023}
}

@inproceedings{zou2023xdecoder,
  title={{Generalized Decoding for Pixel, Image, and Language}},
  author={Zou, Xueyan and Dou, Zi-Yi and Yang, Jianwei and Gan, Zhe and Li, Linjie and Li, Chunyuan and Dai, Xiyang and Behl, Harkirat and Wang, Jianfeng and Yuan, Lu and others},
  booktitle=CVPR,
  pages={15116--15127},
  year={2023}
}

@inproceedings{zou2023seem,
  title={{Segment Everything Everywhere All at Once}},
  author={Zou, Xueyan and Yang, Jianwei and Zhang, Hao and Li, Feng and Li, Linjie and Wang, Jianfeng and Wang, Lijuan and Gao, Jianfeng and Lee, Yong Jae},
  booktitle=NIPS,
  volume={36},
  pages={19769--19782},
  year={2023}
}

@inproceedings{nextchat,
author = {Zhang, Ao and Yao, Yuan and Ji, Wei and Liu, Zhiyuan and Chua, Tat-Seng},
title = {{NExT-Chat: An LMM for Chat, Detection and Segmentation}},
year = {2024},
booktitle = ICML,
articleno = {2487},
numpages = {18},
}

@inproceedings{ren2024pixellm,
  title={{PixelLM: Pixel Reasoning with Large Multimodal Model}},
  author={Ren, Zhongwei and Huang, Zhicheng and Wei, Yunchao and Zhao, Yao and Fu, Dongmei and Feng, Jiashi and Jin, Xiaojie},
  booktitle=CVPR,
  pages={26374--26383},
  year={2024}
}

@inproceedings{zhang2024groundhog,
  title={{ROUNDHOG: Grounding Large Language Models to Holistic Segmentation}},
  author={Zhang, Yichi and Ma, Ziqiao and Gao, Xiaofeng and Shakiah, Suhaila and Gao, Qiaozi and Chai, Joyce},
  booktitle=CVPR,
  pages={14227--14238},
  year={2024}
}

@article{wei2024lasagna,
  title={{LaSagnA: Language-based Segmentation Assistant for Complex Queries}},
  author={Wei, Cong and Tan, Haoxian and Zhong, Yujie and Yang, Yujiu and Ma, Lin},
  journal={arXiv preprint arXiv:2404.08506},
  year={2024}
}

@inproceedings{jang2025m2sa,
title={{MMR: A Large-scale Benchmark Dataset for Multi-target and Multi-granularity Reasoning Segmentation}},
author={Donggon Jang and Yucheol Cho and Suin Lee and Taehyeon Kim and Daeshik Kim},
booktitle=ICLR,
year={2025}
}

@inproceedings{rasheed2024glamm,
  title={{GLaMM: Pixel Grounding Large Multimodal Model}},
  author={Rasheed, Hanoona and Maaz, Muhammad and Shaji, Sahal and Shaker, Abdelrahman and Khan, Salman and Cholakkal, Hisham and Anwer, Rao M and Xing, Eric and Yang, Ming-Hsuan and Khan, Fahad S},
  booktitle=CVPR,
  pages={13009--13018},
  year={2024}
}

@inproceedings{bai2024videolisa,
  title={{One Token to Seg Them All: Language Instructed Reasoning Segmentation in Videos}},
  author={Bai, Zechen and He, Tong and Mei, Haiyang and Wang, Pichao and Gao, Ziteng and Chen, Joya and Liu, Lei and Zhang, Zheng and Shou, Mike Z},
  booktitle=NIPS,
  volume={37},
  pages={6833--6859},
  year={2024}
}

@inproceedings{ji2025robobrain,
  title={{RoboBrain: A Unified Brain Model for Robotic Manipulation from Abstract to Concrete}},
  author={Ji, Yuheng and Tan, Huajie and Shi, Jiayu and Hao, Xiaoshuai and Zhang, Yuan and Zhang, Hengyuan and Wang, Pengwei and Zhao, Mengdi and Mu, Yao and An, Pengju and others},
  booktitle=CVPR,
  pages={1724--1734},
  year={2025}
}

@inproceedings{xu2025a0,
  title={{A0: An Affordance-Aware Hierarchical Model for General Robotic Manipulation}},
  author={Xu, Rongtao and Zhang, Jian and Guo, Minghao and Wen, Youpeng and Yang, Haoting and Lin, Min and Huang, Jianzheng and Li, Zhe and Zhang, Kaidong and Wang, Liqiong and others},
  booktitle=ICCV,
  pages={13491--13501},
  year={2025}
}

@article{ma2024glover,
  title={{GLOVER: Generalizable Open-Vocabulary Affordance Reasoning for Task-Oriented Grasping}},
  author={Ma, Teli and Wang, Zifan and Zhou, Jiaming and Wang, Mengmeng and Liang, Junwei},
  journal={arXiv preprint arXiv:2411.12286},
  year={2024}
}

@inproceedings{yan2024visa,
  title={{VISA: Reasoning Video Object Segmentation via Large Language Models}},
  author={Yan, Cilin and Wang, Haochen and Yan, Shilin and Jiang, Xiaolong and Hu, Yao and Kang, Guoliang and Xie, Weidi and Gavves, Efstratios},
  booktitle=ECCV,
  pages={98--115},
  year={2024},
}

@inproceedings{fei2024vitron,
  title={{Vitron: A Unified Pixel-level Vision LLM for Understanding, Generating, Segmenting, Editing}},
  author={Fei, Hao and Wu, Shengqiong and Zhang, Hanwang and Chua, Tat-Seng and Yan, Shuicheng},
  booktitle=NIPS,
  volume={37},
  pages={57207--57239},
  year={2024}
}

@article{yuan2025sa2va,
  title={{Sa2VA: Marrying SAM2 with LLaVA for Dense Grounded Understanding of Images and Videos}},
  author={Yuan, Haobo and Li, Xiangtai and Zhang, Tao and Sun, Yueyi and Huang, Zilong and Xu, Shilin and Ji, Shunping and Tong, Yunhai and Qi, Lu and Feng, Jiashi and others},
  journal={arXiv preprint arXiv:2501.04001},
  year={2025}
}

@inproceedings{wu2022referformer,
  title={{Language as Queries for Referring Video Object Segmentation}},
  author={Wu, Jiannan and Jiang, Yi and Sun, Peize and Yuan, Zehuan and Luo, Ping},
  booktitle=CVPR,
  pages={4974--4984},
  year={2022}
}

@inproceedings{ding2023mevis,
  title={{MeViS: A Large-scale Benchmark for Video Segmentation with Motion Expressions}},
  author={Ding, Henghui and Liu, Chang and He, Shuting and Jiang, Xudong and Loy, Chen Change},
  booktitle=ICCV,
  pages={2694--2703},
  year={2023}
}

@inproceedings{zhong2022regionclip,
  title={{RegionCLIP: Region-based Language-Image Pretraining}},
  author={Zhong, Yiwu and Yang, Jianwei and Zhang, Pengchuan and Li, Chunyuan and Codella, Noel and Li, Liunian Harold and Zhou, Luowei and Dai, Xiyang and Yuan, Lu and Li, Yin and others},
  booktitle=CVPR,
  pages={16793--16803},
  year={2022}
}

@inproceedings{
zhu2021deformable,
title={{Deformable DETR: Deformable Transformers for End-to-End Object Detection}},
author={Xizhou Zhu and Weijie Su and Lewei Lu and Bin Li and Xiaogang Wang and Jifeng Dai},
booktitle=ICLR,
year={2021},
}

@inproceedings{
kuo2023openvocabulary,
title={{Open-Vocabulary Object Detection upon Frozen Vision and Language Models}},
author={Weicheng Kuo and Yin Cui and Xiuye Gu and AJ Piergiovanni and Anelia Angelova},
booktitle=ICLR,
year={2023},
}

@inproceedings{du2022learning,
  title={{Learning to Prompt for Open-Vocabulary Object Detection with Vision-Language Model}},
  author={Du, Yu and Wei, Fangyun and Zhang, Zihe and Shi, Miaojing and Gao, Yue and Li, Guoqi},
  booktitle=CVPR,
  pages={14084--14093},
  year={2022}
}

@inproceedings{wu2023aligning,
  title={{Aligning Bag of Regions for Open-Vocabulary Object Detection}},
  author={Wu, Size and Zhang, Wenwei and Jin, Sheng and Liu, Wentao and Loy, Chen Change},
  booktitle=CVPR,
  pages={15254--15264},
  year={2023}
}

@inproceedings{minderer2022owlvit,
  title={{Simple Open-Vocabulary Object Detection with Vision Transformers}},
  author={Minderer, Matthias and Gritsenko, Alexey and Stone, Austin and Neumann, Maxim and Weissenborn, Dirk and Dosovitskiy, Alexey and Mahendran, Aravindh and Arnab, Anurag and Dehghani, Mostafa and Shen, Zhuoran and others},
  booktitle=ECCV,
  pages={728--755},
  year={2022},
}

@inproceedings{ma2023codet,
  title={{CoDet: Co-Occurrence Guided Region-Word Alignment for Open-Vocabulary Object Detection}},
  author={Ma, Chuofan and Jiang, Yi and Wen, Xin and Yuan, Zehuan and Qi, Xiaojuan},
  booktitle=NIPS,
  volume={36},
  pages={71078--71094},
  year={2023}
}

@inproceedings{shin2024PixelCLIP,
  title={{Towards Open-Vocabulary Semantic Segmentation Without Semantic Labels}},
  author={Shin, Heeseong and Kim, Chaehyun and Hong, Sunghwan and Cho, Seokju and Arnab, Anurag and Seo, Paul Hongsuck and Kim, Seungryong},
  booktitle=NIPS,
  volume={37},
  pages={9153--9177},
  year={2024}
}

@inproceedings{dong2023maskclip,
  title={{MaskCLIP: Masked Self-Distillation Advances Contrastive Language-Image Pretraining}},
  author={Dong, Xiaoyi and Bao, Jianmin and Zheng, Yinglin and Zhang, Ting and Chen, Dongdong and Yang, Hao and Zeng, Ming and Zhang, Weiming and Yuan, Lu and Chen, Dong and others},
  booktitle=CVPR,
  pages={10995--11005},
  year={2023}
}

@inproceedings{
wei2023ovparts,
title={{OV-PARTS: Towards Open-Vocabulary Part Segmentation}},
author={Meng Wei and Xiaoyu Yue and Wenwei Zhang and Shu Kong and Xihui Liu and Jiangmiao Pang},
booktitle=NIPS,
year={2023},
}

@inproceedings{liang2023ovseg,
  title={{Open-Vocabulary Semantic Segmentation with Mask-adapted CLIP}},
  author={Liang, Feng and Wu, Bichen and Dai, Xiaoliang and Li, Kunpeng and Zhao, Yinan and Zhang, Hang and Zhang, Peizhao and Vajda, Peter and Marculescu, Diana},
  booktitle=CVPR,
  pages={7061--7070},
  year={2023}
}

@article{yu2023fcclip,
  title={{Convolutions Die Hard: Open-Vocabulary Segmentation with Single Frozen Convolutional CLIP}},
  author={Yu, Qihang and He, Ju and Deng, Xueqing and Shen, Xiaohui and Chen, Liang-Chieh},
  journal=NIPS,
  volume={36},
  pages={32215--32234},
  year={2023}
}

@inproceedings{he2022partimagenet,
  title={{PartImageNet: A Large, High-Quality Dataset of Parts}},
  author={He, Ju and Yang, Shuo and Yang, Shaokang and Kortylewski, Adam and Yuan, Xiaoding and Chen, Jie-Neng and Liu, Shuai and Yang, Cheng and Yu, Qihang and Yuille, Alan},
  booktitle=ECCV,
  pages={128--145},
  year={2022},
}

@inproceedings{pan2023partseg,
  title={{Towards Open-World Segmentation of Parts}},
  author={Pan, Tai-Yu and Liu, Qing and Chao, Wei-Lun and Price, Brian},
  booktitle=CVPR,
  pages={15392--15401},
  year={2023}
}

@INPROCEEDINGS{cvpr2025dualseg,
  author={Wang, Zhengyang and Feng, Tingliang and Lyu, Fan and Shang, Fanhua and Feng, Wei and Wan, Liang},
  booktitle=CVPR, 
  title={{Dual Semantic Guidance for Open Vocabulary Semantic Segmentation}}, 
  year={2025},
  volume={},
  number={},
  pages={20212-20222},
}

@INPROCEEDINGS{cvpr2025dpseg,
  author={Zhao, Ziyu and Li, Xiaoguang and Shi, Linjia and Imanpour, Nasrin and Wang, Song},
  booktitle=CVPR, 
  title={{DPSeg: Dual-Prompt Cost Volume Learning for Open-Vocabulary Semantic Segmentation}}, 
  year={2025},
  volume={},
  number={},
  pages={25346-25356},
}

@INPROCEEDINGS{iros2017ITAFF,
  author={Nguyen, Anh and Kanoulas, Dimitrios and Caldwell, Darwin G. and Tsagarakis, Nikos G.},
  booktitle=IROS, 
  title={{Object-based affordances detection with Convolutional Neural Networks and dense Conditional Random Fields}}, 
  year={2017},
  volume={},
  number={},
  pages={5908-5915},
}

@article{ijcv2022osoad,
author = {Zhai, Wei and Luo, Hongchen and Zhang, Jing and Cao, Yang and Tao, Dacheng},
title = {One-Shot Object Affordance Detection in the Wild},
year = {2022},
volume = {130},
number = {10},
journal =IJCV,
pages = {2472–2500},
numpages = {29},
}

@INPROCEEDINGS{Chuang2017LearningTA,
  title={{Learning to Act Properly: Predicting and Explaining Affordances from Images}},
  author={Ching-Yao Chuang and Jiaman Li and Antonio Torralba and Sanja Fidler},
  booktitle=CVPR,
  year={2017},
  pages={975-983},
}

@INPROCEEDINGS{iccv2023cotdet,
  author={Tang, Jiajin and Zheng, Ge and Yu, Jingyi and Yang, Sibei},
  booktitle=ICCV, 
  title={{CoTDet: Affordance Knowledge Prompting for Task Driven Object Detection}}, 
  year={2023},
  volume={},
  number={},
  pages={3045-3055},
}

@InProceedings{Wang_2024_CVPR,
    author    = {Wang, Xiaohan and Liu, Yuehu and Song, Xinhang and Liu, Yuyi and Zhang, Sixian and Jiang, Shuqiang},
    title     = {{An Interactive Navigation Method with Effect-oriented Affordance}},
    booktitle = CVPR,
    year      = {2024},
    pages     = {16446-16456}
}

@article{jiang2025affordancesam,
  title={{AffordanceSAM: Segment Anything Once More in Affordance Grounding}},
  author={Jiang, Dengyang and Wang, Zanyi and Li, Hengzhuang and Dang, Sizhe and Ma, Teli and Wei, Wei and Dai, Guang and Zhang, Lei and Wang, Mengmeng},
  journal={arXiv preprint arXiv:2504.15650},
  year={2025}
}

@inproceedings{
xu2025weaklysupervised,
title={{Weakly-Supervised Affordance Grounding Guided by Part-Level Semantic Priors}},
author={Peiran Xu and Yadong MU},
booktitle=ICLR,
year={2025},
}

@inproceedings{tang2025closed,
  title={{Closed-Loop Transfer for Weakly-supervised Affordance Grounding}},
  author={Tang, Jiajin and Wei, Zhengxuan and Zheng, Ge and Yang, Sibei},
  booktitle=ICCV,
  pages={9530--9539},
  year={2025}
}

@inproceedings{moon2025selective,
  title={{Selective Contrastive Learning for Weakly Supervised Affordance Grounding}},
  author={Moon, WonJun and Seong, Hyun Seok and Heo, Jae-Pil},
  booktitle=ICCV,
  pages={5210--5220},
  year={2025}
}

@INPROCEEDINGS{cvpr2025reasonmamba,
  author={Wang, Yuxuan and Wu, Aming and Yang, Muli and Min, Yukuan and Zhu, Yihang and Deng, Cheng},
  booktitle=CVPR, 
  title={{Reasoning Mamba: Hypergraph-Guided Region Relation Calculating for Weakly Supervised Affordance Grounding}}, 
  year={2025},
  volume={},
  number={},
  pages={27618-27627},
}

@inproceedings{li2023locate,
  title={{LOCATE: Localize and Transfer Object Parts for Weakly Supervised Affordance Grounding}},
  author={Li, Gen and Jampani, Varun and Sun, Deqing and Sevilla-Lara, Laura},
  booktitle=CVPR,
  pages={10922--10931},
  year={2023}
}

@article{tong2025oval,
  title={{OVAL-Grasp: Open-Vocabulary Affordance Localization for Task Oriented Grasping}},
  author={Tong, Edmond and Balaji, Advaith and Opipari, Anthony and Lewis, Stanley and Zeng, Zhen and Jenkins, Odest Chadwicke},
  journal={arXiv preprint arXiv:2511.20841},
  year={2025}
}

@inproceedings{ju2024roboabc,
  title={{Robo-abc: Affordance generalization beyond categories via semantic correspondence for robot manipulation}},
  author={Ju, Yuanchen and Hu, Kaizhe and Zhang, Guowei and Zhang, Gu and Jiang, Mingrun and Xu, Huazhe},
  booktitle=ECCV,
  pages={222--239},
  year={2024},
}

@article{xue2022useek,
  title={{USEEK: Unsupervised SE(3)-Equivariant 3D Keypoints for Generalizable Manipulation}},
  author={Xue, Zhengrong and Yuan, Zhecheng and Wang, Jiashun and Wang, Xueqian and Gao, Yang and Xu, Huazhe},
  journal={arXiv preprint arXiv:2209.13864},
  year={2022}
}

@inproceedings{jiang2024doduo,
  title={{Doduo: Learning Dense Visual Correspondence from Unsupervised Semantic-Aware Flow}},
  author={Jiang, Zhenyu and Jiang, Hanwen and Zhu, Yuke},
  booktitle=ICRA,
  pages={12420--12427},
  year={2024},
}

@article{huang2024rekep,
  title={{ReKep: Spatio-Temporal Reasoning of Relational Keypoint Constraints for Robotic Manipulation}},
  author={Huang, Wenlong and Wang, Chen and Li, Yunzhu and Zhang, Ruohan and Fei-Fei, Li},
  journal={arXiv preprint arXiv:2409.01652},
  year={2024}
}

@inproceedings{vuong2024lgd,
  title={{Language-driven Grasp Detection}},
  author={Vuong, An Dinh and Vu, Minh Nhat and Huang, Baoru and Nguyen, Nghia and Le, Hieu and Vo, Thieu and Nguyen, Anh},
  booktitle=CVPR,
  pages={17902--17912},
  year={2024}
}

@inproceedings{noh2025graspsam,
  title={{GraspSAM: When Segment Anything Model Meets Grasp Detection}},
  author={Noh, Sangjun and Kim, Jongwon and Nam, Dongwoo and Back, Seunghyeok and Kang, Raeyoung and Lee, Kyoobin},
  booktitle=ICRA,
  pages={14023--14029},
  year={2025},
}

@INPROCEEDINGS{iroslightlgd,
  author={Nguyen, Nghia and Vu, Minh Nhat and Huang, Baoru and Vuong, An and Le, Ngan and Vo, Thieu and Nguyen, Anh},
  booktitle=IROS, 
  title={{Lightweight Language-driven Grasp Detection using Conditional Consistency Model}}, 
  year={2024},
  volume={},
  number={},
  pages={13719-13725}
}

@INPROCEEDINGS{iros2025graspmamba,
  author={Nguyen, Huy Hoang and Vuong, An and Nguyen, Anh and Reid, Ian and Vu, Minh Nhat},
  booktitle=IROS, 
  title={{GraspMamba: A Mamba-based Language-driven Grasp Detection Framework with Hierarchical Feature Learning}}, 
  year={2025},
  volume={},
  number={},
  pages={15808-15815},
}

@INPROCEEDINGS{iros2024masklgd,
  author={Van Vo, Tuan and Vu, Minh Nhat and Huang, Baoru and Vuong, An and Le, Ngan and Vo, Thieu and Nguyen, Anh},
  booktitle=IROS, 
  title={Language-driven Grasp Detection with Mask-guided Attention}, 
  year={2024},
  volume={},
  number={},
  pages={7492-7498},
}

@inproceedings{nips2020gpt3,
author = {Brown, Tom B. and Mann, Benjamin and Ryder, Nick and Subbiah, Melanie and Kaplan, Jared and Dhariwal, Prafulla and Neelakantan, Arvind and Shyam, Pranav and Sastry, Girish and Askell, Amanda and Agarwal, Sandhini and Herbert-Voss, Ariel and Krueger, Gretchen and Henighan, Tom and Child, Rewon and Ramesh, Aditya and Ziegler, Daniel M. and Wu, Jeffrey and Winter, Clemens and Hesse, Christopher and Chen, Mark and Sigler, Eric and Litwin, Mateusz and Gray, Scott and Chess, Benjamin and Clark, Jack and Berner, Christopher and McCandlish, Sam and Radford, Alec and Sutskever, Ilya and Amodei, Dario},
title = {{Language Models are Few-Shot Learners}},
year = {2020},
booktitle = NIPS,
articleno = {159},
numpages = {25},
}

@inproceedings{nips2022rlhf,
author = {Ouyang, Long and Wu, Jeff and Jiang, Xu and Almeida, Diogo and Wainwright, Carroll L. and Mishkin, Pamela and Zhang, Chong and Agarwal, Sandhini and Slama, Katarina and Ray, Alex and Schulman, John and Hilton, Jacob and Kelton, Fraser and Miller, Luke and Simens, Maddie and Askell, Amanda and Welinder, Peter and Christiano, Paul and Leike, Jan and Lowe, Ryan},
title = {{Training Language Models to Follow Instructions with Human Feedback}},
year = {2022},
booktitle =NIPS,
articleno = {2011},
numpages = {15},
}

@inproceedings{munasinghe2025videoglamm,
  title={{VideoGLaMM: A Large Multimodal Model for Pixel-Level Visual Grounding in Videos}},
  author={Munasinghe, Shehan and Gani, Hanan and Zhu, Wenqi and Cao, Jiale and Xing, Eric and Khan, Fahad Shahbaz and Khan, Salman},
  booktitle=CVPR,
  pages={19036--19046},
  year={2025}
}

@inproceedings{zheng2025villa,
  title={{ViLLa: Video Reasoning Segmentation with Large Language Model}},
  author={Zheng, Rongkun and Qi, Lu and Chen, Xi and Wang, Yi and Wang, Kun and Zhao, Hengshuang},
  booktitle=ICCV,
  pages={23667--23677},
  year={2025}
}

@inproceedings{lin2025glus,
  title={{GLUS: Global-Local Reasoning Unified into A Single Large Language Model for Video Segmentation}},
  author={Lin, Lang and Yu, Xueyang and Pang, Ziqi and Wang, Yu-Xiong},
  booktitle=CVPR,
  pages={8658--8667},
  year={2025}
}

@inproceedings{deng2025groundmore,
  title={{Motion-Grounded Video Reasoning: Understanding and Perceiving Motion at Pixel Level}},
  author={Deng, Andong and Chen, Tongjia and Yu, Shoubin and Yang, Taojiannan and Spencer, Lincoln and Tian, Yapeng and Mian, Ajmal Saeed and Bansal, Mohit and Chen, Chen},
  booktitle=CVPR,
  pages={8625--8636},
  year={2025}
}

@inproceedings{li2023r2vos,
  title={{Towards Robust Referring Video Object Segmentation with Cyclic Relational Consistency}},
  author={Li, Xiang and Wang, Jinglu and Xu, Xiaohao and Li, Xiao and Raj, Bhiksha and Lu, Yan},
  booktitle=ICCV,
  pages={22236--22245},
  year={2023}
}

@article{tancik2020fouriermapping,
  title={{Fourier Features Let Networks Learn High Frequency Functions in Low Dimensional Domains}},
  author={Tancik, Matthew and Srinivasan, Pratul and Mildenhall, Ben and Fridovich-Keil, Sara and Raghavan, Nithin and Singhal, Utkarsh and Ramamoorthi, Ravi and Barron, Jonathan and Ng, Ren},
  journal=NIPS,
  volume={33},
  pages={7537--7547},
  year={2020}
}

@inproceedings{refcoco,
    title = {{ReferItGame: Referring to Objects in Photographs of Natural Scenes}},
    author = "Kazemzadeh, Sahar  and
      Ordonez, Vicente  and
      Matten, Mark  and
      Berg, Tamara",
    booktitle = EMNLP,
    year = "2014",
}

@inproceedings{refcocog,
  title={{Generation and Comprehension of Unambiguous Object Descriptions}},
  author={Mao, Junhua and Huang, Jonathan and Toshev, Alexander and Camburu, Oana and Yuille, Alan L and Murphy, Kevin},
  booktitle=CVPR,
  pages={11--20},
  year={2016}
}

@InProceedings{urvos,
author="Seo, Seonguk
and Lee, Joon-Young
and Han, Bohyung",
title={{URVOS: Unified Referring Video Object Segmentation Network with a Large-Scale Benchmark}},
booktitle=ECCV,
year="2020",
pages="208--223",
}

@INPROCEEDINGS{ade20k,
  author={Zhou, Bolei and Zhao, Hang and Puig, Xavier and Fidler, Sanja and Barriuso, Adela and Torralba, Antonio},
  booktitle=CVPR, 
  title={{Scene Parsing through ADE20K Dataset}}, 
  year={2017},
  pages={5122-5130},
}

@inproceedings{caesar2018cocostuff,
  title={{COCO-Stuff: Thing and Stuff Classes in Context}},
  author={Caesar, Holger and Uijlings, Jasper and Ferrari, Vittorio},
  booktitle=CVPR,
  pages={1209--1218},
  year={2018}
}

@INPROCEEDINGS {mapillary,
author = { Neuhold, Gerhard and Ollmann, Tobias and Bulo, Samuel Rota and Kontschieder, Peter },
booktitle =ICCV,
title = {{ The Mapillary Vistas Dataset for Semantic Understanding of Street Scenes }},
year = {2017},
pages = {5000-5009},
}

@inproceedings{ramanathan2023paco,
  title={{PACO: Parts and Attributes of Common Objects}},
  author={Ramanathan, Vignesh and Kalia, Anmol and Petrovic, Vladan and Wen, Yi and Zheng, Baixue and Guo, Baishan and Wang, Rui and Marquez, Aaron and Kovvuri, Rama and Kadian, Abhishek and others},
  booktitle=CVPR,
  pages={7141--7151},
  year={2023}
}

@inproceedings{chen2014pacopart,
  title={{Detect What You Can: Detecting and Representing Objects using Holistic Models and Body Parts}},
  author={Chen, Xianjie and Mottaghi, Roozbeh and Liu, Xiaobai and Fidler, Sanja and Urtasun, Raquel and Yuille, Alan},
  booktitle=CVPR,
  pages={1971--1978},
  year={2014}
}

@article{davis2017,
  title={The 2017 DAVIS Challenge on Video Object Segmentation},
  author={Pont-Tuset, Jordi and Perazzi, Federico and Caelles, Sergi and Arbel{\'a}ez, Pablo and Sorkine-Hornung, Alex and Van Gool, Luc},
  journal={arXiv preprint arXiv:1704.00675},
  year={2017}
}

@inproceedings{wang2023lvvis,
  title={{Towards Open-Vocabulary Video Instance Segmentation}},
  author={Wang, Haochen and Yan, Cilin and Wang, Shuai and Jiang, Xiaolong and Tang, Xu and Hu, Yao and Xie, Weidi and Gavves, Efstratios},
  booktitle=ICCV,
  pages={4057--4066},
  year={2023}
}

@inproceedings{athar2025vicas,
  title={{ViCaS: A Dataset for Combining Holistic and Pixel-level Video Understanding using Captions with Grounded Segmentation}},
  author={Athar, Ali and Deng, Xueqing and Chen, Liang-Chieh},
  booktitle=CVPR,
  pages={19023--19035},
  year={2025}
}

@misc{alibaba2025qwenvlmax,
  author       = {Alibaba Cloud},
  title        = {{Qwen-VL-Max: Alibaba's large visual language model}},
  year         = {2025},
  howpublished = {\url{https://modelstudio.console.alibabacloud.com/?tab=doc\#/doc/?type=model&url=2840914_2&modelId=qwen-vl-max}},
  note         = {Accessed: 2025-11-06},
}

@misc{deepmind_gemini_flash,
  title        = {{Gemini 2.5 Flash}},
  author       = {{Google DeepMind}},
  howpublished = {\url{https://deepmind.google/models/gemini/flash/}},
  note         = {Accessed: 2025-11-06},
  year         = {2025}
}

@inproceedings{sun2023vlpart,
  title={{Going Denser with Open-Vocabulary Part Segmentation}},
  author={Sun, Peize and Chen, Shoufa and Zhu, Chenchen and Xiao, Fanyi and Luo, Ping and Xie, Saining and Yan, Zhicheng},
  booktitle=ICCV,
  pages={15453--15465},
  year={2023}
}

@inproceedings{liu2024groundingdino,
  title={{Grounding DINO: Marrying DINO with Grounded Pre-Training for Open-Set Object Detection}},
  author={Liu, Shilong and Zeng, Zhaoyang and Ren, Tianhe and Li, Feng and Zhang, Hao and Yang, Jie and Jiang, Qing and Li, Chunyuan and Yang, Jianwei and Su, Hang and others},
  booktitle=ECCV,
  pages={38--55},
  year={2024},
}

@inproceedings{xiao2024florence2,
  title={{Florence-2: Advancing a Unified Representation for a Variety of Vision Tasks}},
  author={Xiao, Bin and Wu, Haiping and Xu, Weijian and Dai, Xiyang and Hu, Houdong and Lu, Yumao and Zeng, Michael and Liu, Ce and Yuan, Lu},
  booktitle=CVPR,
  pages={4818--4829},
  year={2024}
}

@inproceedings{wu2025ragnet,
  title={{RAGNet: Large-scale Reasoning-based Affordance Segmentation Benchmark towards General Grasping}},
  author={Wu, Dongming and Fu, Yanping and Huang, Saike and Liu, Yingfei and Jia, Fan and Liu, Nian and Dai, Feng and Wang, Tiancai and Anwer, Rao Muhammad and Khan, Fahad Shahbaz and others},
  booktitle=ICCV,
  pages={11980--11990},
  year={2025}
}

@inproceedings{qian20233doi,
  title={{Understanding 3D Object Interaction from a Single Image}},
  author={Qian, Shengyi and Fouhey, David F},
  booktitle=ICCV,
  pages={21753--21763},
  year={2023}
}

@inproceedings{guo2023handal,
  title={{HANDAL: A Dataset of Real-World Manipulable Object Categories with Pose Annotations, Affordances, and Reconstructions}},
  author={Guo, Andrew and Wen, Bowen and Yuan, Jianhe and Tremblay, Jonathan and Tyree, Stephen and Smith, Jeffrey and Birchfield, Stan},
  booktitle=IROS,
  pages={11428--11435},
  year={2023},
}

@inproceedings{zhu2023egoobjects,
  title={{EgoObjects: A Large-Scale Egocentric Dataset for Fine-Grained Object Understanding}},
  author={Zhu, Chenchen and Xiao, Fanyi and Alvarado, Andr{\'e}s and Babaei, Yasmine and Hu, Jiabo and El-Mohri, Hichem and Culatana, Sean and Sumbaly, Roshan and Yan, Zhicheng},
  booktitle=ICCV,
  pages={20110--20120},
  year={2023}
}

@inproceedings{o2024openx,
  title={{Open X-Embodiment: Robotic Learning Datasets and RT-X Models}},
  author={O’Neill, Abby and Rehman, Abdul and Maddukuri, Abhiram and Gupta, Abhishek and Padalkar, Abhishek and Lee, Abraham and Pooley, Acorn and Gupta, Agrim and Mandlekar, Ajay and Jain, Ajinkya and others},
  booktitle=ICRA,
  pages={6892--6903},
  year={2024},
}

@article{james2020rlbench,
  title={{RLBench: The Robot Learning Benchmark \& Learning Environment}},
  author={James, Stephen and Ma, Zicong and Arrojo, David Rovick and Davison, Andrew J},
  journal=RAL,
  volume={5},
  number={2},
  pages={3019--3026},
  year={2020},
}

@inproceedings{fang2020graspnet,
  title={{GraspNet-1Billion: A Large-Scale Benchmark for General Object Grasping}},
  author={Fang, Hao-Shu and Wang, Chenxi and Gou, Minghao and Lu, Cewu},
  booktitle=CVPR,
  pages={11444--11453},
  year={2020}
}

@inproceedings{depierre2018jacquard,
  title={{Jacquard: A Large Scale Dataset for Robotic Grasp Detection}},
  author={Depierre, Amaury and Dellandr{\'e}a, Emmanuel and Chen, Liming},
  booktitle=IROS,
  pages={3511--3516},
  year={2018},
}

@inproceedings{perez2018film,
  title={{FiLM: Visual Reasoning with a General Conditioning Layer}},
  author={Perez, Ethan and Strub, Florian and De Vries, Harm and Dumoulin, Vincent and Courville, Aaron},
  booktitle=AAAI,
  pages={3942--3951},
  year={2018}
}

@article{morrison2018ggcnn,
  title={{Closing the Loop for Robotic Grasping: A Real-time, Generative Grasp Synthesis Approach}},
  author={Morrison, Douglas and Corke, Peter and Leitner, J{\"u}rgen},
  journal={arXiv preprint arXiv:1804.05172},
  year={2018}
}

@INPROCEEDINGS{grcnn,
  author={Kumra, Sulabh and Joshi, Shirin and Sahin, Ferat},
  booktitle=IROS, 
  title={{Antipodal Robotic Grasping using Generative Residual Convolutional Neural Network}}, 
  year={2020},
  volume={},
  number={},
  pages={9626-9633},
}

@INPROCEEDINGS{cornell,
  author={Yun Jiang and Moseson, Stephen and Saxena, Ashutosh},
  booktitle=ICRA, 
  title={{Efficient Grasping from RGBD Images: Learning Using a New Rectangle Representation}}, 
  year={2011},
  volume={},
  number={},
  pages={3304-3311},
}

@INPROCEEDINGS{ocidgrasp,
  author={Ainetter, Stefan and Fraundorfer, Friedrich},
  booktitle=ICRA, 
  title={{End-to-end Trainable Deep Neural Network for Robotic Grasp Detection and Semantic Segmentation from RGB}}, 
  year={2021},
  volume={},
  number={},
  pages={13452-13458},
}

@INPROCEEDINGS{vmrd,
  author={Zhang, Hanbo and Lan, Xuguang and Bai, Site and Zhou, Xinwen and Tian, Zhiqiang and Zheng, Nanning},
  booktitle=IROS, 
  title={{ROI-based Robotic Grasp Detection for Object Overlapping Scenes}}, 
  year={2019},
  volume={},
  number={},
  pages={4768-4775},
}

@INPROCEEDINGS{mttr,
  author={Botach, Adam and Zheltonozhskii, Evgenii and Baskin, Chaim},
  booktitle=CVPR, 
  title={{End-to-End Referring Video Object Segmentation with Multimodal Transformers}}, 
  year={2022},
  volume={},
  number={},
  pages={4975-4985},
}

@article{wei2024hyperseg,
  title={HyperSeg: Towards Universal Visual Segmentation with Large Language Model},
  author={Wei, Cong and Zhong, Yujie and Tan, Haoxian and Liu, Yong and Zhao, Zheng and Hu, Jie and Yang, Yujiu},
  journal={arXiv preprint arXiv:2411.17606},
  year={2024}
}

@article{wei2024instructseg,
  title={InstructSeg: Unifying Instructed Visual Segmentation with Multi-modal Large Language Models},
  author={Wei, Cong and Zhong, Yujie and Tan, Haoxian and Zeng, Yingsen and Liu, Yong and Zhao, Zheng and Yang, Yujiu},
  journal={arXiv preprint arXiv:2412.14006},
  year={2024}
}
